\documentclass{article}

\PassOptionsToPackage{numbers, compress}{natbib}
\usepackage{subcaption}
 \usepackage[preprint]{neurips_2025}

\usepackage{hyperref}       
\usepackage{url}            
\usepackage{booktabs}       
\usepackage{amsfonts}       
\usepackage{nicefrac}       
\usepackage{microtype}      
\usepackage{xcolor}         

\usepackage{graphicx}
\usepackage{booktabs}

\usepackage{multirow}
\usepackage{fontawesome}

\usepackage[accsupp]{axessibility}  

\usepackage{hyperref}

\usepackage{orcidlink}

\usepackage{wrapfig}
\usepackage{marvosym}

\usepackage{tcolorbox}
\tcbuselibrary{skins, breakable}

\newtcolorbox{mycode}{
  breakable,
  colback=gray!10,
  colframe=gray!10,
  fontupper=\small,
  boxrule=0pt,
  arc=2pt,
}

\usepackage{fancyvrb}

\newtcolorbox{finding}[1]{
  enhanced,
  colback=cyan!6!white,
  colframe=cyan!25!white,
  fonttitle=\bfseries\color{black},
  title={#1},
  left=6pt, right=6pt, top=4pt, bottom=4pt,
  boxrule=0.8pt,
}

\title{\textit{MME-CoF-Pro}: Evaluating Reasoning Coherence in Video Generative Models with Text and Visual Hints}

\author{%
  Yu Qi$^{*\dagger1}$,\ Xinyi Xu$^{*1}$,\ Ziyu Guo\textsuperscript{\Letter}$^{*\dagger2}$,\ Siyuan Ma$^{*23}$, Renrui Zhang$^{4}$, Xinyan Chen$^{2}$ \\[0.5em]
  \textbf{Ruichuan An$^{5}$, Ruofan Xing$^{1}$, Jiayi Zhang$^{1}$, Haojie Huang$^{1}$, Pheng-Ann Heng$^{2}$}\\[0.5em]
  \textbf{Jonathan Tremblay$^{6}$, Lawson L.S. Wong$^{1}$} \\[1em]
    $^1$ Northeastern University  $^2$ The Chinese University of Hong Kong   $^3$ Westlake University \\[0.5em]
     $^4$ ByteDance Seed $^5$ Peking University $^6$ NVIDIA
}

\begin{document}

\maketitle

\vspace{-3mm}

\begin{abstract}
  Video generative models show emerging reasoning behaviors. It is essential to ensure that generated events remain causally consistent across frames for reliable deployment, a property we define as \textbf{\textit{reasoning coherence}}. To bridge the gap in literature for missing reasoning coherence evaluation, we propose \textbf{\textit{MME-CoF-Pro}}, a comprehensive video reasoning benchmark to assess reasoning coherence in video models. Specifically, MME-CoF-Pro contains 303 samples across 16 categories, ranging from visual logical to scientific reasoning. It introduces Reasoning Score as evaluation metric for assessing process-level necessary intermediate reasoning steps, and includes three evaluation settings, (a) no hint (b) text hint and (c) visual hint, enabling a controlled investigation into the underlying mechanisms of reasoning hint guidance. Evaluation results in 7 open and closed-source video models reveals insights including: (1) Video generative models exhibit weak reasoning coherence, decoupled from generation quality. (2) Text hints boost apparent correctness but often cause inconsistency and hallucinated reasoning (3) Visual hints benefit structured perceptual tasks but struggle with fine-grained perception. Website: \textbf{\textcolor{red}{\url{https://video-reasoning-coherence.github.io/}}}

  \end{abstract}

\begin{figure}[h]
\centering
\includegraphics[width=0.9\linewidth]{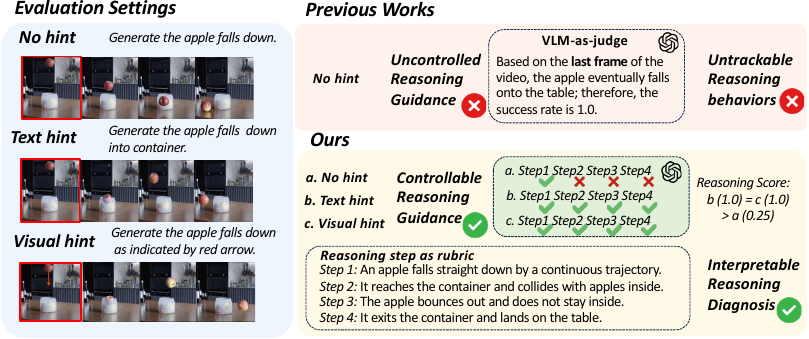}
\caption{
\textbf{\textit{MME-CoF-Pro} evaluates video model capacity to model world behaviour.}
This evaluation is shown under the numbered multi steps section (shown in darker green), this matches with ``reasoning steps as rubric'' section (human generated to match expected behaviour). 
Using different prompting mechanisms, our benchmark evaluates video model behaviour at a fine grain.
Red box indicates input image. 
}
\label{fig:teaser}
\vspace{-5mm}
\end{figure}

\section{Introduction}

\label{sec:intro}

\begin{figure}[t]
\centering
\includegraphics[width=\linewidth]{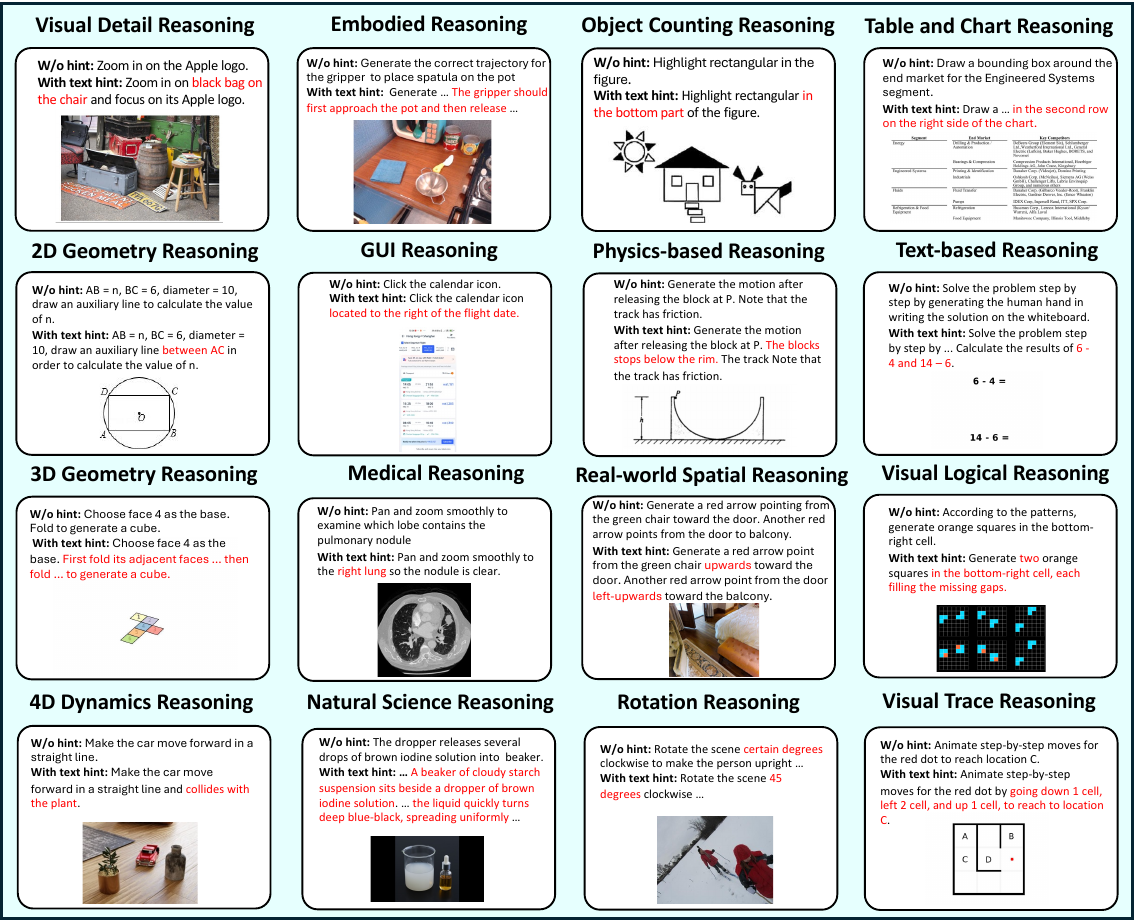}
\caption{\textbf{\textit{MME-CoF-Pro}} consists of 303 samples from 16 basic reasoning categories, covering from \textit{Visual Detail Reasoning} to \textit{Visual Trace Reasoning}. Each sample is provided with no-hint and text-hint evaluation prompt. In text-hint setting, models are provided with additional textual details to facilitate reasoning.
}
\label{fig:mme_cof_pro_text_hint}
\vspace{-5mm}
\end{figure}

\begin{figure}[t]
\centering
\includegraphics[width=\linewidth]{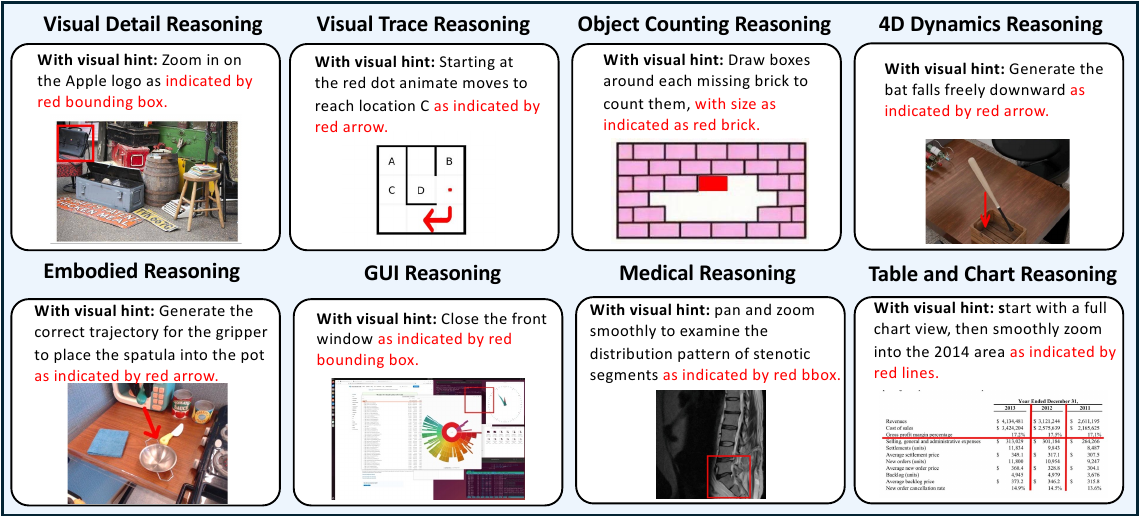}
\caption{\textbf{Visual Hints in \textit{MME-CoF-Pro}}. For controlled comparison, we additionally add visual hints to 8 categories which requires visual reasoning capabilities, using bounding boxes and arrows to indicate positions, trajectories to facilitate reasoning.
}
\label{fig:mme_cof_pro_visual_hints}
\vspace{-5mm}
\end{figure}

Video generative models have rapidly progressed, producing high-fidelity outputs with realistic temporal dynamics. 
Emerging evidence~\cite{wiedemer2025video,gao2025seedance} suggests reasoning abilities such as planning and causal inference. 
It is now possible to prompt a generative video model to simulate an apple falling onto a jar (Fig.~\ref{fig:teaser}), predict the causal outcome of closing a window pane on a computer screen, or even perform zero-shot object identification from a single image.
While these emerging abilities remain largely understudied 
by existing benchmarks, it is essential to ensure that generated events remain causally consistent across frames for reliable deployment. We formalize this gap as \textit{reasoning coherence}: the extent to which generated events follow consistent and plausible cause–effect dynamics across frames.

Recent efforts such as V-ReasonBench~\cite{luo2025v} and Gen-ViRe~\cite{liu2025can} have taken steps toward evaluating reasoning in video generative models, introducing benchmarks across spatial, physical, and abstract dimensions.
While these approaches move beyond surface-level fidelity, they primarily assess outcome correctness through last-frame verification. 
Concurrently, VIPER~\cite{li2025viper} and VBVR~\cite{wang2026very} introduced process-aware and large-scale verifiable evaluation of intermediate reasoning steps, but do not target the consistency of cause-and-effect chains specifically. 
In parallel, 
Morpheus~\cite{zhang2025morpheus} and VideoPhy~\cite{bansal2024videophy} evaluate physical plausibility of individual phenomena but do not assess whether multi-step causal relationships remain coherent across the full generated sequence. 
While these directions are promising, none explicitly capture what we define as \textbf{\textit{reasoning coherence}}: the degree to which generated video events maintain consistent cause-and-effect relationships over time, spanning long-horizon dependencies, interaction reasoning, and the fidelity of intermediate transitions.

\begin{wrapfigure}[18]{r}{0.5\textwidth}
    \vspace{-1em}
    \centering
    \includegraphics[width=0.5\textwidth]{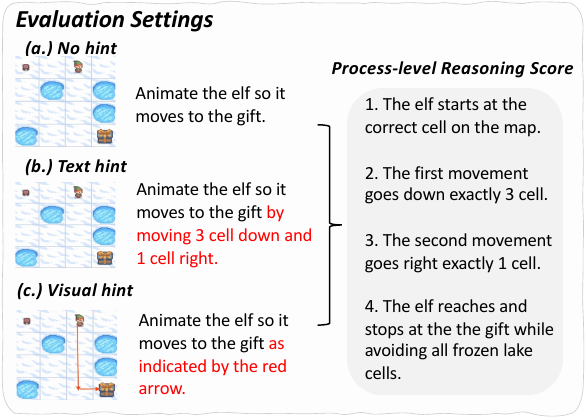}
    \caption{\textbf{Setting Comparisons.} All three settings share the same evaluation metrics and identical instructions except for the hint components.}
    \label{fig:hint_comparison}
    \vspace{-5mm}
\end{wrapfigure}

To bridge this gap, we introduce \textbf{\textit{MME-CoF-Pro}}, a benchmark for evaluating reasoning coherence in video generative models. In greater details, it covers 16 categories and 303 samples, as illustrated in Fig.~\ref{fig:mme_cof_pro_text_hint}. We introduce two key components for evaluating reasoning coherence, the reasoning score, and visual and text prompt-hints. 
First, the \textbf{Reasoning Score (RS)} provides process-level evaluation by assessing intermediate reasoning steps which are necessary in order to accomplish a task, allowing fine-grained analysis of how reasoning unfolds across frames. Each sample in MME-CoF-Pro is paired with a human-validated step-level rubric, scored automatically over \textit{k} equidistant frames sampled from the generated video.
Second, we design three controllable hint prompting settings: \textbf{(a) no hints} (standard setting), \textbf{(b) text hints} (Fig.~\ref{fig:mme_cof_pro_text_hint}), and \textbf{(c) visual hints} (Fig.~\ref{fig:mme_cof_pro_visual_hints}). 
The hint-based settings provide explicit reasoning scaffolds through visual or text guidance, whereas the no-hint setting requires models to reason without external assistance. 
As shown in Fig.~\ref{fig:hint_comparison}, all remaining parts of instructions are kept identical across settings so any performance differences can be directly attributed to the presence of structured reasoning guidance.

Evaluation results of \textbf{\textit{MME-CoF-Pro}} across 7 open and closed-source models reveal insights: 
\textbf{(1)} Video generative models exhibit generally weak reasoning coherence, and their reasoning performance shows no clear correlation with generation quality.
\textbf{(2)} Explicit text hints may improve apparent correctness, but the gains are often superficial and can induce hallucinated reasoning, inflating scores without genuine understanding and leading to partial comprehension or overfitting to reasoning patterns.
\textbf{(3)} Visual hints are more effective for structured and spatially guided tasks but are less reliable for fine-grained visual tasks. They also introduce hallucinations by being mistakenly rendered as part of the scene rather than purely as guidance.
Our contributions can be summarized as follows:

\begin{itemize}
    \item We introduce \textbf{\textit{MME-CoF-Pro}}, the first video reasoning benchmark that explicitly decouples reasoning hints evaluation which covers 303 samples across 16 categories (Fig.~\ref{fig:overview_statistics}).

    \item We propose Reasoning Score, a process-level evaluation metric for evaluate intermediate reasoning steps instead of generation quality against human validated rubrics.

    \item Results on 7 open and closed-source video models uncover key insights including the performance of reasoning capabilities, limitations of explicit hints, the role of visual and text hints.
    
\end{itemize}

\section{Related Works}

\subsection{Video Generative Models as Reasoning Models}

Recent advances in large reasoning models such as DeepSeek-R1~\cite{guo2025deepseek} have accelerated video reasoning. Prior work enhances MLLMs and VLMs~\cite{xu2021vlm, hong2024cogvlm2, lu2024enhancing} with causal reasoning~\cite{maaz2025video, yi2019clevrer, li2025videochat}, spatio-temporal modeling~\cite{yuan2025videorefer, cheng2024videollama, liu2025stop, liao2024videoinsta, li2024videomamba}, and chain-of-thought~\cite{arnab2025temporal, rasheed2025video, fei2024video, zhang2025video}, using RL or supervised fine-tuning~\cite{hu2025sf2t, wang2024knowledge, wang2023foundation}. Recent large-scale video generative models~\cite{wan2025wan, ali2025world, chi2025wow, chiwow, wang2024worlddreamer, seedance2025seedance, gao2025seedance, team2025klingavatar, liu2024sora} implicitly learn temporal dynamics, causal structures, and physical regularities through predictive video modeling. By forecasting future frames, diffusion-based and autoregressive frameworks ~\cite{ho2022imagen, yan2021videogpt} capture object interactions, motion patterns, and long-horizon dependencies. Such temporal consistency and coherent scene evolution suggest that these models acquire structured world knowledge from large-scale video data, even without explicit reasoning supervision.

\subsection{Resaoning Evaluation in Video Generative Models}

Existing image and video reasoning benchmarks propose reasoning-related tasks, but they do not explicitly isolate and control reasoning as an independent evaluation variable in either the input design or the evaluation metrics~\cite{yang2025reasoning, tong2025thinking, luo2025v, haodong2025tivibench, liu2025can, he2025ruler, li2025viper, tian2025envision, liang2025worldlens, an2026geniusgenerativefluidintelligence, an2026visionlanguagemodelsassess, duan2025worldscore}. Prior work evaluates generation quality and creativity in T2V(Text-to-Video) models~\cite{huang2024vbench,liu2024evalcrafter}, comprehensive reasoning tasks~\cite{guo2025video}, maze-based spatial reasoning~\cite{maaz2025video}, synthetic dynamics prediction~\cite{tong2025thinking,luo2025v}, structured reasoning patterns~\cite{chen2025tivibench}, and scientific problem-solving~\cite{deng2025scivideobench, hu2025text2world}. 

\textbf{(1)}~Existing benchmarks embed reasoning guidance implicitly within task prompts rather than treating it as an explicit, controllable input. Without varying guidance independently, these benchmarks cannot measure how explicit reasoning instructions affect performance.
\textbf{(2)}~Existing evaluation metrics mainly focus on instruction alignment~\cite{guo2025video}, visual consistency~\cite{maaz2025video, guo2025video}, rule-based correctness~\cite{he2025ruler}, pass@k based final-frame correctness~\cite{tong2025thinking, luo2025v, chen2025tivibench}, spatial accuracy by hard coded rules~\cite{wang2026very}, goal-conditioned success rate~\cite{maes2026stable}, process-consistency~\cite{li2025viper} but do not evaluate intermediate reasoning processes, limiting their ability to diagnose reasoning trajectory failures or providing targeted improvement suggestions.

In contrast to previous approaches, our proposed \textbf{\textit{MME-CoF-Pro}} treats reasoning prompts as an explicit, controllable variable—injecting text and visual hints into the input while keeping the remaining instructions unchanged. By keeping all other task components identical, any performance difference can be causally attributed to reasoning capability and hint guidance. Finally, we introduce \textbf{\textit{Reasoning Score}}, a process-level evaluation metric for step-wise reasoning coherence (generated with humans), enabling interpretable diagnosis of where and why models fail.

\section{The MME-CoF-Pro Benchmark}

\begin{figure*}[t]
\centering
\begin{minipage}{\textwidth}
    \subcaptionbox{Key Statistics.\label{fig:overview:a}}[0.30\textwidth]{
        \centering
        \fontsize{8pt}{\baselineskip}\selectfont
        \renewcommand\tabcolsep{0.9pt}
        \renewcommand\arraystretch{0.7}
        \scalebox{0.70}{
            \begin{tabular}{lc}
            \toprule
            \textbf{Statistic} & \textbf{Number} \\
            \midrule
            Total questions & 673 \\
            - no hint questions & 303  \\
            - text hint questions & 303  \\
            - visual hint questions & 67 \\
            Total samples & 303  \\
            Total images & 370 \\
            \midrule
             Total categories  & 16 \\
             Categories with visual hints & 8 \\
            \midrule
            Max question words & 191 \\
            Avg question words & 81.6 \\
            \midrule
            Max reasoning step & 10\\
            Avg reasoning step & 4.6\\
            \midrule
            Avg extra word for text hint & 17.4 \\
            Avg extra word for visual hint & 5.2 \\
            \bottomrule
            \end{tabular}
        }
    }
    \hfill
    \subcaptionbox{Category Distribution.\label{fig:overview:b}}[0.295\textwidth]{
        \centering
        \includegraphics[width=\linewidth]{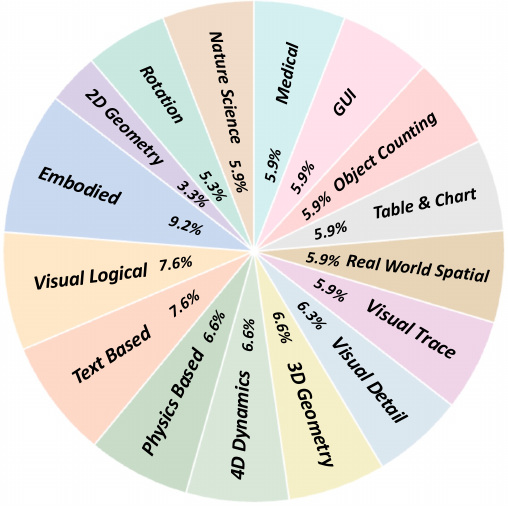}
    }
    \hfill
    \subcaptionbox{Evaluation  Map.\label{fig:overview:c}}[0.365\textwidth]{
        \centering
        \includegraphics[width=\linewidth, clip, trim=0 0 0 0]{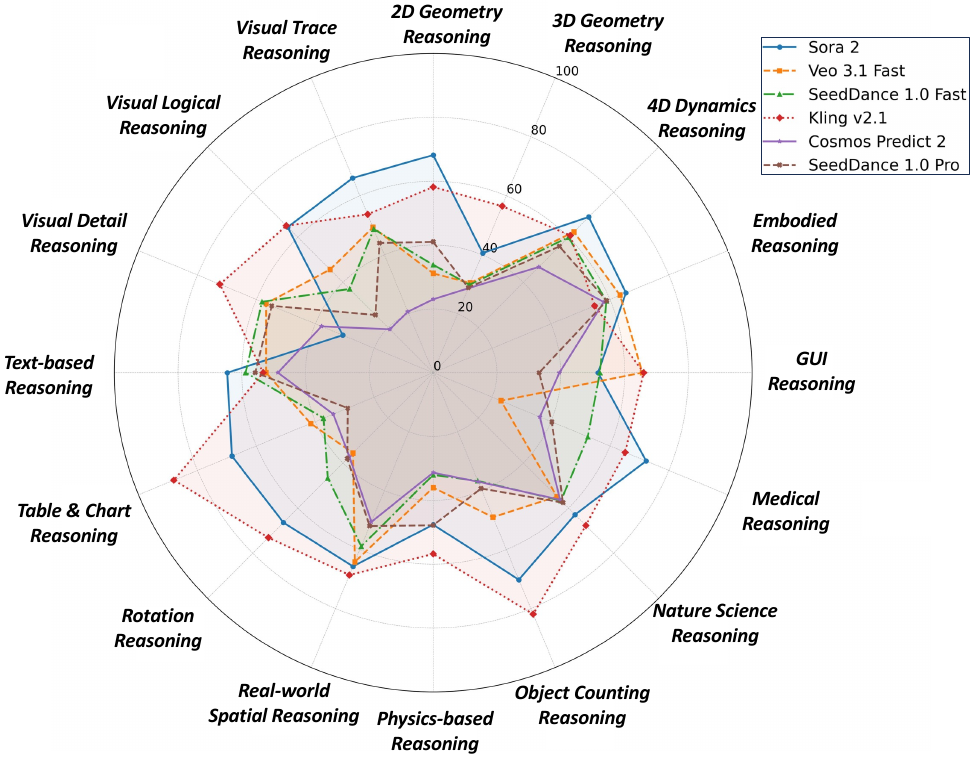}
    }
\end{minipage}
\caption{{Statistics, Distribution and Evaluation Results of MME-CoF-Pro.}}
\label{fig:overview_statistics}
\end{figure*}

\subsection{Overview of MME-CoF-Pro Benchmark.}
In order to study reasoning behaviors of video generative models under 
controlled hint conditions, we propose \textit{\textbf{MME-CoF-Pro}}, the first benchmark 
that treats reasoning guidance as an explicitly controllable variable and 
incorporates \textbf{\textit{Reasoning Score}} as process-level evaluation metric for interpretable reasoning coherence diagnosis, 
as shown in Fig.~\ref{fig:mme_cof_pro_text_hint},
~\ref{fig:mme_cof_pro_visual_hints}, and~\ref{fig:overview_statistics}.
\textit{MME-CoF-Pro} contains 303 samples across 16 reasoning categories, spanning a 
broad range of reasoning capabilities, 
from perceptual tasks such as visual detail reasoning, to physically grounded and causal 
tasks including 4D Dynamics Reasoning. 
All samples include no-hint and text-hint settings. For a subset of 8 perceptually demanding categories, designated \textit{MME-CoF-Pro-mini}, we additionally introduce a visual-hint setting (see Fig.~\ref{fig:mme_cof_pro_visual_hints}).
\textbf{(2)}~We further annotate each sample with key reasoning steps as intermediate checkpoints (Fig.~\ref{fig:evaluation_metric}) and provide details in Section~\ref{section_reasoning_step}, allowing fine-grained diagnosis of reasoning trajectories. These annotations define the evaluation criteria for the Reasoning Score (RS).

\subsubsection{Evaluation settings with no hint, text hint, and visual hint} 

As illustrated in Fig.~\ref{fig:mme_cof_pro_text_hint} and~\ref{fig:mme_cof_pro_visual_hints}, \textit{\textbf{MME-CoF-Pro}} evaluates each sample 
under controlled settings that differ only in the type of reasoning guidance 
provided. The \textbf{(a) no-hint} setting serves as the standard baseline, 
where models must reason purely from the task instruction without any 
additional scaffolding. The \textbf{(b) text-hint} setting augments the prompt 
with explicit textual descriptions of key reasoning steps, providing 
structured guidance to facilitate inference. 
For the 8 visually demanding categories in \textbf{\textit{MME-CoF-Pro-mini}}, we provide an additional \textbf{(c) visual-hint} setting (Fig.~\ref{fig:mme_cof_pro_visual_hints}) in which bounding boxes or directional arrows are drawn on the input image to direct attention to relevant regions or motion directions.
Since every part of the instruction except the hint remains identical across conditions, performance differences isolate the effect of reasoning guidance, supporting controlled and causal analysis of how models reason under different guidance types.
Each hint guidance also shares the same reasoning score rubric for fair comparison.

\subsubsection{Reasoning Score (RS) over process-level key step.}
\label{section_reasoning_step}

\begin{figure}[t]
\centering
\includegraphics[width=\linewidth]{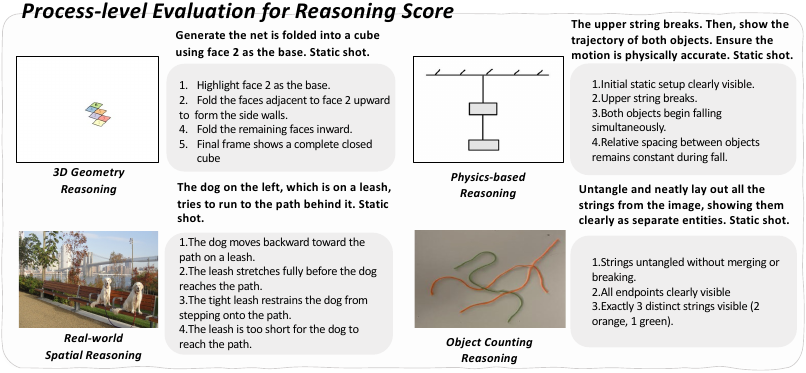}
\caption{\textbf{Process-level Evaluation Metric for Reasoning Score in \textit{MME-CoF-Pro.}} Videos are evaluated against fine-grained reasoning steps, enabling interpretable assessment beyond final-answer accuracy.
}
\label{fig:evaluation_metric}
\vspace{-3mm}
\end{figure}

\textit{MME-CoF-Pro} introduces a process-level evaluation framework that assesses reasoning coherence for each intermediate step which is necessary to accomplish a task. 
As illustrated in Fig.~\ref{fig:evaluation_metric}, each question is annotated as shown in Section~\ref{data_annotation_pipeline_section} with a sequence of key reasoning steps, where each step represents a critical checkpoint for a correct video generation. Formally, given a generated video $V$ and $N$ annotated reasoning steps $\{s_1, s_2, \ldots, s_N\}$, the Reasoning Score is computed as:
\begin{equation}
    \text{RS} = \frac{1}{N} \sum_{i=1}^{N} \mathbb{1}[s_i \text{ is correctly completed}]
\end{equation}
where each step $s_i$ is evaluated independently by a judge model, such as Gemini-2.5-Flash~\cite{comanici2025gemini}. 
This design allows fine-grained diagnosis of where along the reasoning chain a video generative model succeeds or fails, and supports more reliable cross-model comparison by capturing partial correctness rather than discarding it.

\vspace{-3mm}
\subsection{Category Distribution}

The 16 categories in MME-CoF-Pro are organized into four groups, progressing 
from low-level perceptual understanding to high-level task-oriented reasoning,
ensuring comprehensive coverage of the reasoning capabilities required for 
video generation.

\noindent\textbf{Perceptual Reasoning} covers fine-grained visual understanding, 
including \textit{Visual Detail Reasoning} (identifying attributes by zooming 
into specific regions), \textit{Rotation Reasoning} (interpreting scenes at 
non-standard orientations and restoring correct viewpoints), and \textit{Object 
Counting Reasoning} (accurately enumerating objects of specific types within 
cluttered scenes). 
\newline\noindent\textbf{Spatial and Structural Reasoning} requires understanding of 
geometric and spatial relationships, spanning \textit{Visual Trace Reasoning} 
(animating step-by-step navigation on mazes while avoiding obstacles), 
\textit{Real World Spatial Reasoning} (inferring egocentric and allocentric 
spatial relationships in real-world environments), and \textit{2D}/\textit{3D 
Geometry Reasoning} (constructing auxiliary lines over planar figures and 
reasoning about 3D object structure across viewpoints). 
\newline\noindent\textbf{Physical and Causal Reasoning} evaluates grounded understanding 
of how the world behaves over time, covering \textit{Physics-Based Reasoning} 
(animating motion governed by physical laws such as gravity and collision), 
\textit{4D Dynamics Reasoning} (predicting object interactions over time such 
as occlusion, collision, and containment), and \textit{Nature Science Domain} 
(reasoning grounded in domain knowledge across chemistry, biology, and optics). 
\newline\noindent\textbf{Task-oriented Reasoning} covers reasoning embedded in 
real-world application contexts, including \textit{Embodied Reasoning} 
(generating correct robotic manipulation trajectories), \textit{GUI Reasoning} 
(identifying and interacting with interface elements in desktop or mobile 
screenshots), \textit{Medical Reasoning} (analyzing clinical images such as 
CT scans and X-rays), \textit{Table and Chart Reasoning} (locating specific 
data entries via zoom or bounding box operations), \textit{Text-Based Reasoning} 
(mathematical problem solving and code execution tracing), and \textit{Visual 
Logical Reasoning} (pattern completion and rule inference over structured 
visual grids and board games).

\subsection{Benchmark Curation}

\subsubsection{Data Source.}

\textit{MME-CoF-Pro} is curated from 27 existing real-world and synthetic benchmarks, including  where each of the 16 reasoning categories is constructed by selectively sampling datasets that best represent the target reasoning skill (see Appendix for details). Real-world datasets provide ecological validity and grounded visual understanding, while synthetic benchmarks enable precise geometric, spatial, and procedural reasoning with controllable layouts and clear annotations. We respect all original licenses and usage terms and use only publicly available data. We provide more details in Appendix.

\subsubsection{Data Annotation Pipeline.}
\label{data_annotation_pipeline_section}

To ensure annotation reliability, each sample was annotated and reviewed by multiple human annotators. The reasoning steps were first drafted by Gemini-2.5-Pro~\cite{comanici2025gemini} and subsequently refined through three rounds of manual verification and correction by domain experts.  Images with visual hints are manually annotated by human annotators using Photopea, strictly preserving the original image resolution. During this process, annotators cross-checked each reasoning step to ensure that it was (1) necessary for the intended reasoning trajectory and (2) unambiguously verifiable from the generated video. Disagreements were resolved through discussion until consensus was reached. The annotated reasoning steps represent minimal necessary checkpoints required for successful task completion, which reduces ambiguity 
while preserving evaluation consistency.



\section{Experiment and Analysis}

After detailing the experimental setup including models, metrics, hint configurations, and human study design in Section~\ref{experiment_set_up_section}, we validate the proposed Reasoning Score via human evaluation in Section~\ref{human_study_section}. Section~\ref{main_result_reasoning_score_evaluation} assesses how well current video generative models reason and how reasoning relates to generation quality, while Sections~\ref{main_result_text_hints_evaluation} and~\ref{main_result_visual_hints_evaluation} explore whether reasoning hints enhance reasoning and how hint modality shapes the outcome.

\vspace{-2mm}

\subsection{Experiment Setup.}
\label{experiment_set_up_section}

\paragraph{Models.} We evaluated 7 closed and open-source models, Veo-3.1~\cite{wiedemer2025video}, Veo-3.1-fast~\cite{wiedemer2025video}, Sora-2~\cite{peng2025open}, Seedance-1.0-pro~\cite{gao2025seedance}, Seedance-1.0-fast~\cite{gao2025seedance}, Kling-v2.1~\cite{team2025klingavatar} and Cosmos-Predict2-14B~\cite{agarwal2025cosmos}. We use default configurations of these video models, with further inference configurations provided in Appendix. 

\paragraph{Evaluation Protocols.} Following recent video reasoning benchmarks, we adopt an VLM-as-a-judge protocol for automated evaluation, employing Gemini-2.5-Flash~\cite{comanici2025gemini} as our judge model. 
During evaluation, each generated video is uniformly sampled at every 10th frame. 
Videos are 4 or 5 seconds long depending on the generative model, all at 24 fps. The reliability of judge model is validated via human study in Section ~\ref{human_study_section}, where RS achieves the highest Spearman correlation (0.61) with human scores.

\paragraph{Evaluation Metrics.} We evaluate models from two perspectives: \textit{Reasoning Score (RS)} and \textit{Generation Quality (GQ)}. \textbf{(1)} For RS, Gemini uses the annotated process-level reasoning steps, as shown in Fig.~\ref{fig:evaluation_metric}, to identify whether each reasoning step is successfully achieved in the output video. \textbf{Reasoning Score is reported as percentages}, higher is better.
\textbf{(2)} GQ is evaluated along five dimensions: Consistency Score (CS), temporal consistency, visual stability, hallucination, and physics grounding. 
CS measures cross-frame content coherence and the absence of abrupt content changes. 
Temporal consistency evaluates motion smoothness across frames. 
Visual stability assesses the steadiness of camera, objects, and scene composition. 
Hallucination penalizes unexpected or spurious visual artifacts. Physics grounding measures whether motions and interactions are physically plausible. We report the percentage value of CS and the average (Avg) across all five dimensions representing Generation Quality, and higher is better.

\paragraph{Evaluation Baselines.} We evaluate three settings: \textit{no hint}, \textit{text hint}, and \textit{visual hint}. 
The \textit{no hint} and \textit{text hint} settings are tested on the full 
\textit{MME-CoF-Pro} (16 categories), where \textit{no hint} serves as the baseline to 
measure gains from text hints. The \textit{visual hint} setting is evaluated on 
\textit{MME-CoF-Pro-mini}, a curated subset covering 8 perceptually intensive categories, 
with its corresponding \textit{no hint} results as the baseline.

\paragraph{Human Study Setup.}
\label{human_study_setup}
To validate that RS serves as an effective and relatively independent measure of video reasoning capability, we conduct a human study on 10 randomly selected generated videos with 10 human participants, each of whom scores the videos according to the annotated reasoning steps. Furthermore, we compare RS with two alternative metrics adopted from prior work: \textit{Instruction Alignment}~\cite{guo2025video} and \textit{Pass@5 last-frame correctness}~\cite{tong2025thinking}. The Spearsman correlation is computed between the average of human scores and each metric to assess their respective reliability in capturing reasoning performance.

\subsection{Result Analysis of Reasoning Score}
\label{main_result_reasoning_score_evaluation}

\paragraph{Overall models fall short in reasoning coherence.}

In Tab.~\ref{tab:main_result}, we report the RS under no-hint and text-hint settings. Model-level averages show clear stratification: frontier proprietary models lead, with Veo achieving the highest RS (56), followed by Sora (50), while Seedance, Cosmos, and Kling lag behind. Notably, even the best model only slightly exceeds 50, indicating that reasoning in video generation remains challenging.

\clearpage
\begin{table*}[ht]
\centering
\setlength\tabcolsep{6.3pt}
\footnotesize
\caption{
\textbf{Evaluation results of MME-CoF-Pro.} We report the percentage of \textit{Reasoning Score (RS)} and \textit{Consistency Score (CS)} in \textbf{no-hint settings}, with text-hint performance shown in parentheses. Color indicates the value of change brought by \textbf{text-hint}: \colorbox{green!12}{$\uparrow$~improvement} and \colorbox{red!12}{$\downarrow$~degradation}. We additionally evaluate \textit{Hallucination Score}, \textit{Visual Stability}, and \textit{Physics Grounding}, and present the overall average score with \textit{Consistency Score} here. Details are provided in the Appendix.
}
\vspace{-3mm}
\label{tab:main_result}
\resizebox{0.96\textwidth}{!}{%
\begin{tabular}{cc|cccccc|c}
\toprule
\multicolumn{1}{c|}{\multirow{2}{*}{\textbf{Category}}} & \multirow{2}{*}{\textbf{Metric}} & \multicolumn{6}{c|}{\textbf{Closed-Source Models}} & \multicolumn{1}{c}{\textbf{Open-Source Models}} \\ \cmidrule(l){3-9}
\multicolumn{1}{c|}{} & & Veo3.1~\cite{wiedemer2025video} & Veo3.1-fast~\cite{wiedemer2025video} & Sora2~\cite{peng2025open} & \begin{tabular}[c]{@{}c@{}}Seedance\\ 1.0-pro\end{tabular}~\cite{gao2025seedance} & \begin{tabular}[c]{@{}c@{}}Seedance\\ 1.0-fast\end{tabular}~\cite{gao2025seedance} & Kling-v2.1~\cite{team2025klingavatar} & \begin{tabular}[c]{@{}c@{}}Cosmos\\ Predict\end{tabular}~\cite{agarwal2025cosmos} \\ \midrule

\multicolumn{1}{c|}{\multirow{3}*{\begin{tabular}[c]{@{}c@{}}Visual Detail\\ Reasoning\end{tabular}}} & RS   & 61.5(\colorbox{green!12}{+18.6}) & 65.7(\colorbox{green!12}{+6.0}) & 40.3(\colorbox{green!12}{+27.8}) & 19.8(\colorbox{green!12}{+29.3}) & 26.1(\colorbox{green!12}{+11.7}) & 2.0(\colorbox{red!12}{-2.0}) & 30.2(\colorbox{green!12}{+8.1}) \\[1pt]
\multicolumn{1}{c|}{}                                 & CS   & 51.2(0.0) & 47.6(\colorbox{red!12}{-1.8}) & 22.8(\colorbox{green!12}{+13.3}) & 50.0(\colorbox{green!12}{+7.6}) & 55.6(\colorbox{red!12}{-13.3}) & 64.7(\colorbox{red!12}{-16.6}) & 36.1(\colorbox{red!12}{-10.6}) \\[1pt]
\multicolumn{1}{c|}{}                                 & \textbf{Avg} & 56.8(\colorbox{red!12}{-1.1}) & 56.7(\colorbox{red!12}{-3.2}) & 30.7(\colorbox{green!12}{+13.7}) & 54.8(\colorbox{green!12}{+6.6}) & 58.1(\colorbox{red!12}{-5.3}) & 72.3(\colorbox{red!12}{-7.3}) & 37.9(\colorbox{red!12}{-8.6}) \\[4pt] \midrule \addlinespace[4pt]

\multicolumn{1}{c|}{\multirow{3}*{\begin{tabular}[c]{@{}c@{}}Embodied\\ Reasoning\end{tabular}}} & RS   & 81.7(\colorbox{red!12}{-0.2}) & 78.5(\colorbox{green!12}{+0.8}) & 38.4(\colorbox{red!12}{-3.9}) & 48.7(\colorbox{green!12}{+6.8}) & 57.0(\colorbox{red!12}{-6.4}) & 2.5(\colorbox{red!12}{-1.4}) & 49.4(\colorbox{green!12}{+20.1}) \\[1pt]
\multicolumn{1}{c|}{}                                 & CS   & 72.9(\colorbox{green!12}{+1.1}) & 62.5(\colorbox{green!12}{+8.2}) & 61.8(\colorbox{red!12}{-1.1}) & 56.9(\colorbox{red!12}{-3.3}) & 52.8(\colorbox{green!12}{+0.1}) & 44.1(\colorbox{red!12}{-3.8}) & 58.6(\colorbox{red!12}{-6.6}) \\[1pt]
\multicolumn{1}{c|}{}                                 & \textbf{Avg} & 76.1(\colorbox{green!12}{+1.6}) & 63.4(\colorbox{green!12}{+9.9}) & 65.4(\colorbox{red!12}{-4.3}) & 58.9(\colorbox{red!12}{-3.6}) & 58.8(\colorbox{red!12}{-2.2}) & 54.8(\colorbox{red!12}{-2.4}) & 57.9(\colorbox{red!12}{-4.4}) \\[4pt] \midrule \addlinespace[4pt]

\multicolumn{1}{c|}{\multirow{3}*{\begin{tabular}[c]{@{}c@{}}Object Counting\\ Reasoning\end{tabular}}} & RS   & 51.7(\colorbox{green!12}{+5.8}) & 55.8(\colorbox{red!12}{-2.1}) & 57.1(\colorbox{green!12}{+5.1}) & 35.4(\colorbox{green!12}{+14.2}) & 31.2(\colorbox{green!12}{+4.6}) & 11.8(\colorbox{red!12}{-3.8}) & 16.6(\colorbox{red!12}{-5.3}) \\[1pt]
\multicolumn{1}{c|}{}                                 & CS   & 47.2(\colorbox{red!12}{-0.6}) & 46.7(\colorbox{green!12}{+8.3}) & 69.4(\colorbox{red!12}{-10.6}) & 36.5(\colorbox{red!12}{-1.8}) & 30.0(\colorbox{green!12}{+3.9}) & 75.3(\colorbox{red!12}{-6.5}) & 36.1(\colorbox{green!12}{+3.3}) \\[1pt]
\multicolumn{1}{c|}{}                                 & \textbf{Avg} & 46.1(\colorbox{green!12}{+3.4}) & 49.0(\colorbox{green!12}{+7.1}) & 70.2(\colorbox{red!12}{-2.9}) & 39.3(\colorbox{red!12}{-2.2}) & 36.7(\colorbox{green!12}{+3.2}) & 81.9(\colorbox{red!12}{-4.0}) & 37.0(\colorbox{green!12}{+11.3}) \\[4pt] \midrule \addlinespace[4pt]

\multicolumn{1}{c|}{\multirow{3}*{\begin{tabular}[c]{@{}c@{}}Table and Chart\\ Reasoning\end{tabular}}} & RS   & 49.4(\colorbox{green!12}{+4.4}) & 52.3(\colorbox{red!12}{-5.4}) & 61.3(\colorbox{green!12}{+0.2}) & 33.7(\colorbox{red!12}{-11.2}) & 29.9(\colorbox{green!12}{+15.3}) & 33.5(\colorbox{red!12}{-4.8}) & 23.8(\colorbox{green!12}{+10.5}) \\[1pt]
\multicolumn{1}{c|}{}                                 & CS   & 31.1(\colorbox{green!12}{+3.3}) & 38.3(\colorbox{red!12}{-11.1}) & 72.2(\colorbox{red!12}{-2.8}) & 25.7(\colorbox{red!12}{-7.9}) & 33.3(\colorbox{green!12}{+2.8}) & 85.7(\colorbox{red!12}{-2.1}) & 33.9(\colorbox{green!12}{+12.2}) \\[1pt]
\multicolumn{1}{c|}{}                                 & \textbf{Avg} & 33.6(\colorbox{green!12}{+3.3}) & 41.6(\colorbox{red!12}{-8.4}) & 68.3(\colorbox{red!12}{-2.8}) & 29.0(\colorbox{red!12}{-9.3}) & 37.2(\colorbox{green!12}{+1.8}) & 88.0(\colorbox{red!12}{-5.6}) & 33.9(\colorbox{green!12}{+5.2}) \\[4pt] \midrule \addlinespace[4pt]

\multicolumn{1}{c|}{\multirow{3}*{\begin{tabular}[c]{@{}c@{}}2D Geometry\\ Reasoning\end{tabular}}} & RS   & 29.2(\colorbox{red!12}{-6.9}) & 22.2(\colorbox{red!12}{-4.6}) & 54.2(\colorbox{red!12}{-19.2}) & 22.5(\colorbox{red!12}{-9.2}) & 9.2(\colorbox{green!12}{+25.8}) & 0.0(\colorbox{green!12}{+3.3}) & 9.2(\colorbox{red!12}{-5.8}) \\[1pt]
\multicolumn{1}{c|}{}                                 & CS   & 28.0(\colorbox{red!12}{-4.7}) & 27.8(\colorbox{green!12}{+14.4}) & 64.0(\colorbox{green!12}{+2.0}) & 32.0(\colorbox{green!12}{+12.0}) & 22.0(\colorbox{green!12}{+14.0}) & 46.0(\colorbox{green!12}{+10.0}) & 15.0(\colorbox{green!12}{+9.0}) \\[1pt]
\multicolumn{1}{c|}{}                                 & \textbf{Avg} & 32.4(\colorbox{green!12}{+0.5}) & 31.1(\colorbox{green!12}{+12.2}) & 68.2(\colorbox{green!12}{+2.6}) & 41.0(\colorbox{green!12}{+11.8}) & 33.8(\colorbox{green!12}{+3.8}) & 58.2(\colorbox{green!12}{+12.4}) & 23.0(\colorbox{green!12}{+4.4}) \\[4pt] \midrule \addlinespace[4pt]

\multicolumn{1}{c|}{\multirow{3}*{\begin{tabular}[c]{@{}c@{}}GUI\\ Reasoning\end{tabular}}} & RS   & 69.3(\colorbox{red!12}{-10.1}) & 72.0(\colorbox{red!12}{-4.6}) & 47.2(\colorbox{red!12}{-3.1}) & 15.6(\colorbox{red!12}{-0.4}) & 26.7(\colorbox{green!12}{+1.1}) & 7.5(\colorbox{red!12}{-1.8}) & 18.8(\colorbox{green!12}{+5.7}) \\[1pt]
\multicolumn{1}{c|}{}                                 & CS   & 68.9(\colorbox{red!12}{-6.1}) & 66.7(\colorbox{red!12}{-2.2}) & 57.5(\colorbox{green!12}{+8.8}) & 33.9(\colorbox{green!12}{+2.8}) & 53.9(\colorbox{red!12}{-2.2}) & 77.2(\colorbox{red!12}{-15.6}) & 39.4(\colorbox{green!12}{+17.2}) \\[1pt]
\multicolumn{1}{c|}{}                                 & \textbf{Avg} & 69.7(\colorbox{red!12}{-7.1}) & 65.6(\colorbox{green!12}{+0.2}) & 51.8(\colorbox{green!12}{+14.1}) & 33.2(\colorbox{green!12}{+2.2}) & 52.3(\colorbox{red!12}{-2.1}) & 66.0(\colorbox{red!12}{-1.9}) & 39.6(\colorbox{green!12}{+10.2}) \\[4pt] \midrule \addlinespace[4pt]

\multicolumn{1}{c|}{\multirow{3}*{\begin{tabular}[c]{@{}c@{}}Physics-based\\ Reasoning\end{tabular}}} & RS   & 55.9(\colorbox{green!12}{+6.6}) & 65.8(\colorbox{red!12}{-20.1}) & 62.2(\colorbox{green!12}{+9.5}) & 53.2(\colorbox{red!12}{-4.2}) & 45.5(\colorbox{red!12}{-5.2}) & 18.9(\colorbox{red!12}{-4.5}) & 25.9(\colorbox{green!12}{+7.9}) \\[1pt]
\multicolumn{1}{c|}{}                                 & CS   & 30.5(\colorbox{red!12}{-2.0}) & 36.5(\colorbox{red!12}{-2.5}) & 51.0(\colorbox{green!12}{+6.5}) & 48.5(\colorbox{red!12}{-12.3}) & 34.5(\colorbox{green!12}{+4.0}) & 45.4(\colorbox{green!12}{+20.0}) & 30.5(\colorbox{red!12}{-2.8}) \\[1pt]
\multicolumn{1}{c|}{}                                 & \textbf{Avg} & 27.1(\colorbox{green!12}{+2.5}) & 36.0(\colorbox{red!12}{-4.4}) & 47.6(\colorbox{green!12}{+7.6}) & 47.9(\colorbox{red!12}{-7.9}) & 32.0(\colorbox{green!12}{+5.0}) & 56.8(\colorbox{green!12}{+6.3}) & 31.3(\colorbox{red!12}{-1.3}) \\[4pt] \midrule \addlinespace[4pt]

\multicolumn{1}{c|}{\multirow{3}*{\begin{tabular}[c]{@{}c@{}}Text-based\\ Reasoning\end{tabular}}} & RS   & 62.3(\colorbox{green!12}{+8.4}) & 62.0(\colorbox{green!12}{+4.6}) & 75.4(\colorbox{green!12}{+1.8}) & 35.6(\colorbox{red!12}{-6.8}) & 44.9(\colorbox{red!12}{-1.7}) & 20.7(\colorbox{red!12}{-5.0}) & 29.1(\colorbox{red!12}{-2.0}) \\[1pt]
\multicolumn{1}{c|}{}                                 & CS   & 49.6(\colorbox{green!12}{+6.9}) & 49.6(\colorbox{green!12}{+4.9}) & 55.6(\colorbox{red!12}{-2.2}) & 49.6(\colorbox{red!12}{-2.2}) & 54.8(\colorbox{red!12}{-8.7}) & 34.8(\colorbox{green!12}{+10.4}) & 43.5(\colorbox{red!12}{-1.7}) \\[1pt]
\multicolumn{1}{c|}{}                                 & \textbf{Avg} & 52.5(\colorbox{green!12}{+9.1}) & 52.4(\colorbox{green!12}{+9.9}) & 64.6(\colorbox{red!12}{-5.0}) & 55.8(\colorbox{red!12}{-3.7}) & 58.8(\colorbox{red!12}{-8.1}) & 53.3(\colorbox{red!12}{-0.8}) & 48.7(\colorbox{red!12}{-3.3}) \\[4pt] \midrule \addlinespace[4pt]

\multicolumn{1}{c|}{\multirow{3}*{\begin{tabular}[c]{@{}c@{}}3D Geometry\\ Reasoning\end{tabular}}} & RS   & 43.9(\colorbox{green!12}{+13.2}) & 41.7(\colorbox{green!12}{+7.2}) & 37.7(\colorbox{green!12}{+14.6}) & 36.4(\colorbox{red!12}{-7.8}) & 48.5(\colorbox{red!12}{-23.8}) & 13.4(\colorbox{green!12}{+0.5}) & 22.1(\colorbox{green!12}{+3.1}) \\[1pt]
\multicolumn{1}{c|}{}                                 & CS   & 27.2(\colorbox{green!12}{+4.4}) & 27.2(\colorbox{green!12}{+1.7}) & 42.8(\colorbox{red!12}{-8.3}) & 28.2(\colorbox{red!12}{-7.1}) & 22.8(\colorbox{red!12}{-3.9}) & 50.6(\colorbox{green!12}{+9.4}) & 26.7(\colorbox{green!12}{+5.6}) \\[1pt]
\multicolumn{1}{c|}{}                                 & \textbf{Avg} & 32.2(\colorbox{green!12}{+0.1}) & 30.4(\colorbox{red!12}{-2.8}) & 40.6(\colorbox{red!12}{-2.9}) & 28.8(\colorbox{red!12}{-4.4}) & 29.8(\colorbox{red!12}{-5.1}) & 56.5(\colorbox{green!12}{+4.5}) & 28.6(\colorbox{green!12}{+9.2}) \\[4pt] \midrule \addlinespace[4pt]

\multicolumn{1}{c|}{\multirow{3}*{\begin{tabular}[c]{@{}c@{}}Medical\\ Reasoning\end{tabular}}} & RS   & 41.7(\colorbox{red!12}{-0.2}) & 35.8(\colorbox{green!12}{+3.2}) & 49.6(\colorbox{green!12}{+10.0}) & 32.4(\colorbox{green!12}{+10.1}) & 45.3(\colorbox{red!12}{-6.4}) & 38.7(\colorbox{red!12}{-2.5}) & 31.3(\colorbox{green!12}{+13.0}) \\[1pt]
\multicolumn{1}{c|}{}                                 & CS   & 22.8(\colorbox{red!12}{-1.1}) & 21.7(\colorbox{red!12}{-3.3}) & 67.8(\colorbox{red!12}{-7.2}) & 40.0(0.0) & 51.2(\colorbox{red!12}{-0.6}) & 54.4(\colorbox{green!12}{+21.7}) & 37.8(\colorbox{green!12}{+2.2}) \\[1pt]
\multicolumn{1}{c|}{}                                 & \textbf{Avg} & 27.1(\colorbox{red!12}{-2.6}) & 23.0(\colorbox{red!12}{-2.2}) & 72.3(\colorbox{red!12}{-10.4}) & 40.3(\colorbox{green!12}{+1.0}) & 52.5(\colorbox{green!12}{+3.1}) & 65.2(\colorbox{green!12}{+6.0}) & 36.2(\colorbox{green!12}{+6.2}) \\[4pt] \midrule \addlinespace[4pt]

\multicolumn{1}{c|}{\multirow{3}*{\begin{tabular}[c]{@{}c@{}}Real-world Spatial\\ Reasoning\end{tabular}}} & RS   & 54.0(\colorbox{red!12}{-0.8}) & 58.7(\colorbox{green!12}{+9.6}) & 35.0(\colorbox{green!12}{+26.5}) & 30.6(\colorbox{red!12}{-0.4}) & 26.2(\colorbox{green!12}{+19.1}) & 9.4(\colorbox{red!12}{-2.3}) & 41.4(\colorbox{green!12}{+11.7}) \\[1pt]
\multicolumn{1}{c|}{}                                 & CS   & 57.8(\colorbox{red!12}{-0.7}) & 63.3(\colorbox{green!12}{+1.4}) & 62.6(\colorbox{red!12}{-2.6}) & 53.2(\colorbox{red!12}{-7.0}) & 60.0(\colorbox{red!12}{-7.2}) & 58.4(\colorbox{green!12}{+12.7}) & 52.6(\colorbox{green!12}{+6.3}) \\[1pt]
\multicolumn{1}{c|}{}                                 & \textbf{Avg} & 62.1(\colorbox{red!12}{-0.4}) & 64.2(\colorbox{green!12}{+1.8}) & 65.7(\colorbox{red!12}{-0.3}) & 52.0(0.0) & 59.0(\colorbox{red!12}{-6.7}) & 68.6(\colorbox{green!12}{+6.8}) & 50.6(\colorbox{green!12}{+6.5}) \\[4pt] \midrule \addlinespace[4pt]

\multicolumn{1}{c|}{\multirow{3}*{\begin{tabular}[c]{@{}c@{}}Visual Logical\\ Reasoning\end{tabular}}} & RS   & 72.0(\colorbox{red!12}{-7.4}) & 58.2(\colorbox{green!12}{+8.9}) & 49.4(\colorbox{green!12}{+12.3}) & 22.4(\colorbox{green!12}{+11.2}) & 36.5(\colorbox{green!12}{+3.2}) & 9.8(\colorbox{green!12}{+4.6}) & 4.3(\colorbox{green!12}{+49.3}) \\[1pt]
\multicolumn{1}{c|}{}                                 & CS   & 55.5(\colorbox{red!12}{-5.5}) & 44.5(\colorbox{red!12}{-1.9}) & 64.5(\colorbox{red!12}{-7.6}) & 20.6(\colorbox{green!12}{+2.6}) & 32.6(\colorbox{red!12}{-6.5}) & 48.7(\colorbox{red!12}{-1.7}) & 15.0(\colorbox{green!12}{+14.6}) \\[1pt]
\multicolumn{1}{c|}{}                                 & \textbf{Avg} & 55.5(\colorbox{red!12}{-5.0}) & 45.7(\colorbox{green!12}{+0.3}) & 64.5(\colorbox{red!12}{-6.2}) & 25.6(\colorbox{green!12}{+3.6}) & 37.1(\colorbox{red!12}{-8.4}) & 65.1(0.0) & 19.4(\colorbox{green!12}{+10.4}) \\[4pt] \midrule \addlinespace[4pt]

\multicolumn{1}{c|}{\multirow{3}*{\begin{tabular}[c]{@{}c@{}}4D Dynamics\\ Reasoning\end{tabular}}} & RS   & 66.2(\colorbox{green!12}{+12.4}) & 64.7(\colorbox{green!12}{+11.3}) & 51.3(\colorbox{green!12}{+6.9}) & 61.2(\colorbox{green!12}{+2.7}) & 52.0(\colorbox{green!12}{+9.1}) & 9.7(\colorbox{red!12}{-0.9}) & 52.2(\colorbox{red!12}{-3.4}) \\[1pt]
\multicolumn{1}{c|}{}                                 & CS   & 68.2(\colorbox{red!12}{-1.2}) & 62.8(\colorbox{red!12}{-7.2}) & 73.3(\colorbox{red!12}{-13.9}) & 56.1(\colorbox{red!12}{-11.1}) & 59.4(\colorbox{red!12}{-2.2}) & 65.0(\colorbox{red!12}{-15.6}) & 55.6(\colorbox{red!12}{-11.1}) \\[1pt]
\multicolumn{1}{c|}{}                                 & \textbf{Avg} & 65.4(\colorbox{green!12}{+0.1}) & 62.4(\colorbox{red!12}{-1.7}) & 69.0(\colorbox{red!12}{-11.0}) & 56.0(\colorbox{red!12}{-8.6}) & 60.0(\colorbox{red!12}{-3.7}) & 60.8(\colorbox{red!12}{-6.6}) & 46.8(\colorbox{red!12}{-8.6}) \\[4pt] \midrule \addlinespace[4pt]

\multicolumn{1}{c|}{\multirow{3}*{\begin{tabular}[c]{@{}c@{}}Natural Science\\ Reasoning\end{tabular}}} & RS   & 49.6(\colorbox{green!12}{+1.8}) & 55.3(\colorbox{green!12}{+17.0}) & 60.3(\colorbox{red!12}{-1.1}) & 46.0(\colorbox{red!12}{-5.2}) & 49.6(\colorbox{green!12}{+1.8}) & 11.3(\colorbox{green!12}{+2.8}) & 51.0(\colorbox{green!12}{+2.8}) \\[1pt]
\multicolumn{1}{c|}{}                                 & CS   & 57.3(\colorbox{red!12}{-5.3}) & 51.1(\colorbox{green!12}{+11.7}) & 66.7(\colorbox{red!12}{-4.4}) & 53.8(\colorbox{green!12}{+1.0}) & 57.3(\colorbox{red!12}{-5.3}) & 62.8(\colorbox{red!12}{-26.1}) & 57.5(\colorbox{red!12}{-20.8}) \\[1pt]
\multicolumn{1}{c|}{}                                 & \textbf{Avg} & 56.7(\colorbox{green!12}{+0.7}) & 54.9(\colorbox{green!12}{+9.0}) & 62.9(\colorbox{red!12}{-0.3}) & 57.6(\colorbox{green!12}{+0.6}) & 56.7(\colorbox{green!12}{+0.7}) & 67.7(\colorbox{red!12}{-14.9}) & 56.3(\colorbox{red!12}{-15.3}) \\[4pt] \midrule \addlinespace[4pt]

\multicolumn{1}{c|}{\multirow{3}*{\begin{tabular}[c]{@{}c@{}}Rotation\\ Reasoning\end{tabular}}} & RS   & 60.2(\colorbox{red!12}{-2.5}) & 56.0(\colorbox{green!12}{+4.6}) & 30.1(\colorbox{green!12}{+9.8}) & 41.8(\colorbox{green!12}{+5.0}) & 35.8(\colorbox{red!12}{-10.4}) & 11.9(\colorbox{green!12}{+11.8}) & 25.0(\colorbox{green!12}{+6.0}) \\[1pt]
\multicolumn{1}{c|}{}                                 & CS   & 44.4(\colorbox{green!12}{+2.3}) & 34.4(\colorbox{green!12}{+13.8}) & 71.2(\colorbox{red!12}{-6.2}) & 40.0(\colorbox{green!12}{+9.3}) & 47.5(\colorbox{red!12}{-11.2}) & 79.3(\colorbox{red!12}{-5.3}) & 41.2(\colorbox{red!12}{-4.1}) \\[1pt]
\multicolumn{1}{c|}{}                                 & \textbf{Avg} & 45.8(\colorbox{green!12}{+2.1}) & 35.6(\colorbox{green!12}{+9.6}) & 66.4(\colorbox{red!12}{-0.6}) & 38.1(\colorbox{green!12}{+10.9}) & 46.8(\colorbox{red!12}{-12.5}) & 73.1(\colorbox{green!12}{+2.5}) & 37.2(\colorbox{red!12}{-5.5}) \\[4pt] \midrule \addlinespace[4pt]

\multicolumn{1}{c|}{\multirow{3}*{\begin{tabular}[c]{@{}c@{}}Visual Trace\\ Reasoning\end{tabular}}} & RS   & 45.5(\colorbox{green!12}{+24.0}) & 49.8(\colorbox{green!12}{+17.5}) & 49.7(\colorbox{green!12}{+21.9}) & 35.7(\colorbox{green!12}{+7.5}) & 36.0(\colorbox{green!12}{+12.2}) & 19.5(\colorbox{red!12}{-0.1}) & 27.9(\colorbox{green!12}{+4.1}) \\[1pt]
\multicolumn{1}{c|}{}                                 & CS   & 47.8(\colorbox{green!12}{+6.3}) & 43.3(\colorbox{green!12}{+4.9}) & 63.9(\colorbox{red!12}{-1.7}) & 38.9(\colorbox{green!12}{+0.6}) & 44.4(\colorbox{red!12}{-1.7}) & 40.0(\colorbox{green!12}{+8.9}) & 20.6(\colorbox{green!12}{+1.6}) \\[1pt]
\multicolumn{1}{c|}{}                                 & \textbf{Avg} & 53.0(\colorbox{green!12}{+5.7}) & 49.3(\colorbox{green!12}{+1.8}) & 66.0(\colorbox{green!12}{+2.3}) & 44.0(0.0) & 48.7(\colorbox{red!12}{-4.4}) & 53.7(\colorbox{green!12}{+0.4}) & 20.7(\colorbox{green!12}{+6.2}) \\[4pt] \midrule \addlinespace[4pt]

\multicolumn{1}{c|}{\multirow{3}*{Average}}         & RS   & 55.9(\colorbox{green!12}{+4.2}) & 55.9(\colorbox{green!12}{+3.4}) & 49.9(\colorbox{green!12}{+7.4}) & 35.7(\colorbox{green!12}{+2.6}) & 37.5(\colorbox{green!12}{+3.1}) & 13.8(\colorbox{red!12}{-0.4}) & 28.6(\colorbox{green!12}{+7.9}) \\[1pt]
\multicolumn{1}{c|}{}                                 & CS   & 47.5(\colorbox{red!12}{-0.2}) & 45.3(\colorbox{green!12}{+2.4}) & 60.5(\colorbox{red!12}{-2.4}) & 41.5(\colorbox{red!12}{-1.0}) & 44.5(\colorbox{red!12}{-2.4}) & 58.3(\colorbox{red!12}{-0.0}) & 37.5(\colorbox{green!12}{+0.9}) \\[1pt]
\multicolumn{1}{c|}{}                                 & \textbf{Avg} & 49.5(\colorbox{green!12}{+0.8}) & 47.6(\colorbox{green!12}{+2.4}) & 60.9(\colorbox{red!12}{-0.4}) & 43.9(\colorbox{red!12}{-0.2}) & 47.4(\colorbox{red!12}{-2.6}) & 65.1(\colorbox{red!12}{-0.3}) & 37.8(\colorbox{green!12}{+1.4}) \\[4pt] \midrule \addlinespace[4pt]

\end{tabular}%
}
\vspace{-3pt}
\end{table*}
\clearpage

\begin{figure}[t]
\centering
\includegraphics[width=\linewidth]{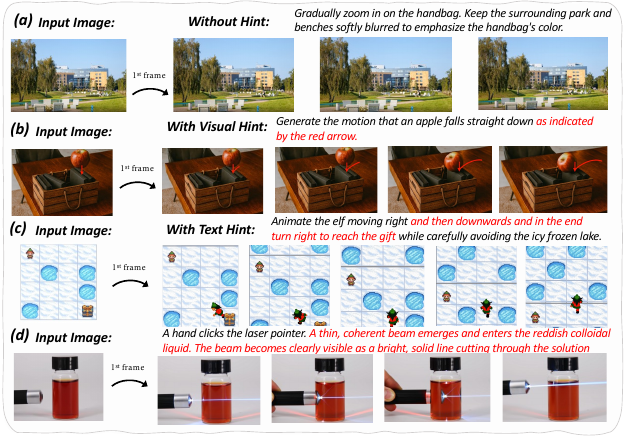}
\caption{\textbf{Failure Examples of Video Generative Models.} Subfigure (a) shows that the Kling model prioritizes perceptual fidelity (e.g., wind and forest dynamics) over the intended reasoning trajectory of gradual zoom-in. Subfigures (b) to (d) reveal hallucinations and inconsistencies arising under different hint conditions.
}
\label{fig:visual_text_hint_hallucination}
\vspace{-5mm}
\end{figure}

\paragraph{Decoupling between generation quality and reasoning score.}
We observe a clear mismatch between overall generation quality (Avg) and RS, indicating that strong visual quality does not necessarily reflect strong reasoning ability. Some models achieve competitive Avg scores while exhibiting low RS, suggesting they generate visually coherent videos without correctly executing reasoning steps. An illustrative case is Kling, which achieves a high Avg score (65.1) but a low RS (13.8), showing that strong visual fidelity can coexist with weak process-level reasoning. As shown in Fig.~\ref{fig:visual_text_hint_hallucination}, the Kling model prioritizes the fidelity of wind dynamics over faithfully following the zoom-in instruction and searching for the handbag. This decoupling suggests that reasoning is a distinct capability, underscoring the need for dedicated reasoning-centric evaluation.





\begin{finding}{Takeaway 1}
Video generative models exhibit generally weak reasoning ability, and their reasoning performance shows no clear correlation with generation quality.
\end{finding}

\subsection{Result Analysis of Human Study}
\label{human_study_section}

\begin{wrapfigure}[7]{r}{0.33\textwidth}
    \centering
    \vspace{-4.7em}
    \includegraphics[width=0.3\textwidth]{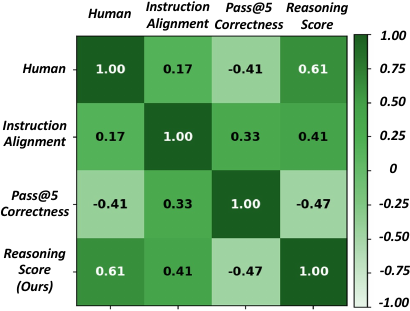}

    \caption{Human study spearman correlation map.} 

    \label{fig:human_study}
\end{wrapfigure}

In Section~\ref{human_study_setup}, we validate whether Reasoning Score (RS) aligns better with human judgment than existing metrics. As shown in Fig.~\ref{fig:human_study}, RS achieves the highest Spearman correlation (0.61), outperforming \textit{Instruction Alignment}~\cite{guo2025video} (0.17) and contrasting with \textit{Pass@5 Correctness}~\cite{li2025viper}(-0.41), which supports the claim that our proposed RS better captures human reasoning behavior and is an effective metric for reasoning coherence evaluation.


\subsection{Result Analysis of Text Hints}
\label{main_result_text_hints_evaluation}

\paragraph{Text hints consistently improve Reasoning Score but hurt consistency.} Across 16 categories, the majority of models show positive RS changes when text hints are provided, for instance, Veo3.1 gains +4.5, Sora2 +7.6, and Cosmos Predict2-14B +6.7 at the overall average level, indicating that textual guidance helps models complete reasoning steps and improve task performance. However, this improvement comes at the cost of consistency: all of the 7 evaluated models experiencing CS degradation, Sora2 and  Seedance 1.0-fast, suggesting that text hints broadly weaken temporal consistency and cross-video stability. Most strikingly, in 4D Dynamics Reasoning, all 7 models suffer CS drops under text hints (ranging from -1.2 to -15.6), raising the question of whether models are genuinely understanding the content or merely leveraging textual cues to produce superficially correct answers. As shown in Fig.~\ref{fig:visual_text_hint_hallucination}(b)(c), these cases illustrate instruction-induced hallucination and inconsistency under textual hints. In the elf example, the model fails to ground the instruction in the original scene and instead `forks' an extra elf to satisfy the motion directive, introducing a non-existent object and causing scene inconsistency. Overall, the model might overfits to textual hints and renders descriptions literally, prioritizing instruction completion over visual faithfulness and physical plausibility.

\begin{finding}{Takeaway 2}
Text hints generally improve Reasoning Score, but consistently 
degrade Consistency Score and cause hallucination, suggesting that explicit guidance may 
shift model attention rather than enhance genuine understanding.
\end{finding}

\begin{table*}[ht]
\centering
\setlength\tabcolsep{6.3pt}
\footnotesize
\caption{
\textbf{Evaluation results (visual-hint).} We report \textit{Reasoning Score (RS)} and \textit{Consistency Score (CS)} in \textbf{no-hint settings} for items that have visual-hint, with \textbf{visual-hint} change in parentheses. Color: \colorbox{green!12}{$\uparrow$~improvement}, \colorbox{red!12}{$\downarrow$~degradation}. Only categories with visual-hint data are shown.
}
\vspace{-3mm}
\label{tab:visual_result}
\resizebox{0.96\textwidth}{!}{%
\begin{tabular}{cc|cccccc|c}
\toprule
\multicolumn{1}{c|}{\multirow{2}{*}{\textbf{Category}}} & \multirow{2}{*}{\textbf{Metric}} & \multicolumn{6}{c|}{\textbf{Closed-Source Models}} & \multicolumn{1}{c}{\textbf{Open-Source Models}} \\ \cmidrule(l){3-9}
\multicolumn{1}{c|}{} & & Veo3.1~\cite{wiedemer2025video} & Veo3.1-fast~\cite{wiedemer2025video} & Sora2~\cite{peng2025open} & \begin{tabular}[c]{@{}c@{}}Seedance\\ 1.0-pro\end{tabular}~\cite{gao2025seedance} & \begin{tabular}[c]{@{}c@{}}Seedance\\ 1.0-fast\end{tabular}~\cite{gao2025seedance} & Kling-v2.1~\cite{team2025klingavatar} & \begin{tabular}[c]{@{}c@{}}Cosmos\\ Predict\end{tabular}~\cite{agarwal2025cosmos} \\ \midrule
\multicolumn{1}{c|}{\multirow{3}*{\begin{tabular}[c]{@{}c@{}}4D Dynamics\\ Reasoning\end{tabular}}} & RS   & 51.7(\colorbox{red!12}{-0.4}) & 70.5(\colorbox{green!12}{+8.5}) & 62.0(\colorbox{red!12}{-3.9}) & 67.4(\colorbox{red!12}{-25.8}) & 51.7(\colorbox{red!12}{-6.6}) & 9.0(\colorbox{red!12}{-5.6}) & 50.8(\colorbox{red!12}{-7.0}) \\[1pt]
\multicolumn{1}{c|}{}                                 & CS   & 63.3(\colorbox{red!12}{-15.6}) & 63.3(\colorbox{green!12}{+2.2}) & 74.4(\colorbox{red!12}{-12.2}) & 53.3(\colorbox{red!12}{-4.4}) & 63.3(\colorbox{red!12}{-21.1}) & 60.0(\colorbox{green!12}{+3.3}) & 56.7(\colorbox{red!12}{-2.2}) \\[1pt]
\multicolumn{1}{c|}{}                                 & \textbf{Avg} & 62.4(\colorbox{red!12}{-10.9}) & 62.2(\colorbox{green!12}{+5.1}) & 71.6(\colorbox{red!12}{-17.1}) & 55.3(\colorbox{red!12}{-5.8}) & 62.4(\colorbox{red!12}{-23.8}) & 61.8(\colorbox{red!12}{-1.3}) & 52.0(\colorbox{green!12}{+0.7}) \\[4pt] \midrule \addlinespace[4pt]

\multicolumn{1}{c|}{\multirow{3}*{\begin{tabular}[c]{@{}c@{}}Embodied\\ Reasoning\end{tabular}}} & RS   & 60.8(\colorbox{green!12}{+22.9}) & 91.2(\colorbox{red!12}{-17.5}) & 68.8(\colorbox{red!12}{-8.3}) & 50.4(\colorbox{green!12}{+8.3}) & 63.8(\colorbox{green!12}{+12.4}) & 9.2(\colorbox{red!12}{-2.9}) & 22.4(\colorbox{green!12}{+20.5}) \\[1pt]
\multicolumn{1}{c|}{}                                 & CS   & 60.0(\colorbox{green!12}{+17.5}) & 80.0(\colorbox{red!12}{-3.8}) & 70.0(\colorbox{green! 12}{+5.0}) & 51.2(\colorbox{green!12}{+8.8}) & 57.1(\colorbox{green!12}{+17.1}) & 23.8(\colorbox{green!12}{+2.5}) & 55.7(\colorbox{green!12}{+12.9}) \\[1pt]
\multicolumn{1}{c|}{}                                 & \textbf{Avg} & 61.5(\colorbox{green!12}{+12.0}) & 81.0(\colorbox{red!12}{-4.2}) & 69.2(\colorbox{green!12}{+2.5}) & 58.8(\colorbox{green!12}{+5.8}) & 60.3(\colorbox{green!12}{+18.3}) & 42.0(\colorbox{green!12}{+2.0}) & 56.3(\colorbox{green!12}{+13.1}) \\[4pt] \midrule \addlinespace[4pt]

\multicolumn{1}{c|}{\multirow{3}*{\begin{tabular}[c]{@{}c@{}}GUI\\ Reasoning\end{tabular}}} & RS   & 27.5(\colorbox{green!12}{+17.5}) & 82.5(\colorbox{green!12}{+5.4}) & 48.3(\colorbox{green!12}{+14.2}) & 25.0(\colorbox{green!12}{+20.0}) & 27.5(\colorbox{green!12}{+17.5}) & 2.1(0.0) & 12.5(0.0) \\[1pt]
\multicolumn{1}{c|}{}                                 & CS   & 50.0(\colorbox{green!12}{+1.2}) & 60.0(\colorbox{red!12}{-3.8}) & 40.0(\colorbox{green!12}{+7.5}) & 35.0(\colorbox{red!12}{-2.5}) & 50.0(\colorbox{green!12}{+1.2}) & 71.2(\colorbox{green!12}{+8.8}) & 35.0(\colorbox{green!12}{+13.8}) \\[1pt]
\multicolumn{1}{c|}{}                                 & \textbf{Avg} & 49.8(\colorbox{green!12}{+1.8}) & 61.0(\colorbox{red!12}{-6.8}) & 38.2(\colorbox{green!12}{+13.0}) & 33.8(\colorbox{red!12}{-6.0}) & 49.8(\colorbox{green!12}{+1.8}) & 64.0(\colorbox{green!12}{+9.5}) & 33.2(\colorbox{green!12}{+10.8}) \\[4pt] \midrule \addlinespace[4pt]

\multicolumn{1}{c|}{\multirow{3}*{\begin{tabular}[c]{@{}c@{}}Medical\\ Reasoning\end{tabular}}} & RS   & 35.7(\colorbox{green!12}{+6.4}) & 31.9(0.0) & 37.9(\colorbox{green!12}{+7.7}) & 21.2(\colorbox{green!12}{+21.5}) & 35.7(\colorbox{red!12}{-6.4}) & 46.4(\colorbox{green!12}{+10.0}) & 23.1(\colorbox{green!12}{+10.6}) \\[1pt]
\multicolumn{1}{c|}{}                                 & CS   & 48.6(\colorbox{red!12}{-12.9}) & 27.5(\colorbox{green!12}{+5.0}) & 47.5(\colorbox{green!12}{+17.5}) & 37.5(\colorbox{green!12}{+23.8}) & 48.6(\colorbox{red!12}{-2.9}) & 65.7(0.0) & 31.2(\colorbox{red!12}{-12.5}) \\[1pt]
\multicolumn{1}{c|}{}                                 & \textbf{Avg} & 48.9(\colorbox{red!12}{-4.9}) & 29.2(\colorbox{green!12}{+4.0}) & 56.8(\colorbox{green!12}{+7.5}) & 40.0(\colorbox{green!12}{+22.5}) & 48.9(\colorbox{green!12}{+1.4}) & 72.0(\colorbox{red!12}{-1.4}) & 26.0(\colorbox{green!12}{+1.2}) \\[4pt] \midrule \addlinespace[4pt]

\multicolumn{1}{c|}{\multirow{3}*{\begin{tabular}[c]{@{}c@{}}Object Counting\\ Reasoning\end{tabular}}} & RS   & 40.0(\colorbox{green!12}{+15.8}) & 60.8(\colorbox{red!12}{-9.2}) & 76.7(\colorbox{green!12}{+0.8}) & 36.7(\colorbox{red!12}{-1.7}) & 37.5(\colorbox{green!12}{+2.9}) & 9.2(\colorbox{green!12}{+10.0}) & 26.7(\colorbox{red!12}{-11.1}) \\[1pt]
\multicolumn{1}{c|}{}                                 & CS   & 40.0(\colorbox{red!12}{-2.5}) & 58.8(\colorbox{red!12}{-6.2}) & 75.0(\colorbox{red!12}{-13.8}) & 48.8(\colorbox{red!12}{-3.8}) & 28.8(\colorbox{green!12}{+3.8}) & 70.0(\colorbox{green!12}{+10.0}) & 35.0(\colorbox{green!12}{+15.0}) \\[1pt]
\multicolumn{1}{c|}{}                                 & \textbf{Avg} & 46.5(\colorbox{red!12}{-6.5}) & 58.0(\colorbox{red!12}{-6.8}) & 76.2(\colorbox{red!12}{-13.5}) & 50.0(\colorbox{red!12}{-0.8}) & 36.5(\colorbox{green!12}{+3.0}) & 82.0(\colorbox{red!12}{-0.5}) & 35.7(\colorbox{green!12}{+16.0}) \\[4pt] \midrule \addlinespace[4pt]

\multicolumn{1}{c|}{\multirow{3}*{\begin{tabular}[c]{@{}c@{}}Table and Chart\\ Reasoning\end{tabular}}} & RS   & 26.7(\colorbox{green!12}{+31.2}) & 52.7(\colorbox{red!12}{-7.3}) & 56.7(\colorbox{red!12}{-2.9}) & 31.7(\colorbox{red!12}{-9.3}) & 26.7(\colorbox{green!12}{+21.9}) & 34.8(\colorbox{green!12}{+17.1}) & 9.5(\colorbox{green!12}{+26.9}) \\[1pt]
\multicolumn{1}{c|}{}                                 & CS   & 33.8(\colorbox{green!12}{+8.8}) & 45.0(\colorbox{red!12}{-16.2}) & 61.2(\colorbox{red!12}{-7.5}) & 28.6(\colorbox{red!12}{-5.7}) & 33.8(\colorbox{green!12}{+8.8}) & 84.3(\colorbox{green!12}{+5.7}) & 17.1(\colorbox{green!12}{+15.7}) \\[1pt]
\multicolumn{1}{c|}{}                                 & \textbf{Avg} & 34.0(\colorbox{green!12}{+6.0}) & 48.2(\colorbox{red!12}{-18.0}) & 55.5(\colorbox{green!12}{+7.8}) & 32.3(\colorbox{red!12}{-6.3}) & 34.0(\colorbox{green!12}{+4.5}) & 88.6(\colorbox{red!12}{-8.0}) & 18.9(\colorbox{green!12}{+13.7}) \\[4pt] \midrule \addlinespace[4pt]

\multicolumn{1}{c|}{\multirow{3}*{\begin{tabular}[c]{@{}c@{}}Visual Detail\\ Reasoning\end{tabular}}} & RS   & 27.8(\colorbox{red!12}{-13.0}) & 45.8(\colorbox{green!12}{+2.8}) & 37.6(\colorbox{green!12}{+17.0}) & 22.6(\colorbox{red!12}{-5.2}) & 27.8(\colorbox{red!12}{-13.0}) & 4.2(\colorbox{red!12}{-4.2}) & 12.5(\colorbox{green!12}{+6.2}) \\[1pt]
\multicolumn{1}{c|}{}                                 & CS   & 54.4(\colorbox{red!12}{-14.4}) & 56.7(\colorbox{red!12}{-1.7}) & 20.0(\colorbox{green!12}{+25.6}) & 47.8(\colorbox{red!12}{-10.0}) & 54.4(\colorbox{red!12}{-13.3}) & 68.8(\colorbox{red!12}{-8.8}) & 42.5(\colorbox{green!12}{+1.2}) \\[1pt]
\multicolumn{1}{c|}{}                                 & \textbf{Avg} & 56.4(\colorbox{red!12}{-10.4}) & 62.7(\colorbox{red!12}{-2.3}) & 23.6(\colorbox{green!12}{+28.0}) & 54.0(\colorbox{red!12}{-15.3}) & 56.4(\colorbox{red!12}{-8.7}) & 77.0(\colorbox{red!12}{-10.8}) & 41.8(\colorbox{green!12}{+4.0}) \\[4pt] \midrule \addlinespace[4pt]

\multicolumn{1}{c|}{\multirow{3}*{\begin{tabular}[c]{@{}c@{}}Visual Trace\\ Reasoning\end{tabular}}} & RS   & 23.7(\colorbox{green!12}{+6.1}) & 35.9(\colorbox{red!12}{-2.8}) & 40.0(\colorbox{green!12}{+10.0}) & 29.4(\colorbox{red!12}{-2.8}) & 23.7(\colorbox{green!12}{+15.5}) & 16.4(0.0) & 15.3(\colorbox{green!12}{+2.1}) \\[1pt]
\multicolumn{1}{c|}{}                                 & CS   & 40.0(\colorbox{green!12}{+2.2}) & 27.8(\colorbox{green!12}{+7.8}) & 61.1(\colorbox{red!12}{-5.6}) & 32.2(\colorbox{red!12}{-6.7}) & 40.0(\colorbox{green!12}{+7.8}) & 27.8(\colorbox{green!12}{+15.6}) & 12.5(\colorbox{green!12}{+17.5}) \\[1pt]
\multicolumn{1}{c|}{}                                 & \textbf{Avg} & 42.0(\colorbox{green!12}{+2.0}) & 34.4(\colorbox{green!12}{+8.2}) & 59.6(\colorbox{red!12}{-2.7}) & 40.0(\colorbox{red!12}{-10.2}) & 42.0(\colorbox{green!12}{+8.7}) & 52.2(\colorbox{green!12}{+3.8}) & 14.5(\colorbox{green!12}{+18.2}) \\[4pt] \midrule \addlinespace[4pt]

\multicolumn{1}{c|}{\multirow{3}*{Average}}         & RS   & 36.9(\colorbox{green!12}{+10.8}) & 58.9(\colorbox{red!12}{-2.5}) & 53.5(\colorbox{green!12}{+4.3}) & 35.5(\colorbox{green!12}{+0.6}) & 36.8(\colorbox{green!12}{+5.5}) & 16.4(\colorbox{green!12}{+3.1}) & 21.6(\colorbox{green!12}{+6.0}) \\[1pt]
\multicolumn{1}{c|}{}                                 & CS   & 48.8(\colorbox{red!12}{-2.0}) & 52.4(\colorbox{red!12}{-2.1}) & 56.2(\colorbox{green!12}{+2.1}) & 41.8(\colorbox{red!12}{-0.1}) & 47.0(\colorbox{green!12}{+0.2}) & 58.9(\colorbox{green!12}{+4.6}) & 35.7(\colorbox{green!12}{+7.7}) \\[1pt]
\multicolumn{1}{c|}{}                                 & \textbf{Avg} & 50.2(\colorbox{red!12}{-1.4}) & 54.6(\colorbox{red!12}{-2.6}) & 56.3(\colorbox{green!12}{+3.2}) & 45.5(\colorbox{red!12}{-2.0}) & 48.8(\colorbox{green!12}{+0.6}) & 67.4(\colorbox{red!12}{-0.8}) & 34.8(\colorbox{green!12}{+9.7}) \\[4pt] \midrule \addlinespace[4pt]
\end{tabular}%
}
\vspace{-3pt}
\end{table*}

\subsection{Result Analysis of Visual Hints}
\label{main_result_visual_hints_evaluation}

Visual hints do not consistently improve performance and can even degrade results on fine-grained visual tasks such as Visual Detail and Object Counting Reasoning. As shown in Tab.~\ref{tab:visual_result}, several models exhibit notable drops in both RS and CS under visual-hint settings for detail-sensitive categories. For example, in Visual Detail Reasoning, Veo3.1 shows a 13.0 drop in RS and 14.4 in CS, while Seedance-1.0-fast drops by 13.0 in RS and 13.3 in CS. We hypothesize that visual hints add salient cues that bias the model toward coarse hint-following rather than precise visual grounding, causing it to overlook subtle details and thus degrade consistency and reasoning accuracy. In contrast, visual hints are more beneficial for structured or spatially guided tasks such as Embodied and GUI Reasoning, where directional cues better align with the reasoning process.

As shown in Fig.~\ref{fig:visual_text_hint_hallucination}, visual hints can also induce hallucination and inconsistency, where the hint itself becomes part of the generated content. In many cases, the arrow used to indicate motion is implicitly rendered or transformed during generation (e.g., in Fig.~\ref{fig:visual_text_hint_hallucination}(b) the arrow turning into a curved trajectory), suggesting that the model treats the visual hint as a drawable object rather than a guiding cue. We hypothesize that this behavior stems from training data distribution biases, where annotated visual markers (e.g., arrows, highlights) frequently co-occur with edited or synthetic content, leading the model to reproduce the hint as part of the scene instead of using it solely for guidance.

\begin{finding}{Takeaway 3}
Visual hints are more effective for structured and spatially guided tasks but are less reliable for fine-grained visual tasks. They also introduce hallucinations by being mistakenly rendered as part of the scene.
\end{finding}

\subsection{Do More Hints Always Improve Reasoning Coherence? A Case Study}

\begin{wrapfigure}[15]{l}{0.69\textwidth}
    \vspace{-0em}
    \centering
    \includegraphics[width=0.69\textwidth]{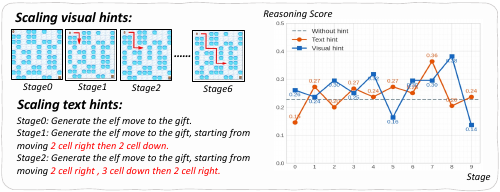}
    \caption{\textbf{Scaling case study on Frozen Lake.} 
    %
    }
    \label{fig:scaling_hint}
    \vspace{-4mm}
\end{wrapfigure}

A natural question arising from our hint analysis is whether 
providing more hints monotonically improves reasoning performance. 
To investigate this, we conduct a preliminary scaling experiment 
on the Frozen Lake~\cite{towers2024gymnasium} using Sora2, progressively increasing 
the number of text and visual hints and measuring  
reasoning score accordingly.

More specifically, as shown in the left part in Fig.~\ref{fig:scaling_hint}, we define the scaling of stage by providing more visual (red arrows for potential directions) and correspondingly, more text instructions. For each setting, we roll out 10 times and report the average reasoning score. Results on the right suggest both text and visual hints generally yield reasoning scores above the no-hint baseline (0.23), suggesting that hint guidance is broadly beneficial. However, neither modality exhibits a monotonic improvement, both curves display substantial fluctuation across stages with no clear upward trend, indicating that current models cannot reliably leverage increasingly detailed hint information in a cumulative manner. This instability suggests that simply increasing hint information is insufficient to guarantee improved reasoning performance. This points to an open research direction: developing models capable of stably grounding multi-step hints into coherent reasoning trajectories.


\section{Conclusion}

In this work, we introduce MME-CoF-Pro, the first benchmark for reasoning coherence evaluation in video generative models. We propose \textit{Reasoning Score} as process-level evaluation metric, and use three hint settings: \textit{no-hint}, \textit{text-hint}, and \textit{visual-hint} to further discover hint-guided performance. Our experiments across seven state-of-the-art models reveal that reasoning ability remains limited and is largely decoupled from generation quality. We further show that hints reshape model behavior in modality-specific ways: text hints generally improve step completion but often reduce consistency and induce hallucinations. Visual hints tend to have task-dependent effects, they are helpful for structured and spatial reasoning, but less reliable for fine-grained perception and sometimes mistakenly incorporated into the scene itself. These findings suggest that current models tend to follow hints rather than truly ground them, highlighting the need for video generative models with stronger visual grounding, instruction understanding abilities and anti-hallucination mechanisms.






\bibliographystyle{splncs04}
\bibliography{main}

\begin{thebibliography}{10}
\providecommand{\url}[1]{\texttt{#1}}
\providecommand{\urlprefix}{URL }
\providecommand{\doi}[1]{https://doi.org/#1}

\bibitem{agarwal2025cosmos}
Agarwal, N., Ali, A., Bala, M., Balaji, Y., Barker, E., Cai, T., Chattopadhyay, P., Chen, Y., Cui, Y., Ding, Y., et~al.: Cosmos world foundation model platform for physical ai. arXiv preprint arXiv:2501.03575  (2025)

\bibitem{ali2025world}
Ali, A., Bai, J., Bala, M., Balaji, Y., Blakeman, A., Cai, T., Cao, J., Cao, T., Cha, E., Chao, Y.W., et~al.: World simulation with video foundation models for physical ai. arXiv preprint arXiv:2511.00062  (2025)

\bibitem{an2026visionlanguagemodelsassess}
An, A., Sun, S., Huang, D., Cheng, M., Gao, Y., Li, J., Qiao, Y., Bian, J.: Can vision language models assess graphic design aesthetics? a benchmark, evaluation, and dataset perspective (2026), \url{https://arxiv.org/abs/2603.01083}

\bibitem{an2026geniusgenerativefluidintelligence}
An, R., Yang, S., Guo, Z., Dai, W., Shen, Z., Li, H., Zhang, R., Wei, X., Li, G., Wu, W., Zhang, W.: Genius: Generative fluid intelligence evaluation suite (2026), \url{https://arxiv.org/abs/2602.11144}

\bibitem{arnab2025temporal}
Arnab, A., Iscen, A., Caron, M., Fathi, A., Schmid, C.: Temporal chain of thought: Long-video understanding by thinking in frames. arXiv preprint arXiv:2507.02001  (2025)

\bibitem{bansal2024videophy}
Bansal, H., Lin, Z., Xie, T., Zong, Z., Yarom, M., Bitton, Y., Jiang, C., Sun, Y., Chang, K.W., Grover, A.: Videophy: Evaluating physical commonsense for video generation. arXiv preprint arXiv:2406.03520  (2024)

\bibitem{brockman2016openai}
Brockman, G., Cheung, V., Pettersson, L., Schneider, J., Schulman, J., Tang, J., Zaremba, W.: Openai gym. arXiv preprint arXiv:1606.01540  (2016)

\bibitem{chen2025tivibench}
Chen, H.H., Lan, D., Shu, W.J., Liu, Q., Wang, Z., Chen, S., Cheng, W., Chen, K., Zhang, H., Zhang, Z., et~al.: Tivibench: Benchmarking think-in-video reasoning for video generative models. arXiv preprint arXiv:2511.13704  (2025)

\bibitem{cheng2024videollama}
Cheng, Z., Leng, S., Zhang, H., Xin, Y., Li, X., Chen, G., Zhu, Y., Zhang, W., Luo, Z., Zhao, D., et~al.: Videollama 2: Advancing spatial-temporal modeling and audio understanding in video-llms. arXiv preprint arXiv:2406.07476  (2024)

\bibitem{chiwow}
Chi, X., Jia, P., Fan, C.K., Ju, X., Mi, W., Zhang, K., Qin, Z., Tian, W., Ge, K., Jia, Y., et~al.: Wow: Scaling embodied omni-world model for generalizable manipulation simulation

\bibitem{chi2025wow}
Chi, X., Jia, P., Fan, C.K., Ju, X., Mi, W., Zhang, K., Qin, Z., Tian, W., Ge, K., Li, H., et~al.: Wow: Towards a world omniscient world model through embodied interaction. arXiv preprint arXiv:2509.22642  (2025)

\bibitem{comanici2025gemini}
Comanici, G., Bieber, E., Schaekermann, M., Pasupat, I., Sachdeva, N., Dhillon, I., Blistein, M., Ram, O., Zhang, D., Rosen, E., et~al.: Gemini 2.5: Pushing the frontier with advanced reasoning, multimodality, long context, and next generation agentic capabilities. arXiv preprint arXiv:2507.06261  (2025)

\bibitem{deng2025scivideobench}
Deng, A., Yang, T., Yu, S., Spencer, L., Bansal, M., Chen, C., Yeung-Levy, S., Wang, X.: Scivideobench: Benchmarking scientific video reasoning in large multimodal models. arXiv preprint arXiv:2510.08559  (2025)

\bibitem{duan2025worldscore}
Duan, H., Yu, H.X., Chen, S., Fei-Fei, L., Wu, J.: Worldscore: A unified evaluation benchmark for world generation. In: Proceedings of the IEEE/CVF International Conference on Computer Vision. pp. 27713--27724 (2025)

\bibitem{fei2024video}
Fei, H., Wu, S., Ji, W., Zhang, H., Zhang, M., Lee, M.L., Hsu, W.: Video-of-thought: Step-by-step video reasoning from perception to cognition. arXiv preprint arXiv:2501.03230  (2024)

\bibitem{gao2023g}
Gao, J., Pi, R., Zhang, J., Ye, J., Zhong, W., Wang, Y., Hong, L., Han, J., Xu, H., Li, Z., et~al.: G-llava: Solving geometric problem with multi-modal large language model. arXiv preprint arXiv:2312.11370  (2023)

\bibitem{gao2025seedance}
Gao, Y., Guo, H., Hoang, T., Huang, W., Jiang, L., Kong, F., Li, H., Li, J., Li, L., Li, X., et~al.: Seedance 1.0: Exploring the boundaries of video generation models. arXiv preprint arXiv:2506.09113  (2025)

\bibitem{guo2025deepseek}
Guo, D., Yang, D., Zhang, H., Song, J., Wang, P., Zhu, Q., Xu, R., Zhang, R., Ma, S., Bi, X., et~al.: Deepseek-r1: Incentivizing reasoning capability in llms via reinforcement learning. arXiv preprint arXiv:2501.12948  (2025)

\bibitem{guo2025rbench}
Guo, M.H., Chu, X., Yang, Q., Mo, Z.H., Shen, Y., Li, P.l., Lin, X., Zhang, J., Chen, X.S., Zhang, Y., et~al.: Rbench-v: A primary assessment for visual reasoning models with multi-modal outputs. arXiv preprint arXiv:2505.16770  (2025)

\bibitem{guo2025video}
Guo, Z., Chen, X., Zhang, R., An, R., Qi, Y., Jiang, D., Li, X., Zhang, M., Li, H., Heng, P.A.: Are video models ready as zero-shot reasoners? an empirical study with the mme-cof benchmark. arXiv preprint arXiv:2510.26802  (2025)

\bibitem{haodong2025tivibench}
Haodong~Chen, H., Lan, D., Shu, W.J., Liu, Q., Wang, Z., Chen, S., Cheng, W., Chen, K., Zhang, H., Zhang, Z., et~al.: Tivibench: Benchmarking think-in-video reasoning for video generative models. arXiv e-prints pp. arXiv--2511 (2025)

\bibitem{he2025ruler}
He, X., Fan, Z., Li, H., Zhuo, F., Xu, H., Cheng, S., Weng, D., Liu, H., Ye, C., Wu, B.: Ruler-bench: Probing rule-based reasoning abilities of next-level video generation models for vision foundation intelligence. arXiv preprint arXiv:2512.02622  (2025)

\bibitem{ho2022imagen}
Ho, J., Chan, W., Saharia, C., Whang, J., Gao, R., Gritsenko, A., Kingma, D.P., Poole, B., Norouzi, M., Fleet, D.J., et~al.: Imagen video: High definition video generation with diffusion models. arXiv preprint arXiv:2210.02303  (2022)

\bibitem{hong2024cogvlm2}
Hong, W., Wang, W., Ding, M., Yu, W., Lv, Q., Wang, Y., Cheng, Y., Huang, S., Ji, J., Xue, Z., et~al.: Cogvlm2: Visual language models for image and video understanding. arXiv preprint arXiv:2408.16500  (2024)

\bibitem{hu2025text2world}
Hu, M., Chen, T., Zou, Y., Lei, Y., Chen, Q., Li, M., Mu, Y., Zhang, H., Shao, W., Luo, P.: Text2world: Benchmarking large language models for symbolic world model generation. In: Findings of the Association for Computational Linguistics: ACL 2025. pp. 26043--26066 (2025)

\bibitem{hu2025sf2t}
Hu, Y., Song, Z., Feng, N., Luo, Y., Yu, J., Chen, Y.P.P., Yang, W.: Sf2t: Self-supervised fragment finetuning of video-llms for fine-grained understanding. In: Proceedings of the Computer Vision and Pattern Recognition Conference. pp. 29108--29117 (2025)

\bibitem{huang2024vbench}
Huang, Z., He, Y., Yu, J., Zhang, F., Si, C., Jiang, Y., Zhang, Y., Wu, T., Jin, Q., Chanpaisit, N., et~al.: Vbench: Comprehensive benchmark suite for video generative models. In: Proceedings of the IEEE/CVF Conference on Computer Vision and Pattern Recognition. pp. 21807--21818 (2024)

\bibitem{jia2025omnispatial}
Jia, M., Qi, Z., Zhang, S., Zhang, W., Yu, X., He, J., Wang, H., Yi, L.: Omnispatial: Towards comprehensive spatial reasoning benchmark for vision language models. arXiv preprint arXiv:2506.03135  (2025)

\bibitem{jin2023mini}
Jin, E., Hu, J., Huang, Z., Zhang, R., Wu, J., Fei-Fei, L., Mart{\'\i}n-Mart{\'\i}n, R.: Mini-behavior: A procedurally generated benchmark for long-horizon decision-making in embodied ai. arXiv preprint arXiv:2310.01824  (2023)

\bibitem{jing2025beyond}
Jing, M., Jia, M., Lin, J., Shen, Z., Gao, H., Xu, M., Li, S.: Beyond classification accuracy: Neural-medbench and the need for deeper reasoning benchmarks. arXiv preprint arXiv:2509.22258  (2025)

\bibitem{kim2024tablevqa}
Kim, Y., Yim, M., Song, K.Y.: Tablevqa-bench: A visual question answering benchmark on multiple table domains. arXiv preprint arXiv:2404.19205  (2024)

\bibitem{li2025imagine}
Li, C., Wu, W., Zhang, H., Xia, Y., Mao, S., Dong, L., Vuli{\'c}, I., Wei, F.: Imagine while reasoning in space: Multimodal visualization-of-thought. arXiv preprint arXiv:2501.07542  (2025)

\bibitem{li2025screenspot}
Li, K., Meng, Z., Lin, H., Luo, Z., Tian, Y., Ma, J., Huang, Z., Chua, T.S.: Screenspot-pro: Gui grounding for professional high-resolution computer use. In: Proceedings of the 33rd ACM International Conference on Multimedia. pp. 8778--8786 (2025)

\bibitem{li2025videochat}
Li, K., He, Y., Wang, Y., Li, Y., Wang, W., Luo, P., Wang, Y., Wang, L., Qiao, Y.: Videochat: Chat-centric video understanding. Science China Information Sciences  \textbf{68}(10),  200102 (2025)

\bibitem{li2024videomamba}
Li, K., Li, X., Wang, Y., He, Y., Wang, Y., Wang, L., Qiao, Y.: Videomamba: State space model for efficient video understanding. In: European conference on computer vision. pp. 237--255. Springer (2024)

\bibitem{li2025unfolding}
Li, L., Bigverdi, M., Gu, J., Ma, Z., Yang, Y., Li, Z., Choi, Y., Krishna, R.: Unfolding spatial cognition: Evaluating multimodal models on visual simulations. arXiv preprint arXiv:2506.04633  (2025)

\bibitem{li2025viper}
Li, Y., Gu, Y., Min, Y., Liu, Z., Du, Y., Zhou, K., Yang, M., Zhao, W.X., Qiu, M.: Viper: Process-aware evaluation for generative video reasoning. arXiv preprint arXiv:2512.24952  (2025)

\bibitem{li2023super}
Li, Z., Wang, X., Stengel-Eskin, E., Kortylewski, A., Ma, W., Van~Durme, B., Yuille, A.L.: Super-clevr: A virtual benchmark to diagnose domain robustness in visual reasoning. In: Proceedings of the IEEE/CVF conference on computer vision and pattern recognition. pp. 14963--14973 (2023)

\bibitem{liang2025worldlens}
Liang, A., Kong, L., Yan, T., Liu, H., Yang, W., Huang, Z., Yin, W., Zuo, J., Hu, Y., Zhu, D., et~al.: Worldlens: Full-spectrum evaluations of driving world models in real world. arXiv preprint arXiv:2512.10958  (2025)

\bibitem{liao2024videoinsta}
Liao, R., Erler, M., Wang, H., Zhai, G., Zhang, G., Ma, Y., Tresp, V.: Videoinsta: Zero-shot long video understanding via informative spatial-temporal reasoning with llms. In: Findings of the Association for Computational Linguistics: EMNLP 2024. pp. 6577--6602 (2024)

\bibitem{liu2025can}
Liu, X., Xu, Z., Li, M., Wang, K., Lee, Y.J., Shang, Y.: Can world simulators reason? gen-vire: A generative visual reasoning benchmark. arXiv preprint arXiv:2511.13853  (2025)

\bibitem{liu2024evalcrafter}
Liu, Y., Cun, X., Liu, X., Wang, X., Zhang, Y., Chen, H., Liu, Y., Zeng, T., Chan, R., Shan, Y.: Evalcrafter: Benchmarking and evaluating large video generation models. In: Proceedings of the IEEE/CVF conference on computer vision and pattern recognition. pp. 22139--22149 (2024)

\bibitem{liu2024sora}
Liu, Y., Zhang, K., Li, Y., Yan, Z., Gao, C., Chen, R., Yuan, Z., Huang, Y., Sun, H., Gao, J., et~al.: Sora: A review on background, technology, limitations, and opportunities of large vision models. arXiv preprint arXiv:2402.17177  (2024)

\bibitem{liu2025stop}
Liu, Z., Xu, K., Su, B., Zou, X., Peng, Y., Zhou, J.: Stop: Integrated spatial-temporal dynamic prompting for video understanding. In: Proceedings of the Computer Vision and Pattern Recognition Conference. pp. 13776--13786 (2025)

\bibitem{lu2024enhancing}
Lu, H., Jian, H., Poppe, R., Salah, A.A.: Enhancing video transformers for action understanding with vlm-aided training. arXiv preprint arXiv:2403.16128  (2024)

\bibitem{lu2022learn}
Lu, P., Mishra, S., Xia, T., Qiu, L., Chang, K.W., Zhu, S.C., Tafjord, O., Clark, P., Kalyan, A.: Learn to explain: Multimodal reasoning via thought chains for science question answering. Advances in neural information processing systems  \textbf{35},  2507--2521 (2022)

\bibitem{luo2025unlocking}
Luo, R., Zheng, Z., Wang, Y., Ni, X., Lin, Z., Jiang, S., Yu, Y., Shi, C., Wang, L., Chu, R., et~al.: Unlocking multimodal mathematical reasoning via process reward model. arXiv preprint arXiv:2501.04686  (2025)

\bibitem{luo2025v}
Luo, Y., Zhao, X., Lin, B., Zhu, L., Tang, L., Liu, Y., Chen, Y.C., Qian, S., Wang, X., You, Y.: V-reasonbench: Toward unified reasoning benchmark suite for video generation models. arXiv preprint arXiv:2511.16668  (2025)

\bibitem{ma2026causalspatial}
Ma, W., Wang, C., Yuan, R., Chen, H., Dai, N., Zhou, S.K., Yang, Y., Yuille, A., Chen, J.: Causalspatial: A benchmark for object-centric causal spatial reasoning. arXiv preprint arXiv:2601.13304  (2026)

\bibitem{maaz2025video}
Maaz, M., Rasheed, H., Khan, F.S., Khan, S.: Video-r2: Reinforcing consistent and grounded reasoning in multimodal language models. arXiv preprint arXiv:2511.23478  (2025)

\bibitem{maes2026stable}
Maes, L., Lidec, Q.L., Haramati, D., Massaudi, N., Scieur, D., LeCun, Y., Balestriero, R.: stable-worldmodel-v1: Reproducible world modeling research and evaluation. arXiv preprint arXiv:2602.08968  (2026)

\bibitem{masry2022chartqa}
Masry, A., Do, X.L., Tan, J.Q., Joty, S., Hoque, E.: Chartqa: A benchmark for question answering about charts with visual and logical reasoning. In: Findings of the association for computational linguistics: ACL 2022. pp. 2263--2279 (2022)

\bibitem{peng2025open}
Peng, X., Zheng, Z., Shen, C., Young, T., Guo, X., Wang, B., Xu, H., Liu, H., Jiang, M., Li, W., et~al.: Open-sora 2.0: Training a commercial-level video generation model in 200 k. arXiv preprint arXiv:2503.09642  (2025)

\bibitem{rasheed2025video}
Rasheed, H., Zumri, M., Maaz, M., Yang, M.H., Khan, F.S., Khan, S.: Video-com: Interactive video reasoning via chain of manipulations. arXiv preprint arXiv:2511.23477  (2025)

\bibitem{royer2024multimedeval}
Royer, C., Menze, B., Sekuboyina, A.: Multimedeval: A benchmark and a toolkit for evaluating medical vision-language models. arXiv preprint arXiv:2402.09262  (2024)

\bibitem{seedance2025seedance}
Seedance, T., Chen, H., Chen, S., Chen, X., Chen, Y., Chen, Y., Chen, Z., Cheng, F., Cheng, T., Cheng, X., et~al.: Seedance 1.5 pro: A native audio-visual joint generation foundation model. arXiv preprint arXiv:2512.13507  (2025)

\bibitem{team2025klingavatar}
Team, K., Chen, J., Ding, Y., Fang, Z., Gai, K., Gao, Y., He, K., Hua, J., Jiang, B., Lao, M., et~al.: Klingavatar 2.0 technical report. arXiv preprint arXiv:2512.13313  (2025)

\bibitem{tian2025envision}
Tian, J., Li, S., He, C., Wu, L., Tan, C.: Envision: Benchmarking unified understanding \& generation for causal world process insights. arXiv preprint arXiv:2512.01816  (2025)

\bibitem{tong2025thinking}
Tong, J., Mou, Y., Li, H., Li, M., Yang, Y., Zhang, M., Chen, Q., Liang, T., Hu, X., Zheng, Y., et~al.: Thinking with video: Video generation as a promising multimodal reasoning paradigm. arXiv preprint arXiv:2511.04570  (2025)

\bibitem{towers2024gymnasium}
Towers, M., Kwiatkowski, A., Terry, J., Balis, J.U., De~Cola, G., Deleu, T., Goul{\~a}o, M., Kallinteris, A., Krimmel, M., KG, A., et~al.: Gymnasium: A standard interface for reinforcement learning environments. arXiv preprint arXiv:2407.17032  (2024)

\bibitem{walke2023bridgedata}
Walke, H.R., Black, K., Zhao, T.Z., Vuong, Q., Zheng, C., Hansen-Estruch, P., He, A.W., Myers, V., Kim, M.J., Du, M., et~al.: Bridgedata v2: A dataset for robot learning at scale. In: Conference on Robot Learning. pp. 1723--1736. PMLR (2023)

\bibitem{wan2025wan}
Wan, T., Wang, A., Ai, B., Wen, B., Mao, C., Xie, C.W., Chen, D., Yu, F., Zhao, H., Yang, J., et~al.: Wan: Open and advanced large-scale video generative models. arXiv preprint arXiv:2503.20314  (2025)

\bibitem{wang2024knowledge}
Wang, G., Zhou, Y., He, Z., Lu, K., Feng, Y., Liu, Z., Wang, G.: Knowledge-guided pre-training and fine-tuning: Video representation learning for action recognition. Neurocomputing  \textbf{571},  127136 (2024)

\bibitem{wang2026very}
Wang, M., Wang, R., Lin, J., Ji, R., Wiedemer, T., Gao, Q., Luo, D., Qian, Y., Huang, L., Hong, Z., et~al.: A very big video reasoning suite. arXiv preprint arXiv:2602.20159  (2026)

\bibitem{wang2025spatialviz}
Wang, S., Pei, M., Sun, L., Deng, C., Shao, K., Tian, Z., Zhang, H., Wang, J.: Spatialviz-bench: An mllm benchmark for spatial visualization. arXiv preprint arXiv:2507.07610  (2025)

\bibitem{wang2024worlddreamer}
Wang, X., Zhu, Z., Huang, G., Wang, B., Chen, X., Lu, J.: Worlddreamer: Towards general world models for video generation via predicting masked tokens. arXiv preprint arXiv:2401.09985  (2024)

\bibitem{wang2025mmbench}
Wang, X., Wu, Z., Xie, J., Ding, Z., Yang, B., Li, Z., Liu, Z., Li, Q., Dong, X., Chen, Z., et~al.: Mmbench-gui: Hierarchical multi-platform evaluation framework for gui agents. arXiv preprint arXiv:2507.19478  (2025)

\bibitem{wang2023foundation}
Wang, Z., Liu, C., Zhang, S., Dou, Q.: Foundation model for endoscopy video analysis via large-scale self-supervised pre-train. In: International conference on medical image computing and computer-assisted intervention. pp. 101--111. Springer (2023)

\bibitem{wiedemer2025video}
Wiedemer, T., Li, Y., Vicol, P., Gu, S.S., Matarese, N., Swersky, K., Kim, B., Jaini, P., Geirhos, R.: Video models are zero-shot learners and reasoners. arXiv preprint arXiv:2509.20328  (2025)

\bibitem{xu2021vlm}
Xu, H., Ghosh, G., Huang, P.Y., Arora, P., Aminzadeh, M., Feichtenhofer, C., Metze, F., Zettlemoyer, L.: Vlm: Task-agnostic video-language model pre-training for video understanding. In: Findings of the Association for Computational Linguistics: ACL-IJCNLP 2021. pp. 4227--4239 (2021)

\bibitem{yan2021videogpt}
Yan, W., Zhang, Y., Abbeel, P., Srinivas, A.: Videogpt: Video generation using vq-vae and transformers. arXiv preprint arXiv:2104.10157  (2021)

\bibitem{yang2025reasoning}
Yang, C., Wan, H., Peng, Y., Cheng, X., Yu, Z., Zhang, J., Yu, J., Yu, X., Zheng, X., Zhou, D., et~al.: Reasoning via video: The first evaluation of video models' reasoning abilities through maze-solving tasks. arXiv preprint arXiv:2511.15065  (2025)

\bibitem{yang2025mmsi}
Yang, S., Xu, R., Xie, Y., Yang, S., Li, M., Lin, J., Zhu, C., Chen, X., Duan, H., Yue, X., et~al.: Mmsi-bench: A benchmark for multi-image spatial intelligence. arXiv preprint arXiv:2505.23764  (2025)

\bibitem{ye2024rotbench}
Ye, J., Wu, Y., Gao, S., Huang, C., Li, S., Li, G., Fan, X., Zhang, Q., Gui, T., Huang, X.J.: Rotbench: A multi-level benchmark for evaluating the robustness of large language models in tool learning. In: Proceedings of the 2024 conference on empirical methods in natural language processing. pp. 313--333 (2024)

\bibitem{yi2019clevrer}
Yi, K., Gan, C., Li, Y., Kohli, P., Wu, J., Torralba, A., Tenenbaum, J.B.: Clevrer: Collision events for video representation and reasoning. arXiv preprint arXiv:1910.01442  (2019)

\bibitem{yuan2025videorefer}
Yuan, Y., Zhang, H., Li, W., Cheng, Z., Zhang, B., Li, L., Li, X., Zhao, D., Zhang, W., Zhuang, Y., et~al.: Videorefer suite: Advancing spatial-temporal object understanding with video llm. In: Proceedings of the Computer Vision and Pattern Recognition Conference. pp. 18970--18980 (2025)

\bibitem{yue2024mmmu}
Yue, X., Ni, Y., Zhang, K., Zheng, T., Liu, R., Zhang, G., Stevens, S., Jiang, D., Ren, W., Sun, Y., et~al.: Mmmu: A massive multi-discipline multimodal understanding and reasoning benchmark for expert agi. In: Proceedings of the IEEE/CVF conference on computer vision and pattern recognition. pp. 9556--9567 (2024)

\bibitem{zhang2025morpheus}
Zhang, C., Cherniavskii, D., Tragoudaras, A., Vozikis, A., Nijdam, T., Prinzhorn, D.W., Bodracska, M., Sebe, N., Zadaianchuk, A., Gavves, E.: Morpheus: Benchmarking physical reasoning of video generative models with real physical experiments. arXiv preprint arXiv:2504.02918  (2025)

\bibitem{zhang2024mavis}
Zhang, R., Wei, X., Jiang, D., Guo, Z., Li, S., Zhang, Y., Tong, C., Liu, J., Zhou, A., Wei, B., et~al.: Mavis: Mathematical visual instruction tuning with an automatic data engine. arXiv preprint arXiv:2407.08739  (2024)

\bibitem{zhang2025video}
Zhang, S., Hao, X., Tang, Y., Zhang, L., Wang, P., Wang, Z., Ma, H., Zhang, S.: Video-cot: A comprehensive dataset for spatiotemporal understanding of videos based on chain-of-thought. In: Proceedings of the 33rd ACM International Conference on Multimedia. pp. 12745--12752 (2025)

\bibitem{zhuang2025math}
Zhuang, W., Huang, X., Zhang, X., Zeng, J.: Math-puma: Progressive upward multimodal alignment to enhance mathematical reasoning. In: Proceedings of the AAAI Conference on Artificial Intelligence. vol.~39, pp. 26183--26191 (2025)

\end{thebibliography}

\clearpage

\section{Reproducibility Statement}

We provide detailed information about the dataset sources, benchmark composition, prompts, and inference configurations in Appendix to facilitate reproducibility.

\section{Ethical Statement}

Our benchmark is designed to evaluate spatial reasoning and physical understanding in video-based environments. 
All scenes involve synthetic or everyday objects performing simple physical interactions (e.g., movement, collision, occlusion, and spatial relationships). 
The benchmark does not contain sensitive, harmful, or violent content.

Human annotators and volunteers are involved in the evaluation pipeline as well as to assess model outputs. 
Before participating, annotators are provided with clear instructions describing the annotation tasks and evaluation criteria. 
All annotators and volunteers provide informed consent prior to participation and take part in the annotation process voluntarily.  They may withdraw from the annotation process at any time.

The benchmark is intended solely for academic research and aims to facilitate systematic evaluation of spatial reasoning capabilities in video generation models. It is not intended for any commercial, surveillance, or decision-making applications.

\section{Data Source}

MME-CoF-Pro comprises 27 diverse benchmarks and datasets, carefully selected to ensure comprehensive coverage of relevant tasks and evaluation scenarios. We acknowledge all original data sources in Tab.~\ref{tab:appendix_mme_cof_source} and strictly adhere to their respective licenses and usage terms. More specifically, the distribution of the data source for each category is listed as follows in Tab.~\ref{tab:appendix_mme_cof_source}.

\begin{table}[htbp]
\centering
\caption{Overview of MME-CoF-Pro by Reasoning Category}
\label{tab:appendix_mme_cof_source}
\small
\begin{tabular}{ll}
\toprule
\textbf{Category} & \textbf{Benchmark} \\
\midrule
Visual Detail Reasoning        & V* Bench~\cite{huang2024vbench} \\
\midrule
\multirow{2}{*}{3D Geometry Reasoning}
                               & STARE~\cite{li2025unfolding} \\
                               & SpatialViz-Bench~\cite{wang2025spatialviz} \\
\midrule
\multirow{2}{*}{Real-world Spatial Reasoning}
                               & MMSI-Bench~\cite{yang2025mmsi} \\
                               & OmniSpatial~\cite{jia2025omnispatial} \\
\midrule
\multirow{6}{*}{Visual Trace Reasoning}
                               & MVoT~\cite{li2025imagine} \\
                               & FrozenLake~\cite{brockman2016openai} \\
                               & Mini-BEHAVIOR~\cite{jin2023mini} \\
                               & OmniSpatial~\cite{jia2025omnispatial} \\
                               & RBench-V~\cite{guo2025rbench} \\
                               & SpatialViz-Bench~\cite{wang2025spatialviz} \\
\midrule
\multirow{3}{*}{Physics-based Reasoning}
                               & MMMU~\cite{yue2024mmmu} \\
                               & ScienceQA~\cite{lu2022learn} \\
                               & RBench-V~\cite{guo2025rbench} \\
\midrule
Natural Science Domain         & ScienceQA~\cite{lu2022learn} \\
\midrule
\multirow{5}{*}{2D Geometry Reasoning}
                               & G-LLaVA~\cite{gao2023g} \\
                               & Math-PUMA~\cite{zhuang2025math} \\
                               & MAVIS~\cite{zhang2024mavis} \\
                               & RBench-V~\cite{guo2025rbench} \\
                               & URSA~\cite{luo2025unlocking} \\
\midrule
\multirow{2}{*}{Table and Chart Reasoning}
                               & ChartQA~\cite{masry2022chartqa} \\
                               & TableVQA~\cite{kim2024tablevqa} \\
\midrule
\multirow{2}{*}{Object Counting Reasoning}
                               & RBench-V~\cite{guo2025rbench} \\
                               & Super-CLEVR~\cite{li2023super} \\
\midrule
\multirow{2}{*}{GUI Reasoning}
                               & ScreenSpot-Pro~\cite{li2025screenspot} \\
                               & MMBench-GUI~\cite{wang2025mmbench} \\
\midrule
\multirow{2}{*}{Medical Reasoning}
                               & Neural-MedBench~\cite{jing2025beyond} \\
                               & MultiMedEval~\cite{royer2024multimedeval} \\
\midrule
Rotation Reasoning             & RotBench~\cite{ye2024rotbench} \\
\midrule
Embodied Reasoning             & BridgeData-v2~\cite{walke2023bridgedata} \\
\midrule
4D Dynamics Reasoning          & CausalSpatial~\cite{ma2026causalspatial} \\
\midrule
Natural Science Domain         & RULER-Bench~\cite{he2025ruler} \\
\midrule
\multirow{2}{*}{Text-based Reasoning}
                               & Thinking with Video~\cite{tong2025thinking} \\
                               & V-ReasonBench~\cite{luo2025v} \\
\bottomrule
\end{tabular}
\end{table}

\section{Human Annotation Pipeline}

To ensure annotation reliability, each sample was annotated and reviewed by multiple human annotators. The reasoning steps were initially drafted by Gemini 2.5 Pro~\cite{comanici2025gemini} and subsequently refined through three rounds of manual verification and correction by 10 human annotators. Each annotator is a domain expert in computer science or is pursuing a relevant undergraduate degree in a related field. Images with visual hints were manually annotated using Photopea, while strictly preserving the original image resolution. During this process, annotators carefully cross-checked each reasoning step to ensure that it was (1) necessary for the intended reasoning trajectory and (2) unambiguously verifiable from the generated video. Any disagreements were resolved through discussion until consensus was reached. The final annotated reasoning steps represent the minimal set of necessary checkpoints required for successful task completion. This design reduces ambiguity while preserving evaluation consistency across annotators.

\section{Human Study Experiment Details}

More specifically, ten human participants took part in the study on a fully voluntary basis after providing informed consent. 
Participants were informed about the purpose of the study and were free to withdraw from the annotation process at any time without any penalty or consequences.  The study involves only simple evaluation of model-generated videos and poses minimal risk to participants. 
No personally identifiable information was collected during the annotation process. The annotator group maintained a balanced distribution in terms of age and gender. Each participant was asked to score 10 videos numerically based on the coherence of the reasoning process rather than the visual generation quality. 
A portion of the study questionnaire is provided below.

\begin{mycode}

Human Study: Reasoning Quality Evaluation

Please evaluate the quality of the reasoning process for each question (IDs 1–10).

Score range: 0–10. Decimal scores are allowed (e.g., 7.5).

Evaluation target: Based on whether the model behavior in the video satisfies the requirements of each question below, provide an overall assessment of the reasoning quality.

Video description: The videos are generated by the Sora model. Each prompt corresponds to one video:
videos/1/video.mp4 ~ videos/10/video.mp4.
The question ID matches the corresponding folder number.

\end{mycode}

The evaluation prompts and scoring protocols for \textbf{Instruction Alignment}~\cite{guo2025video} and \textbf{Pass@5 Correctness}~\cite{li2025viper} follow those defined in the original papers. We run the evaluation on the same batch of video, and calculate the Spearman Correlation.

\section{Model Inference Configurations}


All models are evaluated using their default inference configurations
and recommended resolutions provided by the official APIs or repository. We provide details in Tab.~\ref{tab:appendix_inference_config}.

\begin{table}[h]
\centering
\small
\caption{Default inference configurations for evaluated video generation models. When configurable, we follow the recommended settings from the official implementations or APIs.}
\begin{tabular}{lcccc}
\toprule
Model & Resolution & Duration & FPS \\
\midrule
Veo-3.1~\cite{wiedemer2025video} & 1280$\times$720 & 8s & 24 \\
Veo-3.1-fast~\cite{wiedemer2025video} & 1280$\times$720 & 8s & 24 \\
Sora-2~\cite{peng2025open} & 1280$\times$720 & 4s & 30 \\
Seedance-1.0-pro~\cite{gao2025seedance} & 1920$\times$1280 & 5s & 24 \\
Seedance-1.0-fast~\cite{gao2025seedance} & 1280$\times$720 & 5s & 24 \\
Kling-v2.1~\cite{team2025klingavatar} & 1280$\times$720 & 5s & 24 \\
Cosmos Predict-2.0-14B~\cite{agarwal2025cosmos} & 1280$\times$720 & 5s & 24 \\
\bottomrule
\end{tabular}
\label{tab:appendix_inference_config}
\end{table}



\section{Detailed Experiment Results}

We provide the full experiment results in Tab.~\ref{tab:appendix_text_result} and Tab.~\ref{tab:appendix_visual_result} regarding the four dimensions under generation quality: \textit{Temporal Consistency}, \textit{Visual Stability}, \textit{Hallucination}, \textit{Physics Grounding}. The remaining \textit{Consistency Score}, which is used to evaluate the content consistency, is reported in the main paper. \textit{Reasoning Score}, which is used to evaluation reasoning coherence, is also reported in the main paper.

\begin{table*}[ht]
\centering
\setlength\tabcolsep{6.3pt}
\footnotesize
\caption{
\textbf{Evaluation results (text-hint): Temporal Consistency, Visual Stability, Hallucination, Physics Grounding.} We report no-hint score (0-9 mean as \%); parenthetical change is \textbf{text-hint} minus no-hint. \colorbox{green!12}{$\uparrow$~improvement}, \colorbox{red!12}{$\downarrow$~degradation}.
}
\vspace{-3mm}
\label{tab:appendix_text_result}
\resizebox{0.90\textwidth}{!}{%
\begin{tabular}{cc|cccccc|c}
\toprule
\multicolumn{1}{c|}{\multirow{2}{*}{\textbf{Category}}} & \multirow{2}{*}{\textbf{Metric}} & \multicolumn{6}{c|}{\textbf{Closed-Source Models}} & \multicolumn{1}{c}{\textbf{Open-Source Models}} \\ \cmidrule(l){3-9}
\multicolumn{1}{c|}{} & & Veo3.1~\cite{wiedemer2025video} & Veo3.1-fast~\cite{wiedemer2025video} & Sora2~\cite{peng2025open} & \begin{tabular}[c]{@{}c@{}}Seedance\\ 1.0-pro\end{tabular}~\cite{gao2025seedance} & \begin{tabular}[c]{@{}c@{}}Seedance\\ 1.0-fast\end{tabular}~\cite{gao2025seedance} & Kling-v2.1~\cite{team2025klingavatar} & \begin{tabular}[c]{@{}c@{}}Cosmos\\ Predict\end{tabular} \\ \midrule
\multicolumn{1}{c|}{\multirow{4}*{\begin{tabular}[c]{@{}c@{}}Visual Detail\\ Reasoning\end{tabular}}} & TC   & 64.8(\colorbox{red!12}{-7.4}) & 55.6(\colorbox{red!12}{-2.0}) & 22.8(\colorbox{green!12}{+11.7}) & 60.1(\colorbox{green!12}{+9.2}) & 64.8(\colorbox{red!12}{-7.4}) & 76.4(\colorbox{red!12}{-6.2}) & 37.7(\colorbox{red!12}{-8.0}) \\[1pt]
\multicolumn{1}{c|}{}                                 & VS   & 61.7(\colorbox{red!12}{-1.9}) & 51.6(\colorbox{red!12}{-2.0}) & 36.4(\colorbox{green!12}{+5.6}) & 58.2(\colorbox{green!12}{+7.2}) & 61.7(\colorbox{red!12}{-1.9}) & 88.2(\colorbox{red!12}{-17.4}) & 43.8(\colorbox{red!12}{-7.4}) \\[1pt]
\multicolumn{1}{c|}{}                                 & Hallu   & 66.7(\colorbox{red!12}{-4.9}) & 74.5(\colorbox{red!12}{-6.5}) & 38.3(\colorbox{green!12}{+30.9}) & 66.0(\colorbox{green!12}{+1.3}) & 66.7(\colorbox{red!12}{-4.9}) & 88.2(\colorbox{green!12}{+4.2}) & 45.1(\colorbox{red!12}{-9.3}) \\[1pt]
\multicolumn{1}{c|}{}                                 & PG   & 67.9(\colorbox{red!12}{-0.6}) & 80.4(\colorbox{red!12}{-5.2}) & 47.5(\colorbox{green!12}{+13.0}) & 64.7(\colorbox{green!12}{+10.5}) & 67.9(\colorbox{red!12}{-0.6}) & 79.2(\colorbox{red!12}{-4.9}) & 43.8(\colorbox{red!12}{-11.1}) \\[4pt] \midrule \addlinespace[4pt]

\multicolumn{1}{c|}{\multirow{4}*{\begin{tabular}[c]{@{}c@{}}Embodied\\ Reasoning\end{tabular}}} & TC   & 65.9(\colorbox{red!12}{-4.8}) & 68.3(\colorbox{green!12}{+11.5}) & 76.6(\colorbox{red!12}{-8.7}) & 66.7(\colorbox{red!12}{-3.2}) & 65.9(\colorbox{red!12}{-4.8}) & 72.0(\colorbox{red!12}{-11.9}) & 56.8(\colorbox{red!12}{-1.7}) \\[1pt]
\multicolumn{1}{c|}{}                                 & VS   & 71.0(\colorbox{red!12}{-2.0}) & 71.0(\colorbox{green!12}{+11.5}) & 80.6(\colorbox{red!12}{-8.3}) & 69.4(\colorbox{red!12}{-4.4}) & 71.0(\colorbox{red!12}{-2.0}) & 85.4(\colorbox{red!12}{-1.9}) & 66.7(\colorbox{red!12}{-4.3}) \\[1pt]
\multicolumn{1}{c|}{}                                 & Hallu   & 66.3(\colorbox{red!12}{-4.4}) & 72.6(\colorbox{green!12}{+12.3}) & 71.0(\colorbox{red!12}{-6.7}) & 60.7(\colorbox{red!12}{-0.8}) & 66.3(\colorbox{red!12}{-4.4}) & 88.9(\colorbox{green!12}{+4.2}) & 67.9(\colorbox{red!12}{-6.4}) \\[1pt]
\multicolumn{1}{c|}{}                                 & PG   & 63.5(0.0) & 70.6(\colorbox{green!12}{+10.7}) & 66.7(\colorbox{red!12}{-2.4}) & 69.4(\colorbox{red!12}{-10.3}) & 63.5(0.0) & 9.2(\colorbox{green!12}{+0.4}) & 56.0(\colorbox{green!12}{+4.3}) \\[4pt] \midrule \addlinespace[4pt]

\multicolumn{1}{c|}{\multirow{4}*{\begin{tabular}[c]{@{}c@{}}Object Counting\\ Reasoning\end{tabular}}} & TC   & 41.4(\colorbox{red!12}{-0.6}) & 40.1(\colorbox{green!12}{+12.3}) & 67.3(\colorbox{red!12}{-6.5}) & 48.4(\colorbox{red!12}{-11.8}) & 41.4(\colorbox{red!12}{-0.6}) & 94.8(\colorbox{red!12}{-6.5}) & 40.1(\colorbox{green!12}{+15.4}) \\[1pt]
\multicolumn{1}{c|}{}                                 & VS   & 49.4(0.0) & 64.8(\colorbox{green!12}{+6.8}) & 89.5(\colorbox{red!12}{-5.9}) & 49.7(\colorbox{red!12}{-1.3}) & 49.4(0.0) & 100.0(\colorbox{red!12}{-5.9}) & 48.1(\colorbox{green!12}{+14.8}) \\[1pt]
\multicolumn{1}{c|}{}                                 & Hallu   & 28.4(\colorbox{green!12}{+6.8}) & 51.2(\colorbox{green!12}{+2.5}) & 79.7(\colorbox{red!12}{-1.3}) & 28.8(\colorbox{green!12}{+5.9}) & 28.4(\colorbox{green!12}{+6.8}) & 94.1(\colorbox{green!12}{+2.0}) & 34.6(\colorbox{green!12}{+11.1}) \\[1pt]
\multicolumn{1}{c|}{}                                 & PG   & 51.2(\colorbox{green!12}{+7.4}) & 64.2(\colorbox{green!12}{+8.6}) & 79.1(\colorbox{green!12}{+6.5}) & 51.0(\colorbox{red!12}{-3.3}) & 51.2(\colorbox{green!12}{+7.4}) & 82.4(\colorbox{red!12}{-4.6}) & 42.6(\colorbox{green!12}{+17.9}) \\[4pt] \midrule \addlinespace[4pt]

\multicolumn{1}{c|}{\multirow{4}*{\begin{tabular}[c]{@{}c@{}}Table and Chart\\ Reasoning\end{tabular}}} & TC   & 43.2(\colorbox{red!12}{-1.2}) & 46.3(\colorbox{red!12}{-5.6}) & 64.2(\colorbox{red!12}{-2.5}) & 34.1(\colorbox{red!12}{-7.9}) & 43.2(\colorbox{red!12}{-1.2}) & 100.0(\colorbox{red!12}{-7.1}) & 33.3(\colorbox{green!12}{+1.9}) \\[1pt]
\multicolumn{1}{c|}{}                                 & VS   & 46.3(\colorbox{green!12}{+1.9}) & 51.9(\colorbox{red!12}{-15.4}) & 80.2(\colorbox{green!12}{+3.1}) & 34.1(\colorbox{red!12}{-11.9}) & 46.3(\colorbox{green!12}{+1.9}) & 100.0(\colorbox{red!12}{-7.1}) & 45.1(\colorbox{green!12}{+11.1}) \\[1pt]
\multicolumn{1}{c|}{}                                 & Hallu   & 37.0(\colorbox{red!12}{-3.7}) & 37.7(\colorbox{red!12}{-6.8}) & 80.2(\colorbox{red!12}{-10.5}) & 23.8(\colorbox{red!12}{-9.5}) & 37.0(\colorbox{red!12}{-3.7}) & 96.0(\colorbox{green!12}{+3.2}) & 30.2(\colorbox{green!12}{+1.9}) \\[1pt]
\multicolumn{1}{c|}{}                                 & PG   & 43.2(\colorbox{green!12}{+9.9}) & 52.5(\colorbox{red!12}{-6.8}) & 74.7(\colorbox{red!12}{-2.5}) & 40.5(\colorbox{red!12}{-13.5}) & 43.2(\colorbox{green!12}{+9.9}) & 97.6(\colorbox{red!12}{-17.5}) & 42.0(\colorbox{green!12}{+0.6}) \\[4pt] \midrule \addlinespace[4pt]

\multicolumn{1}{c|}{\multirow{4}*{\begin{tabular}[c]{@{}c@{}}2D Geometry\\ Reasoning\end{tabular}}} & TC   & 48.9(\colorbox{red!12}{-7.8}) & 29.6(\colorbox{green!12}{+27.2}) & 71.1(\colorbox{green!12}{+10.0}) & 51.1(\colorbox{green!12}{+8.9}) & 48.9(\colorbox{red!12}{-7.8}) & 60.0(\colorbox{green!12}{+40.0}) & 23.3(\colorbox{green!12}{+13.3}) \\[1pt]
\multicolumn{1}{c|}{}                                 & VS   & 52.2(\colorbox{red!12}{-1.1}) & 51.9(\colorbox{green!12}{+9.9}) & 86.7(\colorbox{green!12}{+7.8}) & 63.3(\colorbox{green!12}{+17.8}) & 52.2(\colorbox{red!12}{-1.1}) & 71.1(\colorbox{green!12}{+28.9}) & 32.2(\colorbox{green!12}{+12.2}) \\[1pt]
\multicolumn{1}{c|}{}                                 & Hallu   & 25.6(\colorbox{green!12}{+11.1}) & 32.1(\colorbox{red!12}{-2.5}) & 76.7(\colorbox{red!12}{-7.8}) & 30.0(\colorbox{green!12}{+11.1}) & 25.6(\colorbox{green!12}{+11.1}) & 100.0(\colorbox{red!12}{-11.1}) & 26.7(\colorbox{red!12}{-3.3}) \\[1pt]
\multicolumn{1}{c|}{}                                 & PG   & 36.7(\colorbox{green!12}{+3.3}) & 28.4(\colorbox{green!12}{+17.3}) & 73.3(\colorbox{green!12}{+2.2}) & 47.8(\colorbox{green!12}{+14.4}) & 36.7(\colorbox{green!12}{+3.3}) & 41.1(0.0) & 28.9(\colorbox{red!12}{-7.8}) \\[4pt] \midrule \addlinespace[4pt]

\multicolumn{1}{c|}{\multirow{4}*{\begin{tabular}[c]{@{}c@{}}GUI\\ Reasoning\end{tabular}}} & TC   & 58.6(\colorbox{red!12}{-3.7}) & 65.4(\colorbox{green!12}{+4.3}) & 45.1(\colorbox{green!12}{+23.6}) & 40.7(\colorbox{green!12}{+1.9}) & 58.6(\colorbox{red!12}{-3.7}) & 89.5(\colorbox{red!12}{-4.9}) & 48.8(\colorbox{green!12}{+1.9}) \\[1pt]
\multicolumn{1}{c|}{}                                 & VS   & 65.4(\colorbox{green!12}{+4.9}) & 75.9(\colorbox{green!12}{+4.9}) & 69.4(\colorbox{green!12}{+15.3}) & 50.0(\colorbox{red!12}{-3.1}) & 65.4(\colorbox{green!12}{+4.9}) & 90.1(\colorbox{red!12}{-0.6}) & 53.7(\colorbox{green!12}{+13.0}) \\[1pt]
\multicolumn{1}{c|}{}                                 & Hallu   & 53.7(\colorbox{red!12}{-7.4}) & 71.6(\colorbox{green!12}{+1.9}) & 60.4(\colorbox{green!12}{+20.1}) & 24.7(\colorbox{green!12}{+4.9}) & 53.7(\colorbox{red!12}{-7.4}) & 94.4(\colorbox{green!12}{+4.3}) & 40.7(\colorbox{green!12}{+20.4}) \\[1pt]
\multicolumn{1}{c|}{}                                 & PG   & 53.1(\colorbox{red!12}{-3.1}) & 77.2(\colorbox{red!12}{-7.4}) & 48.6(\colorbox{green!12}{+9.7}) & 31.5(\colorbox{green!12}{+5.6}) & 53.1(\colorbox{red!12}{-3.1}) & 6.8(\colorbox{green!12}{+8.0}) & 32.7(\colorbox{green!12}{+2.5}) \\[4pt] \midrule \addlinespace[4pt]

\multicolumn{1}{c|}{\multirow{4}*{\begin{tabular}[c]{@{}c@{}}Physics-based\\ Reasoning\end{tabular}}} & TC   & 39.4(\colorbox{green!12}{+3.3}) & 37.8(\colorbox{red!12}{-1.1}) & 44.4(\colorbox{green!12}{+6.1}) & 59.0(\colorbox{red!12}{-12.0}) & 39.4(\colorbox{green!12}{+3.3}) & 57.3(\colorbox{green!12}{+14.5}) & 42.6(\colorbox{red!12}{-7.4}) \\[1pt]
\multicolumn{1}{c|}{}                                 & VS   & 47.8(\colorbox{green!12}{+6.7}) & 43.3(\colorbox{red!12}{-2.2}) & 66.1(\colorbox{green!12}{+11.1}) & 61.5(\colorbox{red!12}{-4.3}) & 47.8(\colorbox{green!12}{+6.7}) & 100.0(0.0) & 42.0(\colorbox{red!12}{-1.9}) \\[1pt]
\multicolumn{1}{c|}{}                                 & Hallu   & 33.9(\colorbox{green!12}{+12.8}) & 42.8(\colorbox{red!12}{-5.6}) & 60.0(\colorbox{green!12}{+12.8}) & 49.6(\colorbox{red!12}{-1.7}) & 33.9(\colorbox{green!12}{+12.8}) & 100.0(\colorbox{red!12}{-2.6}) & 28.4(0.0) \\[1pt]
\multicolumn{1}{c|}{}                                 & PG   & 18.3(\colorbox{green!12}{+0.6}) & 35.6(\colorbox{red!12}{-12.8}) & 37.2(\colorbox{green!12}{+5.0}) & 41.9(\colorbox{red!12}{-12.0}) & 18.3(\colorbox{green!12}{+0.6}) & 7.7(\colorbox{green!12}{+0.9}) & 21.6(\colorbox{green!12}{+10.5}) \\[4pt] \midrule \addlinespace[4pt]

\multicolumn{1}{c|}{\multirow{4}*{\begin{tabular}[c]{@{}c@{}}Text-based\\ Reasoning\end{tabular}}} & TC   & 70.5(\colorbox{red!12}{-9.7}) & 61.7(\colorbox{green!12}{+12.2}) & 68.1(\colorbox{red!12}{-8.2}) & 67.6(\colorbox{red!12}{-5.3}) & 70.5(\colorbox{red!12}{-9.7}) & 56.5(\colorbox{red!12}{-5.3}) & 57.0(\colorbox{red!12}{-2.4}) \\[1pt]
\multicolumn{1}{c|}{}                                 & VS   & 74.4(\colorbox{red!12}{-12.1}) & 76.1(\colorbox{green!12}{+6.7}) & 81.2(\colorbox{red!12}{-6.3}) & 75.4(\colorbox{red!12}{-1.9}) & 74.4(\colorbox{red!12}{-12.1}) & 64.3(\colorbox{red!12}{-0.5}) & 65.2(\colorbox{red!12}{-4.3}) \\[1pt]
\multicolumn{1}{c|}{}                                 & Hallu   & 47.8(\colorbox{red!12}{-3.4}) & 41.7(\colorbox{green!12}{+5.6}) & 63.8(\colorbox{red!12}{-5.3}) & 40.6(\colorbox{red!12}{-11.6}) & 47.8(\colorbox{red!12}{-3.4}) & 92.3(\colorbox{red!12}{-10.1}) & 37.7(\colorbox{green!12}{+0.5}) \\[1pt]
\multicolumn{1}{c|}{}                                 & PG   & 72.9(\colorbox{red!12}{-10.1}) & 68.3(\colorbox{green!12}{+13.3}) & 84.1(\colorbox{red!12}{-5.8}) & 71.5(\colorbox{green!12}{+0.5}) & 72.9(\colorbox{red!12}{-10.1}) & 44.4(0.0) & 62.3(\colorbox{red!12}{-10.1}) \\[4pt] \midrule \addlinespace[4pt]

\multicolumn{1}{c|}{\multirow{4}*{\begin{tabular}[c]{@{}c@{}}3D Geometry\\ Reasoning\end{tabular}}} & TC   & 47.5(\colorbox{red!12}{-12.3}) & 41.4(\colorbox{red!12}{-7.4}) & 43.2(\colorbox{green!12}{+3.7}) & 37.3(\colorbox{red!12}{-3.9}) & 47.5(\colorbox{red!12}{-12.3}) & 71.9(\colorbox{green!12}{+15.7}) & 37.0(\colorbox{green!12}{+7.4}) \\[1pt]
\multicolumn{1}{c|}{}                                 & VS   & 38.3(\colorbox{red!12}{-3.7}) & 40.7(\colorbox{red!12}{-7.4}) & 46.3(\colorbox{green!12}{+1.2}) & 41.2(\colorbox{red!12}{-5.9}) & 38.3(\colorbox{red!12}{-3.7}) & 83.7(\colorbox{green!12}{+3.3}) & 40.7(\colorbox{green!12}{+11.7}) \\[1pt]
\multicolumn{1}{c|}{}                                 & Hallu   & 27.2(\colorbox{red!12}{-1.9}) & 29.0(\colorbox{red!12}{-1.2}) & 47.5(\colorbox{red!12}{-13.6}) & 22.2(\colorbox{red!12}{-3.9}) & 27.2(\colorbox{red!12}{-1.9}) & 94.8(\colorbox{red!12}{-2.0}) & 34.6(\colorbox{green!12}{+6.2}) \\[1pt]
\multicolumn{1}{c|}{}                                 & PG   & 27.2(\colorbox{red!12}{-6.2}) & 27.8(\colorbox{red!12}{-1.2}) & 40.7(\colorbox{green!12}{+1.9}) & 28.1(\colorbox{red!12}{-2.6}) & 27.2(\colorbox{red!12}{-6.2}) & 7.2(\colorbox{red!12}{-2.6}) & 16.7(\colorbox{green!12}{+19.8}) \\[4pt] \midrule \addlinespace[4pt]

\multicolumn{1}{c|}{\multirow{4}*{\begin{tabular}[c]{@{}c@{}}Medical\\ Reasoning\end{tabular}}} & TC   & 62.1(\colorbox{green!12}{+3.9}) & 29.6(\colorbox{red!12}{-4.3}) & 71.0(\colorbox{red!12}{-9.9}) & 48.8(\colorbox{green!12}{+4.3}) & 62.1(\colorbox{green!12}{+3.9}) & 78.4(\colorbox{green!12}{+10.5}) & 38.9(\colorbox{green!12}{+1.2}) \\[1pt]
\multicolumn{1}{c|}{}                                 & VS   & 54.9(\colorbox{green!12}{+7.2}) & 23.5(\colorbox{red!12}{-1.2}) & 74.7(\colorbox{red!12}{-5.6}) & 42.6(\colorbox{green!12}{+2.5}) & 54.9(\colorbox{green!12}{+7.2}) & 79.6(\colorbox{green!12}{+9.9}) & 36.4(\colorbox{green!12}{+8.0}) \\[1pt]
\multicolumn{1}{c|}{}                                 & Hallu   & 58.8(\colorbox{red!12}{-3.3}) & 25.3(\colorbox{red!12}{-3.1}) & 90.7(\colorbox{red!12}{-12.3}) & 47.5(\colorbox{red!12}{-6.8}) & 58.8(\colorbox{red!12}{-3.3}) & 85.2(0.0) & 45.1(\colorbox{green!12}{+13.0}) \\[1pt]
\multicolumn{1}{c|}{}                                 & PG   & 58.8(\colorbox{green!12}{+9.8}) & 25.3(0.0) & 90.1(\colorbox{red!12}{-22.2}) & 40.7(\colorbox{green!12}{+5.6}) & 58.8(\colorbox{green!12}{+9.8}) & 58.6(\colorbox{red!12}{-11.1}) & 38.9(\colorbox{green!12}{+9.9}) \\[4pt] \midrule \addlinespace[4pt]

\multicolumn{1}{c|}{\multirow{4}*{\begin{tabular}[c]{@{}c@{}}Real-world Spatial\\ Reasoning\end{tabular}}} & TC   & 60.5(\colorbox{red!12}{-8.6}) & 72.5(\colorbox{red!12}{-7.2}) & 73.9(\colorbox{red!12}{-5.9}) & 54.9(\colorbox{green!12}{+0.6}) & 60.5(\colorbox{red!12}{-8.6}) & 67.3(\colorbox{green!12}{+16.7}) & 48.1(\colorbox{green!12}{+12.3}) \\[1pt]
\multicolumn{1}{c|}{}                                 & VS   & 71.6(\colorbox{red!12}{-11.7}) & 75.2(\colorbox{green!12}{+2.0}) & 80.4(\colorbox{green!12}{+1.3}) & 64.2(0.0) & 71.6(\colorbox{red!12}{-11.7}) & 89.5(\colorbox{green!12}{+4.9}) & 60.5(\colorbox{green!12}{+5.6}) \\[1pt]
\multicolumn{1}{c|}{}                                 & Hallu   & 60.5(\colorbox{red!12}{-2.5}) & 67.3(\colorbox{green!12}{+9.8}) & 57.5(\colorbox{green!12}{+12.4}) & 47.5(\colorbox{green!12}{+9.9}) & 60.5(\colorbox{red!12}{-2.5}) & 87.0(\colorbox{red!12}{-0.6}) & 50.6(\colorbox{green!12}{+9.9}) \\[1pt]
\multicolumn{1}{c|}{}                                 & PG   & 64.2(\colorbox{red!12}{-2.5}) & 74.5(\colorbox{green!12}{+0.7}) & 79.7(\colorbox{red!12}{-2.6}) & 63.6(\colorbox{red!12}{-3.1}) & 64.2(\colorbox{red!12}{-2.5}) & 84.0(\colorbox{red!12}{-8.6}) & 55.6(\colorbox{green!12}{+9.3}) \\[4pt] \midrule \addlinespace[4pt]

\multicolumn{1}{c|}{\multirow{4}*{\begin{tabular}[c]{@{}c@{}}Visual Logical\\ Reasoning\end{tabular}}} & TC   & 44.4(\colorbox{red!12}{-13.5}) & 47.5(\colorbox{red!12}{-3.5}) & 63.1(\colorbox{red!12}{-7.6}) & 31.4(\colorbox{green!12}{+3.9}) & 44.4(\colorbox{red!12}{-13.5}) & 83.1(\colorbox{red!12}{-12.6}) & 25.9(\colorbox{green!12}{+3.2}) \\[1pt]
\multicolumn{1}{c|}{}                                 & VS   & 50.7(\colorbox{red!12}{-11.1}) & 60.1(\colorbox{green!12}{+2.0}) & 84.8(\colorbox{red!12}{-10.6}) & 38.6(\colorbox{green!12}{+3.9}) & 50.7(\colorbox{red!12}{-11.1}) & 95.2(\colorbox{red!12}{-1.9}) & 31.2(\colorbox{green!12}{+6.3}) \\[1pt]
\multicolumn{1}{c|}{}                                 & Hallu   & 34.3(\colorbox{red!12}{-6.3}) & 43.9(\colorbox{green!12}{+2.0}) & 70.2(\colorbox{red!12}{-0.5}) & 20.3(\colorbox{green!12}{+11.1}) & 34.3(\colorbox{red!12}{-6.3}) & 75.4(\colorbox{green!12}{+6.3}) & 11.1(\colorbox{green!12}{+19.0}) \\[1pt]
\multicolumn{1}{c|}{}                                 & PG   & 40.6(\colorbox{red!12}{-8.7}) & 53.0(\colorbox{green!12}{+4.0}) & 68.7(\colorbox{red!12}{-2.0}) & 32.0(\colorbox{green!12}{+5.9}) & 40.6(\colorbox{red!12}{-8.7}) & 54.1(\colorbox{green!12}{+10.1}) & 27.5(\colorbox{green!12}{+11.6}) \\[4pt] \midrule \addlinespace[4pt]

\multicolumn{1}{c|}{\multirow{4}*{\begin{tabular}[c]{@{}c@{}}4D Dynamics\\ Reasoning\end{tabular}}} & TC   & 65.4(\colorbox{red!12}{-1.9}) & 66.0(\colorbox{green!12}{+2.5}) & 78.4(\colorbox{red!12}{-14.2}) & 62.3(\colorbox{red!12}{-11.7}) & 65.4(\colorbox{red!12}{-1.9}) & 83.3(\colorbox{red!12}{-16.0}) & 47.5(\colorbox{red!12}{-11.7}) \\[1pt]
\multicolumn{1}{c|}{}                                 & VS   & 76.5(0.0) & 79.0(0.0) & 88.3(\colorbox{red!12}{-13.6}) & 72.8(\colorbox{red!12}{-8.0}) & 76.5(0.0) & 83.3(\colorbox{green!12}{+4.9}) & 59.9(\colorbox{red!12}{-6.2}) \\[1pt]
\multicolumn{1}{c|}{}                                 & Hallu   & 69.8(\colorbox{red!12}{-5.6}) & 73.5(\colorbox{red!12}{-6.2}) & 80.2(\colorbox{red!12}{-8.6}) & 61.7(\colorbox{red!12}{-11.1}) & 69.8(\colorbox{red!12}{-5.6}) & 98.1(\colorbox{red!12}{-8.0}) & 56.8(\colorbox{red!12}{-10.5}) \\[1pt]
\multicolumn{1}{c|}{}                                 & PG   & 55.6(\colorbox{red!12}{-10.5}) & 58.6(\colorbox{green!12}{+2.5}) & 54.9(\colorbox{red!12}{-9.3}) & 51.9(\colorbox{red!12}{-4.3}) & 55.6(\colorbox{red!12}{-10.5}) & 0.6(0.0) & 34.0(\colorbox{red!12}{-6.8}) \\[4pt] \midrule \addlinespace[4pt]

\multicolumn{1}{c|}{\multirow{4}*{\begin{tabular}[c]{@{}c@{}}Natural Science\\ Reasoning\end{tabular}}} & TC   & 65.2(\colorbox{green!12}{+2.2}) & 57.4(\colorbox{green!12}{+6.2}) & 62.3(\colorbox{green!12}{+3.1}) & 70.1(\colorbox{red!12}{-3.5}) & 65.2(\colorbox{green!12}{+2.2}) & 83.3(\colorbox{red!12}{-32.7}) & 59.3(\colorbox{red!12}{-14.8}) \\[1pt]
\multicolumn{1}{c|}{}                                 & VS   & 74.1(\colorbox{red!12}{-3.0}) & 69.1(\colorbox{green!12}{+4.9}) & 73.5(\colorbox{green!12}{+1.2}) & 77.1(0.0) & 74.1(\colorbox{red!12}{-3.0}) & 96.9(\colorbox{red!12}{-7.4}) & 68.5(\colorbox{red!12}{-16.7}) \\[1pt]
\multicolumn{1}{c|}{}                                 & Hallu   & 56.3(\colorbox{green!12}{+7.4}) & 64.2(\colorbox{green!12}{+13.0}) & 80.2(\colorbox{red!12}{-7.4}) & 64.6(\colorbox{red!12}{-1.4}) & 56.3(\colorbox{green!12}{+7.4}) & 99.4(\colorbox{red!12}{-4.3}) & 68.5(\colorbox{red!12}{-25.0}) \\[1pt]
\multicolumn{1}{c|}{}                                 & PG   & 55.6(\colorbox{green!12}{+3.0}) & 57.4(\colorbox{green!12}{+13.0}) & 59.3(\colorbox{green!12}{+6.2}) & 48.6(\colorbox{green!12}{+2.8}) & 55.6(\colorbox{green!12}{+3.0}) & 26.5(\colorbox{red!12}{-9.3}) & 52.8(\colorbox{red!12}{-5.6}) \\[4pt] \midrule \addlinespace[4pt]

\multicolumn{1}{c|}{\multirow{4}*{\begin{tabular}[c]{@{}c@{}}Rotation\\ Reasoning\end{tabular}}} & TC   & 52.1(\colorbox{red!12}{-9.7}) & 39.6(\colorbox{green!12}{+16.0}) & 76.4(\colorbox{red!12}{-2.8}) & 40.5(\colorbox{green!12}{+11.1}) & 52.1(\colorbox{red!12}{-9.7}) & 94.1(0.0) & 44.4(\colorbox{red!12}{-2.6}) \\[1pt]
\multicolumn{1}{c|}{}                                 & VS   & 56.9(\colorbox{red!12}{-12.5}) & 40.3(\colorbox{green!12}{+11.8}) & 81.9(\colorbox{green!12}{+3.5}) & 43.7(\colorbox{green!12}{+12.7}) & 56.9(\colorbox{red!12}{-12.5}) & 94.8(\colorbox{red!12}{-0.7}) & 45.8(\colorbox{red!12}{-6.5}) \\[1pt]
\multicolumn{1}{c|}{}                                 & Hallu   & 50.0(\colorbox{red!12}{-10.4}) & 42.4(\colorbox{green!12}{+3.5}) & 78.5(\colorbox{red!12}{-2.1}) & 46.0(\colorbox{green!12}{+7.1}) & 50.0(\colorbox{red!12}{-10.4}) & 94.1(\colorbox{green!12}{+3.7}) & 38.6(\colorbox{red!12}{-7.8}) \\[1pt]
\multicolumn{1}{c|}{}                                 & PG   & 47.9(\colorbox{red!12}{-24.3}) & 37.5(\colorbox{green!12}{+6.9}) & 52.8(\colorbox{green!12}{+4.9}) & 37.3(\colorbox{green!12}{+19.0}) & 47.9(\colorbox{red!12}{-24.3}) & 34.8(\colorbox{green!12}{+17.0}) & 32.0(\colorbox{red!12}{-9.2}) \\[4pt] \midrule \addlinespace[4pt]

\multicolumn{1}{c|}{\multirow{4}*{\begin{tabular}[c]{@{}c@{}}Visual Trace\\ Reasoning\end{tabular}}} & TC   & 50.0(\colorbox{red!12}{-5.6}) & 49.7(\colorbox{green!12}{+9.8}) & 59.9(\colorbox{green!12}{+9.3}) & 44.4(\colorbox{red!12}{-2.5}) & 50.0(\colorbox{red!12}{-5.6}) & 63.6(\colorbox{red!12}{-9.3}) & 14.4(\colorbox{green!12}{+17.0}) \\[1pt]
\multicolumn{1}{c|}{}                                 & VS   & 74.1(\colorbox{red!12}{-13.6}) & 69.3(\colorbox{red!12}{-3.9}) & 90.7(\colorbox{green!12}{+0.6}) & 64.2(\colorbox{green!12}{+6.2}) & 74.1(\colorbox{red!12}{-13.6}) & 89.5(\colorbox{red!12}{-3.1}) & 33.3(\colorbox{green!12}{+12.4}) \\[1pt]
\multicolumn{1}{c|}{}                                 & Hallu   & 56.8(\colorbox{red!12}{-7.4}) & 56.2(\colorbox{red!12}{-4.6}) & 80.9(\colorbox{red!12}{-4.3}) & 54.3(\colorbox{red!12}{-11.7}) & 56.8(\colorbox{red!12}{-7.4}) & 88.3(\colorbox{green!12}{+4.3}) & 27.5(\colorbox{red!12}{-0.7}) \\[1pt]
\multicolumn{1}{c|}{}                                 & PG   & 40.1(\colorbox{green!12}{+3.7}) & 52.3(\colorbox{green!12}{+2.0}) & 64.2(\colorbox{green!12}{+9.3}) & 38.3(\colorbox{green!12}{+7.4}) & 40.1(\colorbox{green!12}{+3.7}) & 12.3(\colorbox{green!12}{+0.6}) & 17.0(\colorbox{green!12}{+9.8}) \\[4pt] \midrule \addlinespace[4pt]


\end{tabular}%
}
\vspace{-3pt}
\end{table*}

\begin{table*}[ht]
\centering
\setlength\tabcolsep{6.3pt}
\footnotesize
\caption{
\textbf{Evaluation results (visual-hint): Temporal Consistency, Visual Stability, Hallucination, Physics Grounding.} No-hint score (\%); parenthetical change is \textbf{visual-hint} minus no-hint. \colorbox{green!12}{$\uparrow$~improvement}, \colorbox{red!12}{$\downarrow$~degradation}. Only categories with visual-hint data.
}
\vspace{-3mm}
\label{tab:appendix_visual_result}
\resizebox{0.96\textwidth}{!}{%
\begin{tabular}{cc|cccccc|c}
\toprule
\multicolumn{1}{c|}{\multirow{2}{*}{\textbf{Category}}} & \multirow{2}{*}{\textbf{Metric}} & \multicolumn{6}{c|}{\textbf{Closed-Source Models}} & \multicolumn{1}{c}{\textbf{Open-Source Models}} \\ \cmidrule(l){3-9}
\multicolumn{1}{c|}{} & & Veo3.1~\cite{wiedemer2025video} & Veo3.1-fast~\cite{wiedemer2025video} & Sora2~\cite{peng2025open} & \begin{tabular}[c]{@{}c@{}}Seedance\\ 1.0-pro\end{tabular}~\cite{gao2025seedance} & \begin{tabular}[c]{@{}c@{}}Seedance\\ 1.0-fast\end{tabular}~\cite{gao2025seedance} & Kling-v2.1~\cite{team2025klingavatar} & \begin{tabular}[c]{@{}c@{}}Cosmos\\ Predict\end{tabular} \\ \midrule
\multicolumn{1}{c|}{\multirow{4}*{\begin{tabular}[c]{@{}c@{}}4D Dynamics\\ Reasoning\end{tabular}}} & TC   & 67.9(\colorbox{red!12}{-9.9}) & 67.9(\colorbox{green!12}{+2.5}) & 74.1(\colorbox{red!12}{-16.0}) & 63.0(\colorbox{red!12}{-8.6}) & 67.9(\colorbox{red!12}{-22.2}) & 88.9(\colorbox{red!12}{-23.5}) & 51.9(\colorbox{green!12}{+6.2}) \\[1pt]
\multicolumn{1}{c|}{}                                 & VS   & 77.8(\colorbox{red!12}{-7.4}) & 77.8(\colorbox{green!12}{+4.9}) & 87.7(\colorbox{red!12}{-21.0}) & 76.5(\colorbox{red!12}{-4.9}) & 77.8(\colorbox{red!12}{-21.0}) & 88.9(\colorbox{green!12}{+11.1}) & 69.1(\colorbox{green!12}{+1.2}) \\[1pt]
\multicolumn{1}{c|}{}                                 & Hallu   & 75.3(\colorbox{red!12}{-12.3}) & 72.8(\colorbox{green!12}{+8.6}) & 86.4(\colorbox{red!12}{-22.2}) & 64.2(\colorbox{red!12}{-4.9}) & 75.3(\colorbox{red!12}{-37.0}) & 98.8(\colorbox{green!12}{+1.2}) & 67.9(\colorbox{red!12}{-3.7}) \\[1pt]
\multicolumn{1}{c|}{}                                 & PG   & 55.6(\colorbox{red!12}{-13.6}) & 56.8(\colorbox{green!12}{+9.9}) & 66.7(\colorbox{red!12}{-22.2}) & 44.4(\colorbox{red!12}{-8.6}) & 55.6(\colorbox{red!12}{-28.4}) & 0.0(0.0) & 37.0(\colorbox{green!12}{+2.5}) \\[4pt] \midrule \addlinespace[4pt]

\multicolumn{1}{c|}{\multirow{4}*{\begin{tabular}[c]{@{}c@{}}Embodied\\ Reasoning\end{tabular}}} & TC   & 69.4(\colorbox{green!12}{+12.5}) & 86.1(\colorbox{red!12}{-9.7}) & 70.8(\colorbox{green!12}{+5.6}) & 63.9(\colorbox{green!12}{+4.2}) & 68.3(\colorbox{green!12}{+19.0}) & 50.0(\colorbox{red!12}{-20.8}) & 60.3(\colorbox{green!12}{+12.7}) \\[1pt]
\multicolumn{1}{c|}{}                                 & VS   & 76.4(\colorbox{green!12}{+2.8}) & 95.8(\colorbox{red!12}{-6.9}) & 79.2(\colorbox{green!12}{+8.3}) & 72.2(\colorbox{green!12}{+9.7}) & 74.6(\colorbox{green!12}{+19.0}) & 75.0(\colorbox{green!12}{+9.7}) & 76.2(\colorbox{green!12}{+7.9}) \\[1pt]
\multicolumn{1}{c|}{}                                 & Hallu   & 75.0(\colorbox{green!12}{+9.7}) & 93.1(\colorbox{green!12}{+1.4}) & 86.1(\colorbox{red!12}{-2.8}) & 61.1(\colorbox{green!12}{+8.3}) & 71.4(\colorbox{green!12}{+12.7}) & 75.0(\colorbox{green!12}{+25.0}) & 66.7(\colorbox{green!12}{+22.2}) \\[1pt]
\multicolumn{1}{c|}{}                                 & PG   & 54.2(\colorbox{green!12}{+22.2}) & 86.1(\colorbox{red!12}{-4.2}) & 70.8(\colorbox{red!12}{-2.8}) & 72.2(0.0) & 57.1(\colorbox{green!12}{+31.7}) & 6.9(\colorbox{red!12}{-5.6}) & 47.6(\colorbox{green!12}{+15.9}) \\[4pt] \midrule \addlinespace[4pt]

\multicolumn{1}{c|}{\multirow{4}*{\begin{tabular}[c]{@{}c@{}}GUI\\ Reasoning\end{tabular}}} & TC   & 56.9(0.0) & 59.7(\colorbox{red!12}{-8.3}) & 34.7(\colorbox{green!12}{+25.0}) & 41.7(\colorbox{red!12}{-13.9}) & 56.9(0.0) & 87.5(\colorbox{green!12}{+12.5}) & 41.7(\colorbox{green!12}{+6.9}) \\[1pt]
\multicolumn{1}{c|}{}                                 & VS   & 62.5(\colorbox{green!12}{+5.6}) & 69.4(\colorbox{red!12}{-2.8}) & 48.6(\colorbox{green!12}{+20.8}) & 51.4(\colorbox{red!12}{-9.7}) & 62.5(\colorbox{green!12}{+5.6}) & 88.9(\colorbox{green!12}{+11.1}) & 45.8(\colorbox{green!12}{+13.9}) \\[1pt]
\multicolumn{1}{c|}{}                                 & Hallu   & 51.4(\colorbox{red!12}{-5.6}) & 66.7(\colorbox{red!12}{-8.3}) & 37.5(\colorbox{green!12}{+12.5}) & 25.0(\colorbox{red!12}{-8.3}) & 51.4(\colorbox{red!12}{-5.6}) & 100.0(0.0) & 34.7(\colorbox{green!12}{+5.6}) \\[1pt]
\multicolumn{1}{c|}{}                                 & PG   & 50.0(\colorbox{green!12}{+8.3}) & 76.4(\colorbox{red!12}{-13.9}) & 47.2(\colorbox{green!12}{+5.6}) & 30.6(\colorbox{green!12}{+1.4}) & 50.0(\colorbox{green!12}{+8.3}) & 0.0(\colorbox{green!12}{+19.4}) & 23.6(\colorbox{green!12}{+18.1}) \\[4pt] \midrule \addlinespace[4pt]

\multicolumn{1}{c|}{\multirow{4}*{\begin{tabular}[c]{@{}c@{}}Medical\\ Reasoning\end{tabular}}} & TC   & 57.1(\colorbox{red!12}{-3.2}) & 37.5(\colorbox{green!12}{+1.4}) & 43.1(\colorbox{green!12}{+31.9}) & 50.0(\colorbox{green!12}{+31.9}) & 57.1(\colorbox{green!12}{+14.3}) & 85.7(\colorbox{red!12}{-12.7}) & 30.6(\colorbox{red!12}{-9.7}) \\[1pt]
\multicolumn{1}{c|}{}                                 & VS   & 55.6(\colorbox{red!12}{-7.9}) & 26.4(\colorbox{green!12}{+9.7}) & 58.3(\colorbox{green!12}{+13.9}) & 45.8(\colorbox{green!12}{+26.4}) & 55.6(\colorbox{green!12}{+4.8}) & 85.7(\colorbox{red!12}{-12.7}) & 30.6(\colorbox{red!12}{-1.4}) \\[1pt]
\multicolumn{1}{c|}{}                                 & Hallu   & 49.2(0.0) & 33.3(0.0) & 79.2(\colorbox{red!12}{-8.3}) & 38.9(\colorbox{green!12}{+20.8}) & 49.2(\colorbox{red!12}{-6.3}) & 76.2(\colorbox{green!12}{+12.7}) & 27.8(\colorbox{green!12}{+9.7}) \\[1pt]
\multicolumn{1}{c|}{}                                 & PG   & 55.6(\colorbox{red!12}{-1.6}) & 34.7(\colorbox{green!12}{+5.6}) & 81.9(\colorbox{red!12}{-15.3}) & 45.8(\colorbox{green!12}{+19.4}) & 55.6(\colorbox{red!12}{-1.6}) & 79.4(\colorbox{green!12}{+4.8}) & 20.8(\colorbox{green!12}{+22.2}) \\[4pt] \midrule \addlinespace[4pt]

\multicolumn{1}{c|}{\multirow{4}*{\begin{tabular}[c]{@{}c@{}}Object Counting\\ Reasoning\end{tabular}}} & TC   & 52.8(\colorbox{red!12}{-8.3}) & 37.5(\colorbox{green!12}{+5.6}) & 63.9(\colorbox{red!12}{-22.2}) & 63.9(\colorbox{red!12}{-6.9}) & 40.3(\colorbox{green!12}{+9.7}) & 88.9(\colorbox{red!12}{-1.4}) & 37.0(\colorbox{green!12}{+14.8}) \\[1pt]
\multicolumn{1}{c|}{}                                 & VS   & 55.6(\colorbox{red!12}{-11.1}) & 75.0(\colorbox{red!12}{-9.7}) & 87.5(\colorbox{red!12}{-16.7}) & 65.3(\colorbox{red!12}{-5.6}) & 45.8(\colorbox{green!12}{+8.3}) & 100.0(\colorbox{red!12}{-11.1}) & 44.4(\colorbox{green!12}{+11.1}) \\[1pt]
\multicolumn{1}{c|}{}                                 & Hallu   & 33.3(\colorbox{red!12}{-8.3}) & 63.9(\colorbox{red!12}{-16.7}) & 93.1(\colorbox{red!12}{-19.4}) & 27.8(\colorbox{green!12}{+18.1}) & 25.0(\colorbox{red!12}{-2.8}) & 88.9(0.0) & 33.3(\colorbox{green!12}{+20.4}) \\[1pt]
\multicolumn{1}{c|}{}                                 & PG   & 72.2(\colorbox{red!12}{-5.6}) & 80.6(\colorbox{red!12}{-9.7}) & 95.8(\colorbox{red!12}{-1.4}) & 66.7(\colorbox{red!12}{-5.6}) & 59.7(\colorbox{red!12}{-2.8}) & 100.0(\colorbox{red!12}{-1.4}) & 44.4(\colorbox{green!12}{+25.9}) \\[4pt] \midrule \addlinespace[4pt]

\multicolumn{1}{c|}{\multirow{4}*{\begin{tabular}[c]{@{}c@{}}Table and Chart\\ Reasoning\end{tabular}}} & TC   & 38.9(\colorbox{red!12}{-1.4}) & 56.9(\colorbox{red!12}{-22.2}) & 55.6(\colorbox{green!12}{+19.4}) & 41.3(\colorbox{red!12}{-1.6}) & 38.9(\colorbox{red!12}{-1.4}) & 100.0(0.0) & 23.8(\colorbox{green!12}{+4.8}) \\[1pt]
\multicolumn{1}{c|}{}                                 & VS   & 48.6(\colorbox{red!12}{-6.9}) & 55.6(\colorbox{red!12}{-20.8}) & 59.7(\colorbox{green!12}{+16.7}) & 34.9(\colorbox{red!12}{-4.8}) & 48.6(\colorbox{red!12}{-6.9}) & 100.0(0.0) & 25.4(\colorbox{green!12}{+6.3}) \\[1pt]
\multicolumn{1}{c|}{}                                 & Hallu   & 37.5(\colorbox{green!12}{+11.1}) & 48.6(\colorbox{red!12}{-16.7}) & 72.2(\colorbox{green!12}{+2.8}) & 31.7(\colorbox{red!12}{-17.5}) & 37.5(\colorbox{green!12}{+2.8}) & 100.0(\colorbox{red!12}{-1.6}) & 7.9(\colorbox{green!12}{+36.5}) \\[1pt]
\multicolumn{1}{c|}{}                                 & PG   & 26.4(\colorbox{green!12}{+20.8}) & 56.9(\colorbox{red!12}{-22.2}) & 52.8(\colorbox{green!12}{+12.5}) & 39.7(\colorbox{red!12}{-4.8}) & 26.4(\colorbox{green!12}{+20.8}) & 98.4(\colorbox{red!12}{-49.2}) & 28.6(\colorbox{green!12}{+11.1}) \\[4pt] \midrule \addlinespace[4pt]

\multicolumn{1}{c|}{\multirow{4}*{\begin{tabular}[c]{@{}c@{}}Visual Detail\\ Reasoning\end{tabular}}} & TC   & 63.0(\colorbox{red!12}{-12.3}) & 59.3(\colorbox{red!12}{-1.9}) & 21.0(\colorbox{green!12}{+22.2}) & 60.5(\colorbox{red!12}{-9.9}) & 63.0(\colorbox{red!12}{-13.6}) & 76.4(\colorbox{red!12}{-9.7}) & 47.2(\colorbox{green!12}{+2.8}) \\[1pt]
\multicolumn{1}{c|}{}                                 & VS   & 63.0(\colorbox{red!12}{-7.4}) & 66.7(\colorbox{red!12}{-9.3}) & 24.7(\colorbox{green!12}{+29.6}) & 59.3(\colorbox{red!12}{-12.3}) & 63.0(\colorbox{red!12}{-6.2}) & 100.0(\colorbox{red!12}{-11.1}) & 54.2(0.0) \\[1pt]
\multicolumn{1}{c|}{}                                 & Hallu   & 64.2(\colorbox{red!12}{-14.8}) & 75.9(\colorbox{red!12}{-7.4}) & 27.2(\colorbox{green!12}{+39.5}) & 63.0(\colorbox{red!12}{-29.6}) & 64.2(\colorbox{red!12}{-12.3}) & 100.0(0.0) & 40.3(\colorbox{green!12}{+8.3}) \\[1pt]
\multicolumn{1}{c|}{}                                 & PG   & 63.0(\colorbox{red!12}{-7.4}) & 83.3(\colorbox{green!12}{+7.4}) & 35.8(\colorbox{green!12}{+35.8}) & 64.2(\colorbox{red!12}{-22.2}) & 63.0(\colorbox{red!12}{-1.2}) & 75.0(\colorbox{red!12}{-29.2}) & 43.1(\colorbox{green!12}{+9.7}) \\[4pt] \midrule \addlinespace[4pt]

\multicolumn{1}{c|}{\multirow{4}*{\begin{tabular}[c]{@{}c@{}}Visual Trace\\ Reasoning\end{tabular}}} & TC   & 45.7(\colorbox{red!12}{-8.6}) & 28.4(\colorbox{green!12}{+13.6}) & 49.4(\colorbox{green!12}{+12.3}) & 40.7(\colorbox{red!12}{-7.4}) & 45.7(\colorbox{green!12}{+1.2}) & 69.1(\colorbox{red!12}{-2.5}) & 8.3(\colorbox{green!12}{+37.5}) \\[1pt]
\multicolumn{1}{c|}{}                                 & VS   & 66.7(\colorbox{red!12}{-8.6}) & 55.6(\colorbox{green!12}{+3.7}) & 86.4(\colorbox{red!12}{-7.4}) & 61.7(\colorbox{red!12}{-18.5}) & 66.7(\colorbox{green!12}{+3.7}) & 88.9(\colorbox{green!12}{+11.1}) & 22.2(\colorbox{green!12}{+29.2}) \\[1pt]
\multicolumn{1}{c|}{}                                 & Hallu   & 45.7(\colorbox{green!12}{+4.9}) & 34.6(\colorbox{green!12}{+14.8}) & 70.4(\colorbox{red!12}{-7.4}) & 43.2(\colorbox{red!12}{-13.6}) & 45.7(\colorbox{green!12}{+8.6}) & 87.7(\colorbox{red!12}{-8.6}) & 25.0(\colorbox{green!12}{+18.1}) \\[1pt]
\multicolumn{1}{c|}{}                                 & PG   & 30.9(\colorbox{green!12}{+21.0}) & 42.0(\colorbox{green!12}{+4.9}) & 56.8(\colorbox{red!12}{-6.2}) & 40.7(\colorbox{red!12}{-9.9}) & 30.9(\colorbox{green!12}{+25.9}) & 13.6(\colorbox{green!12}{+3.7}) & 11.1(\colorbox{red!12}{-2.8}) \\[4pt] \midrule \addlinespace[4pt]


\end{tabular}%
}
\vspace{-3pt}
\end{table*}

\clearpage

\section{Data Examples}

We provide the following examples for visualizations.

\begin{figure}[h]
\centering
\includegraphics[width=\linewidth]{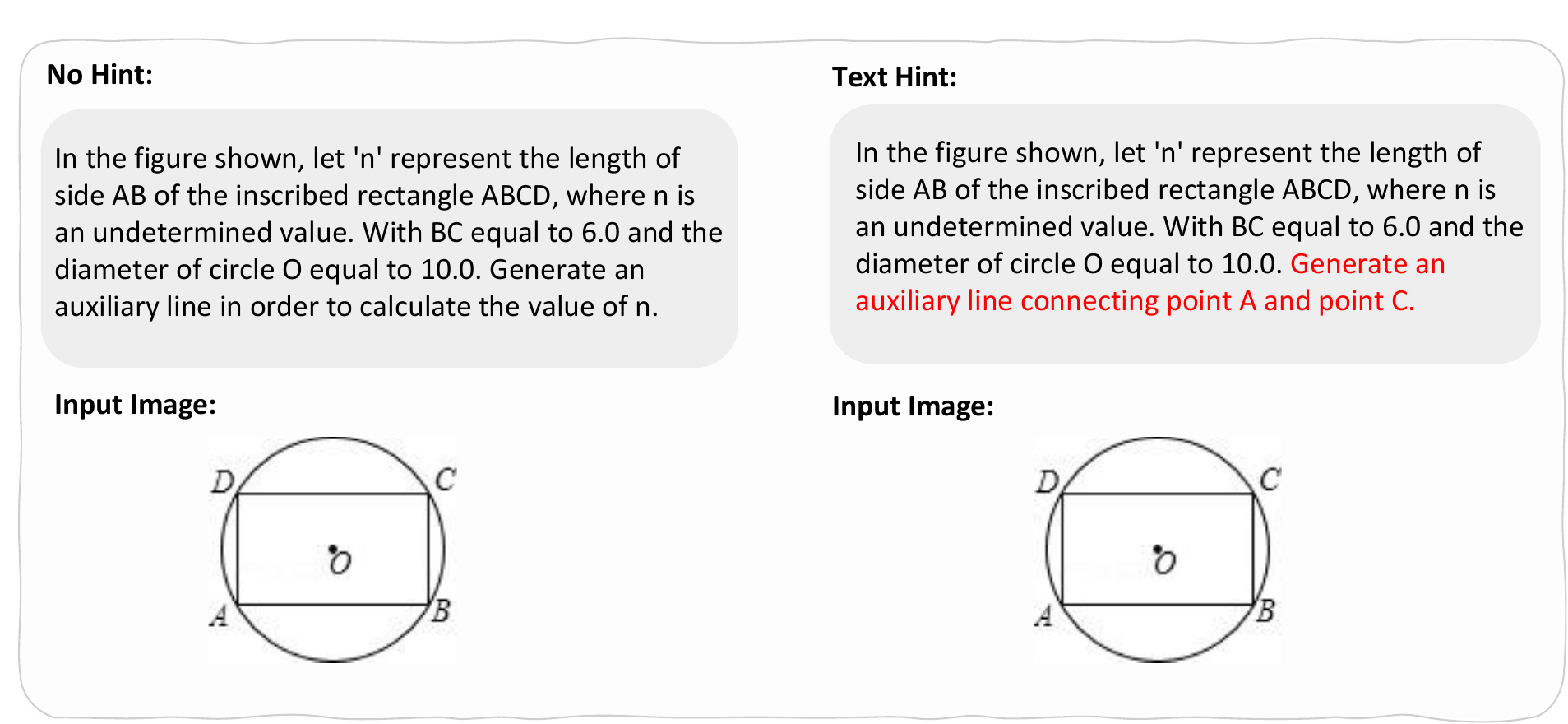}
\end{figure}

\begin{figure}[h]
\centering
\includegraphics[width=\linewidth]{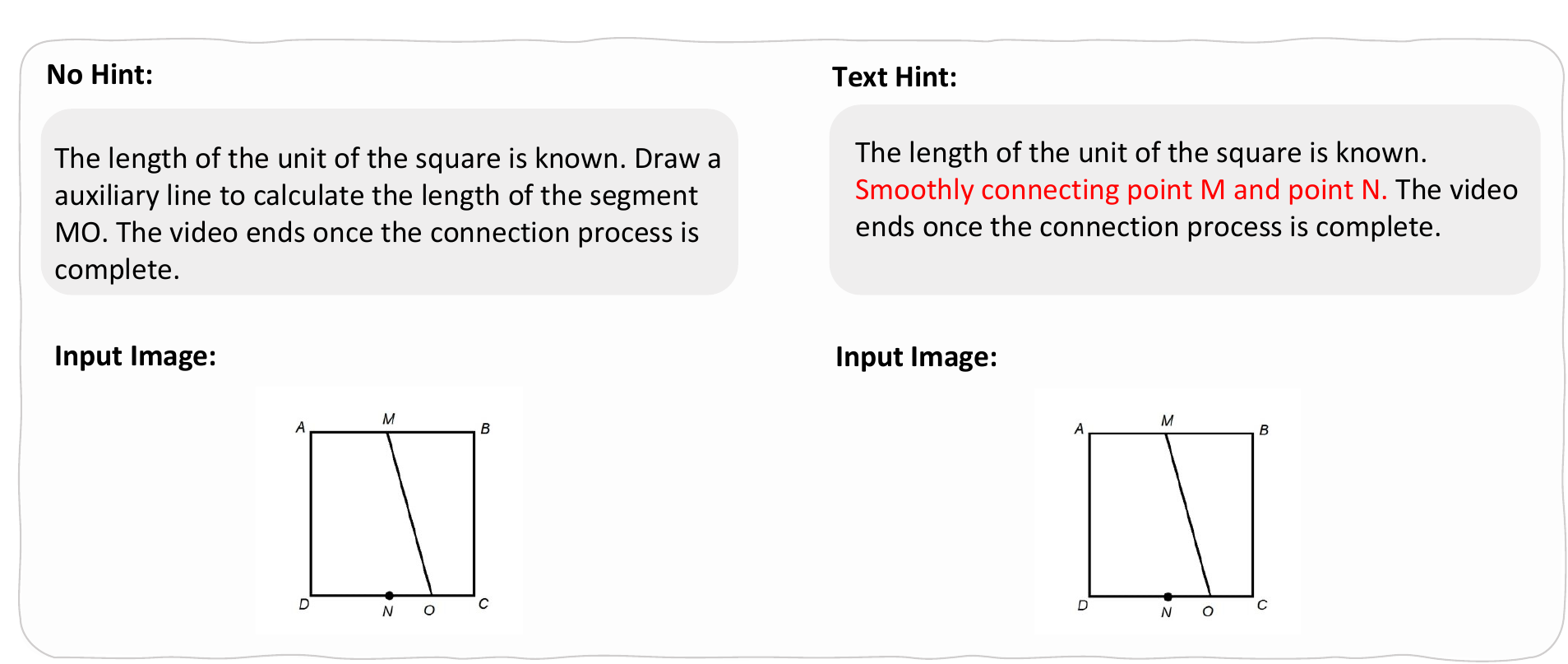}
\end{figure}

\begin{figure}[h]
\centering
\includegraphics[width=\linewidth]{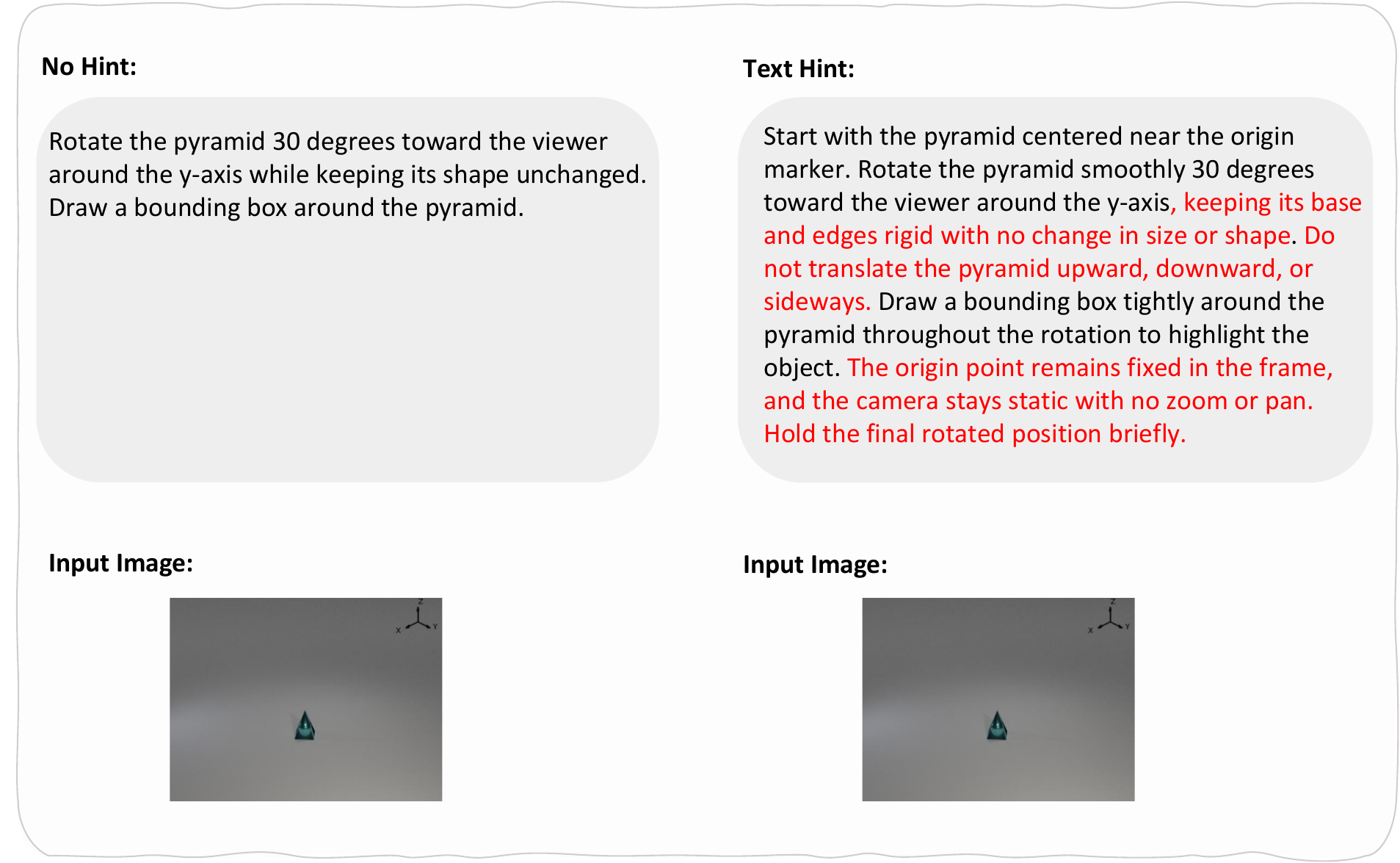}
\end{figure}

\begin{figure}[h]
\centering
\includegraphics[width=\linewidth]{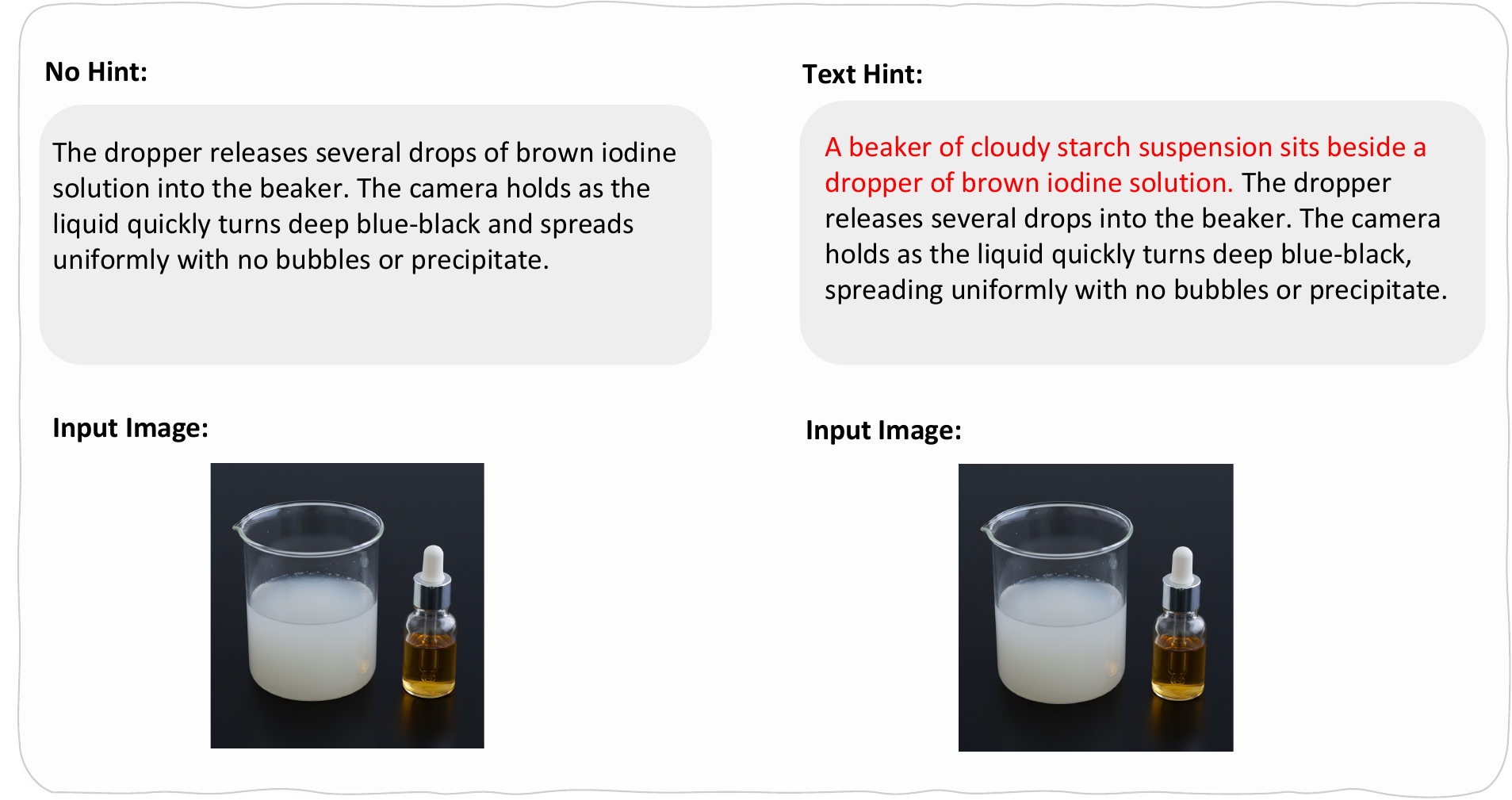}
\end{figure}

\begin{figure}[h]
\centering
\includegraphics[width=\linewidth]{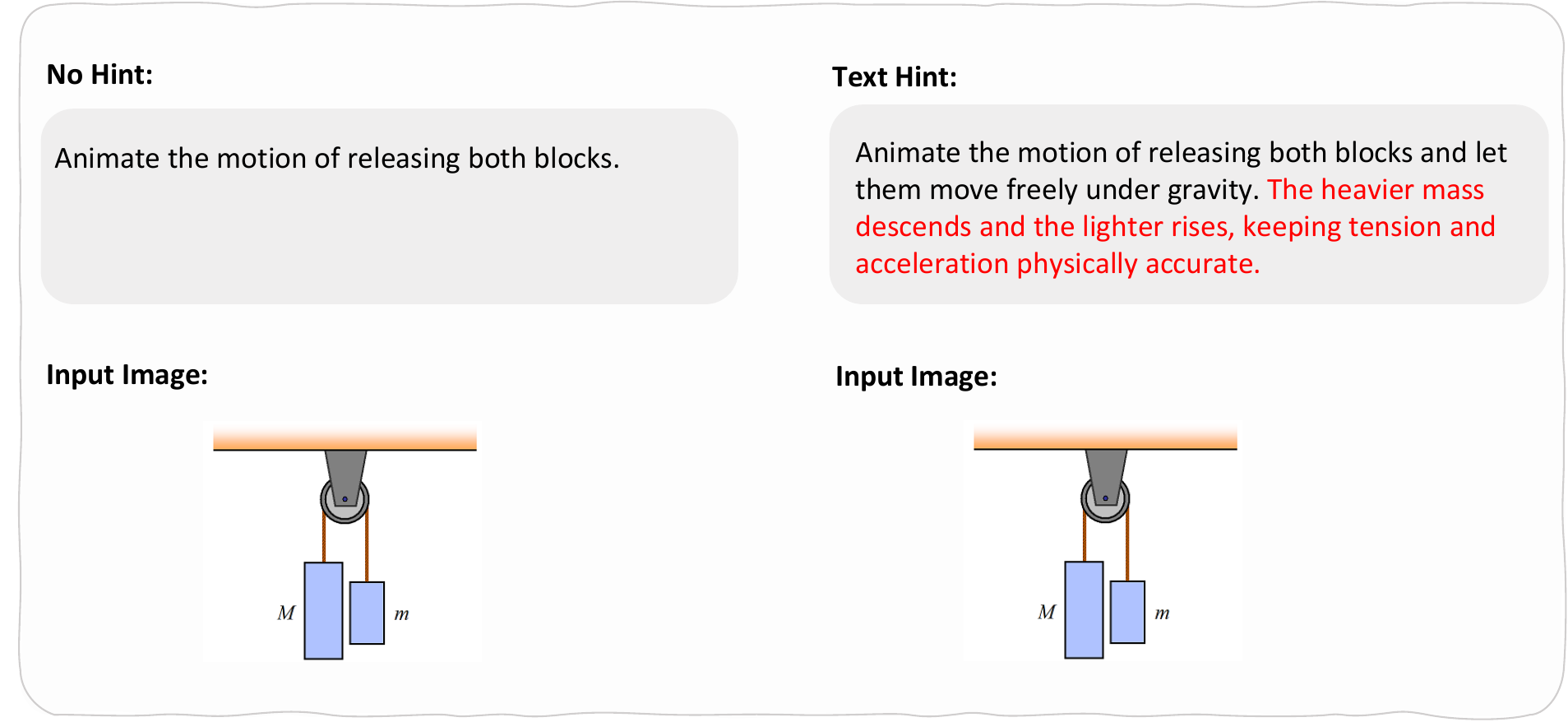}
\end{figure}

\begin{figure}[h]
\centering
\includegraphics[width=\linewidth]{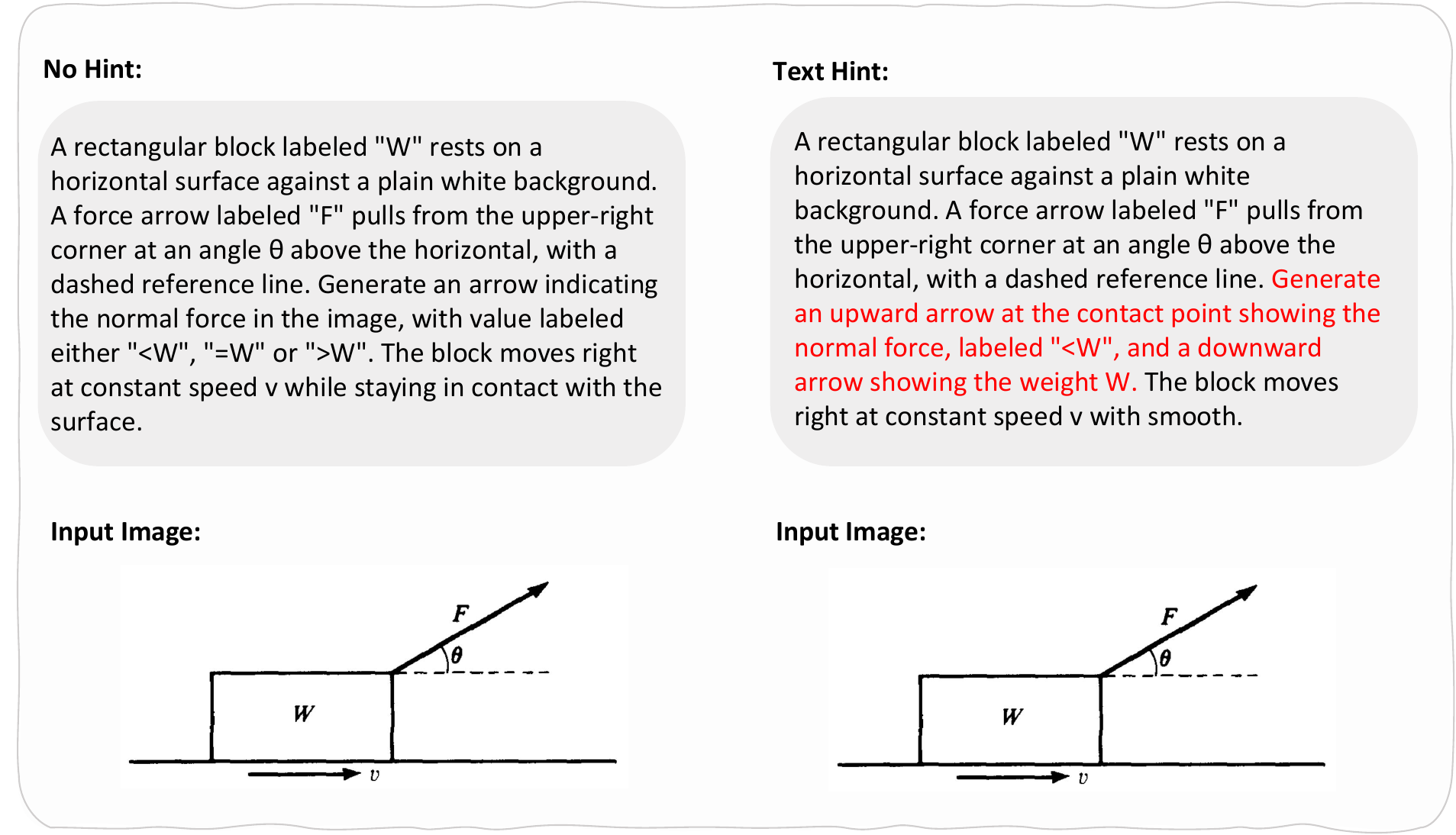}
\end{figure}

\begin{figure}[h]
\centering
\includegraphics[width=\linewidth]{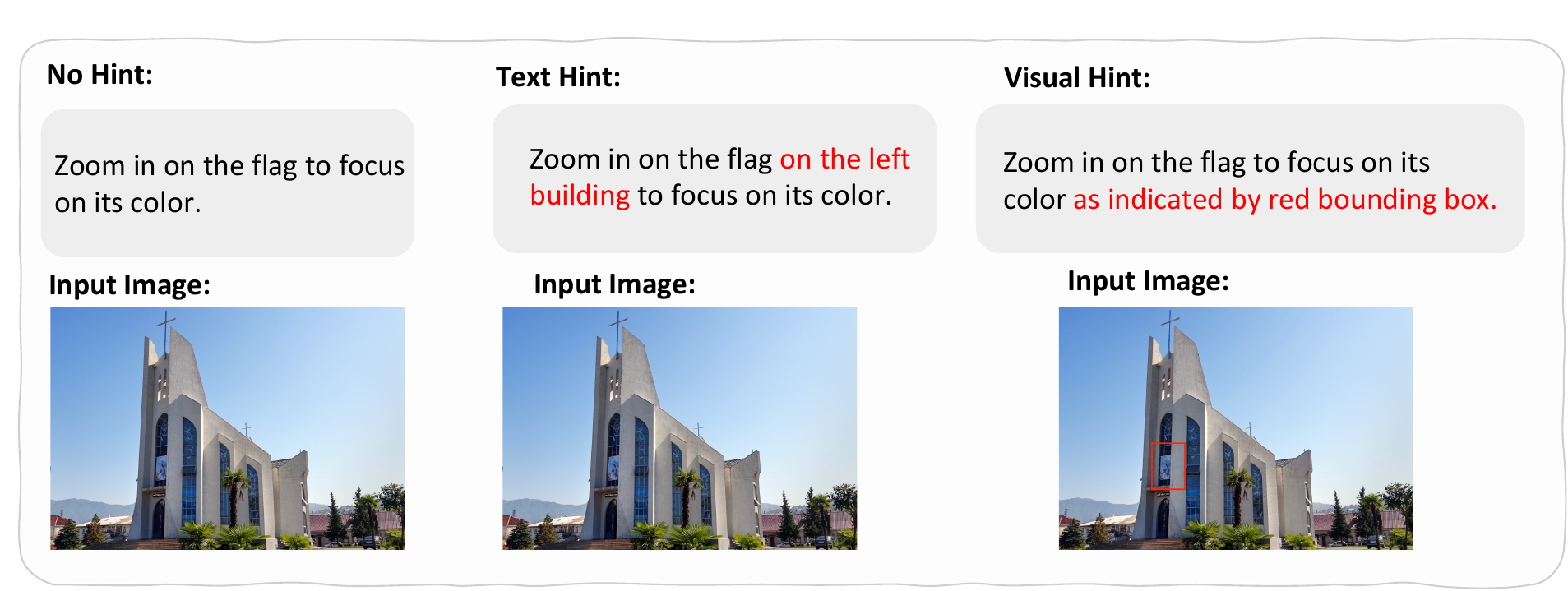}
\end{figure}

\begin{figure}[h]
\centering
\includegraphics[width=\linewidth]{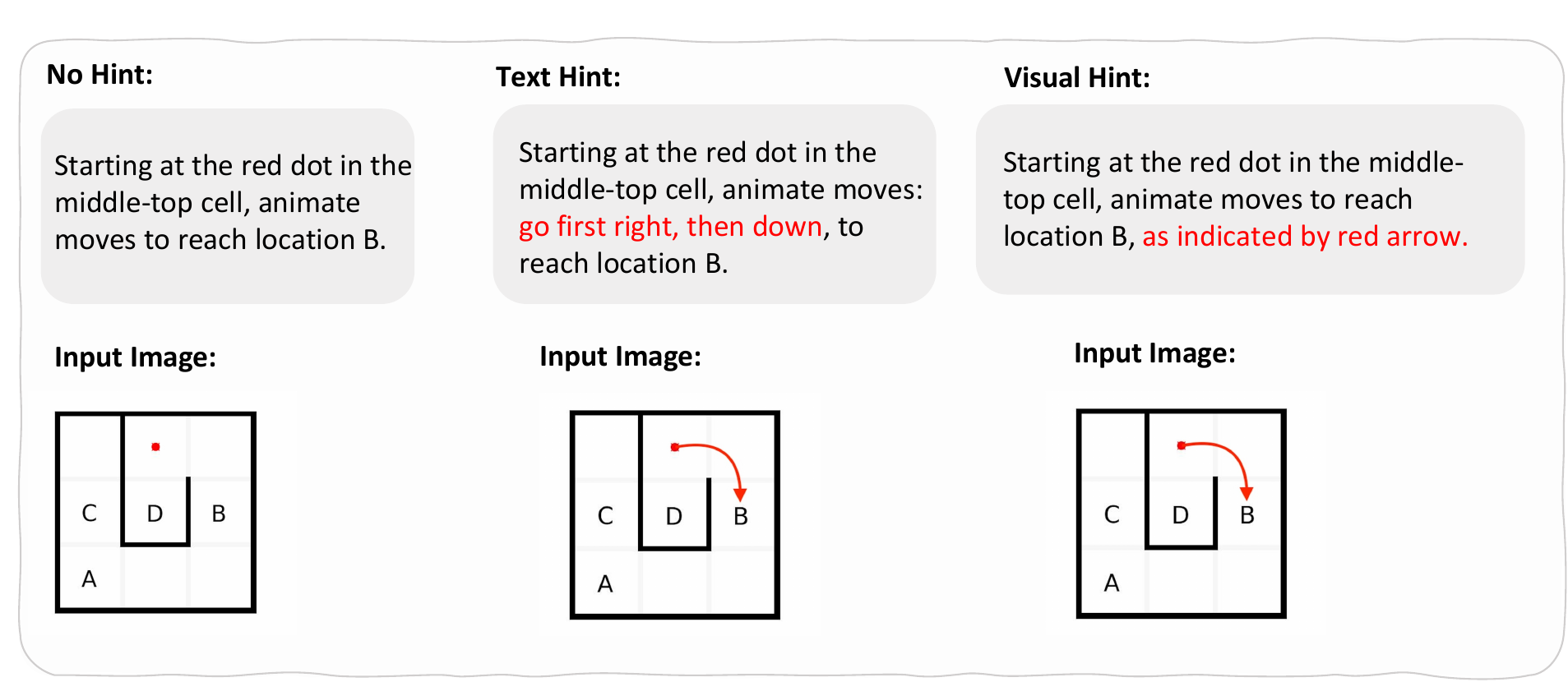}
\end{figure}

\begin{figure}[h]
\centering
\includegraphics[width=\linewidth]{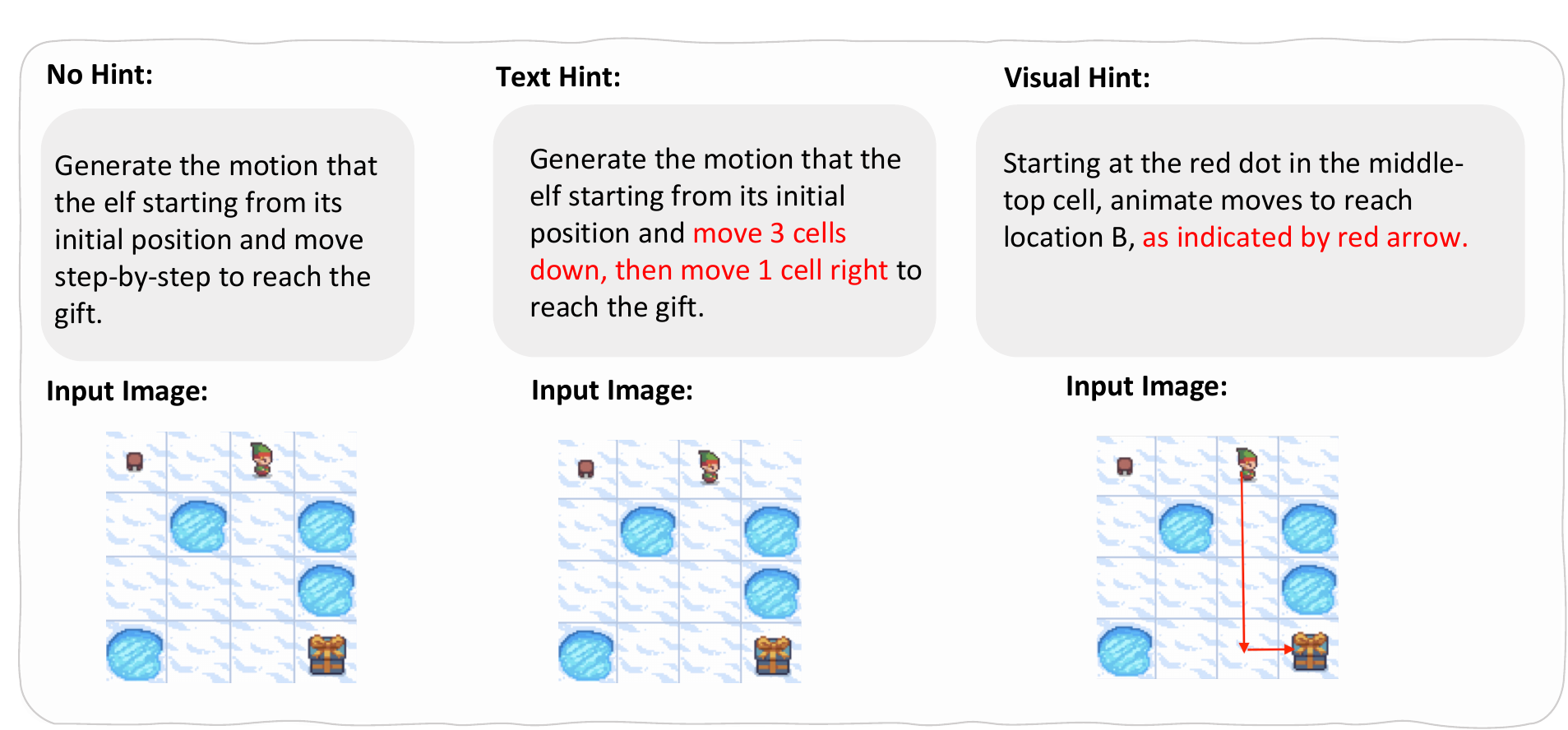}
\end{figure}

\begin{figure}[h]
\centering
\includegraphics[width=\linewidth]{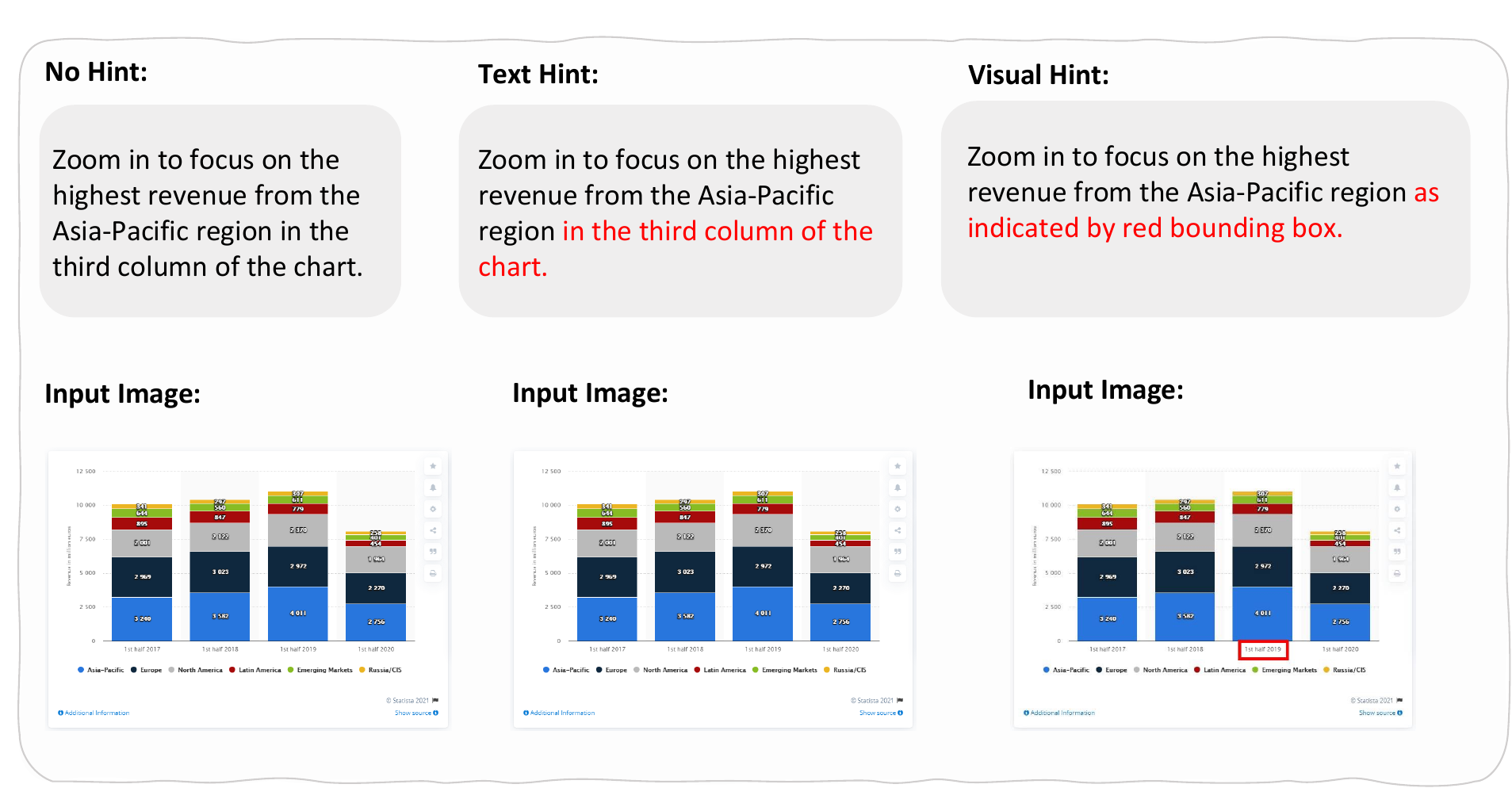}
\end{figure}

\begin{figure}[h]
\centering
\includegraphics[width=\linewidth]{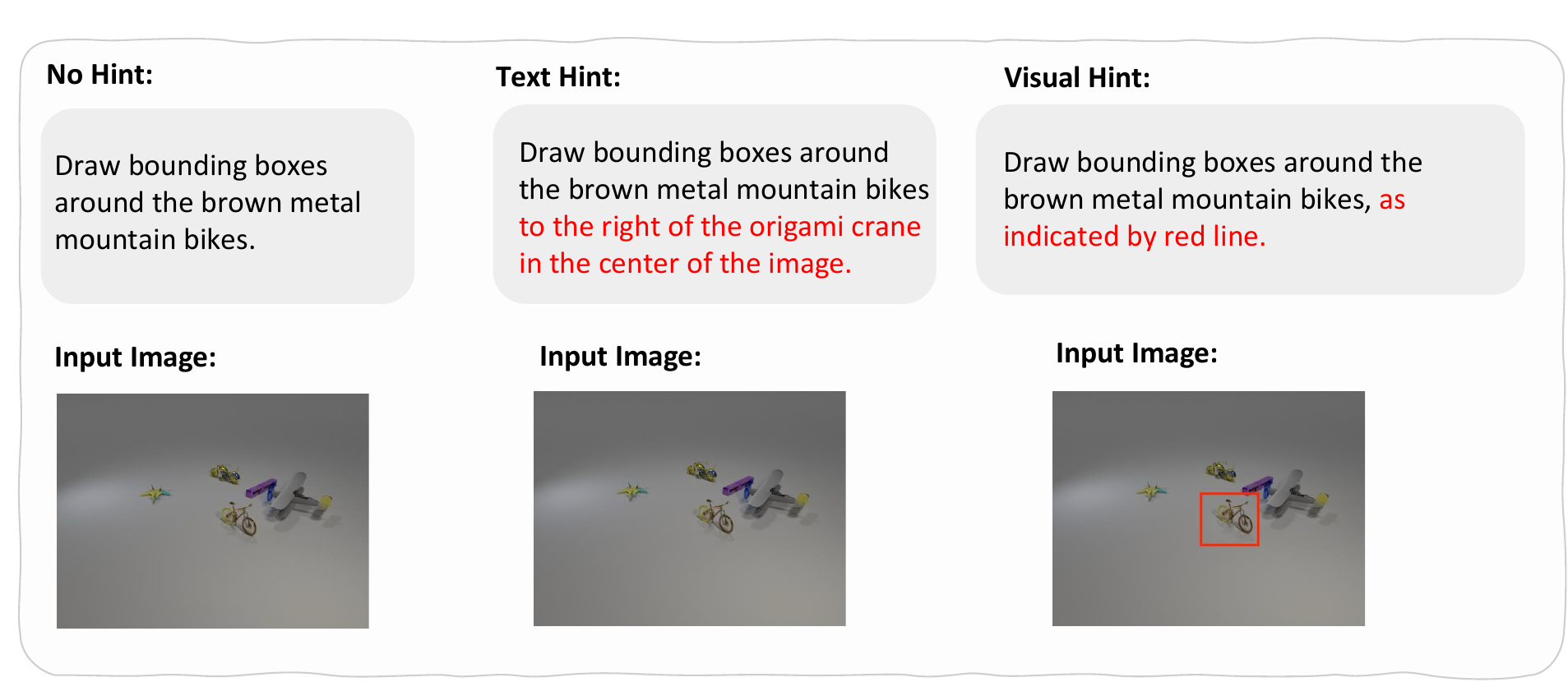}
\end{figure}

\begin{figure}[h]
\centering
\includegraphics[width=\linewidth]{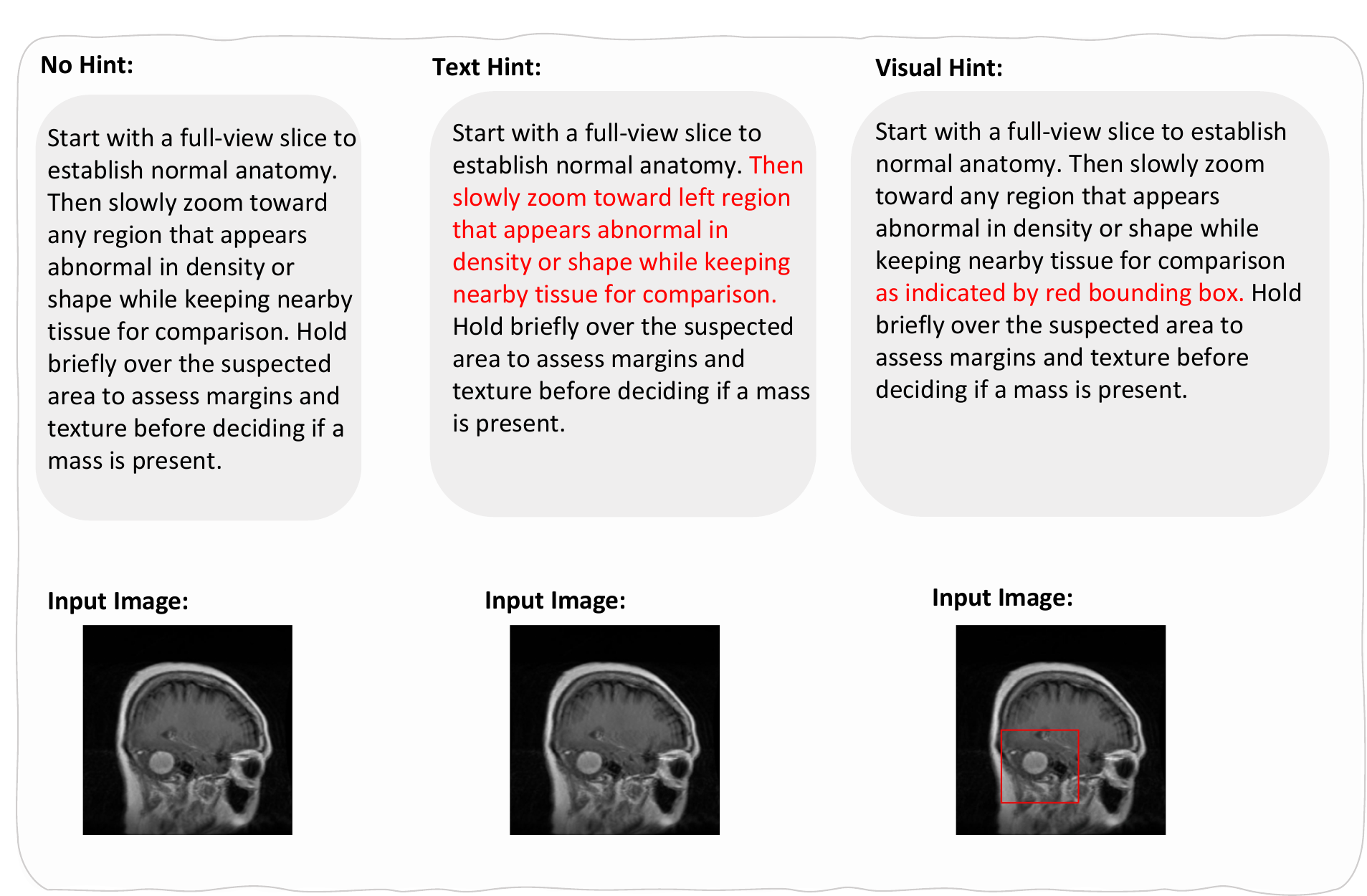}
\end{figure}

\clearpage

\section{Visualization Examples of Text and Visual Hint}

We provide following examples to visualize the effect of text and visual hint to the generated videos.

\begin{figure}[h]
\centering
\includegraphics[width=\linewidth]{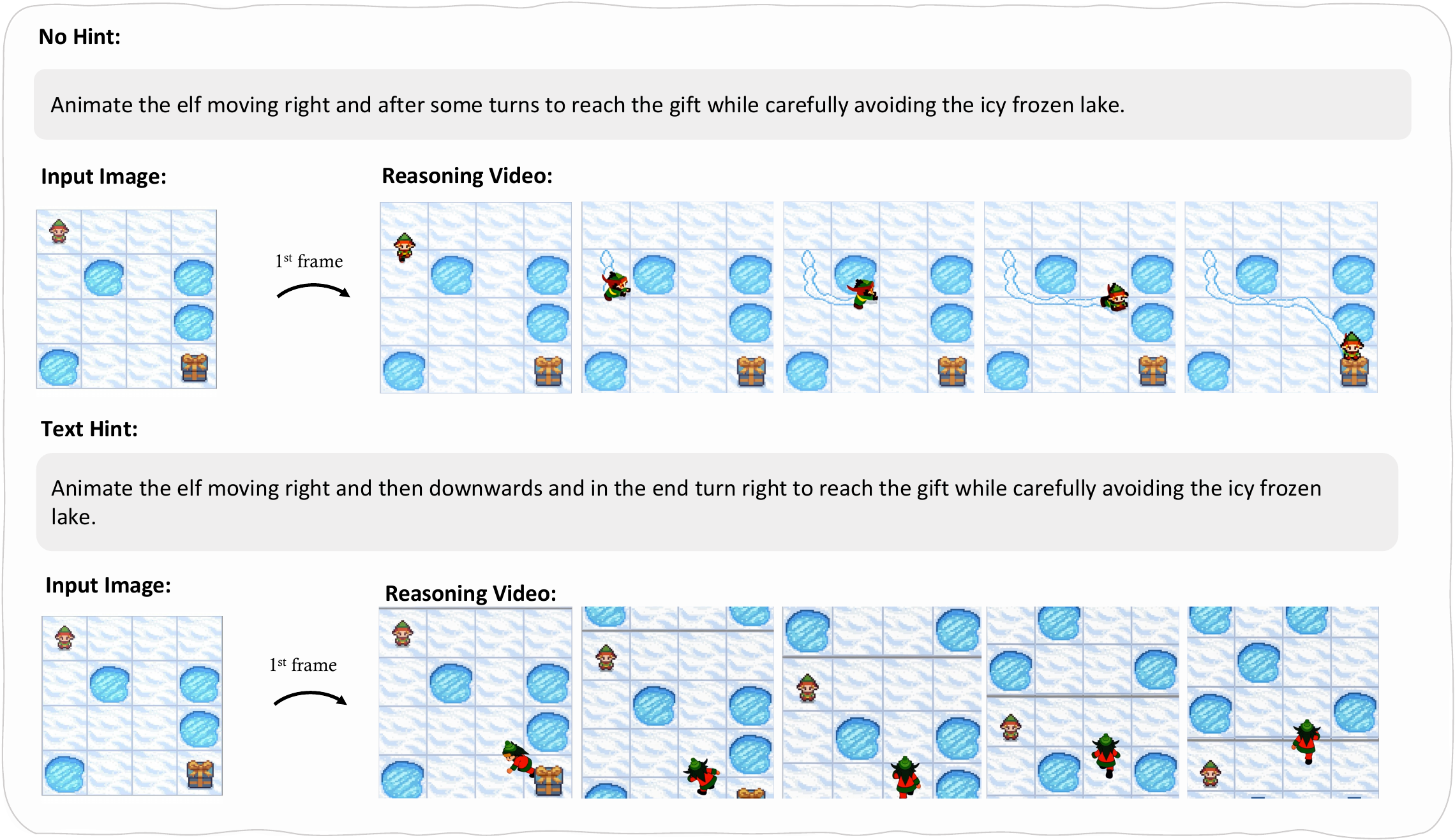}
\end{figure}

\begin{figure}[h]
\centering
\includegraphics[width=\linewidth]{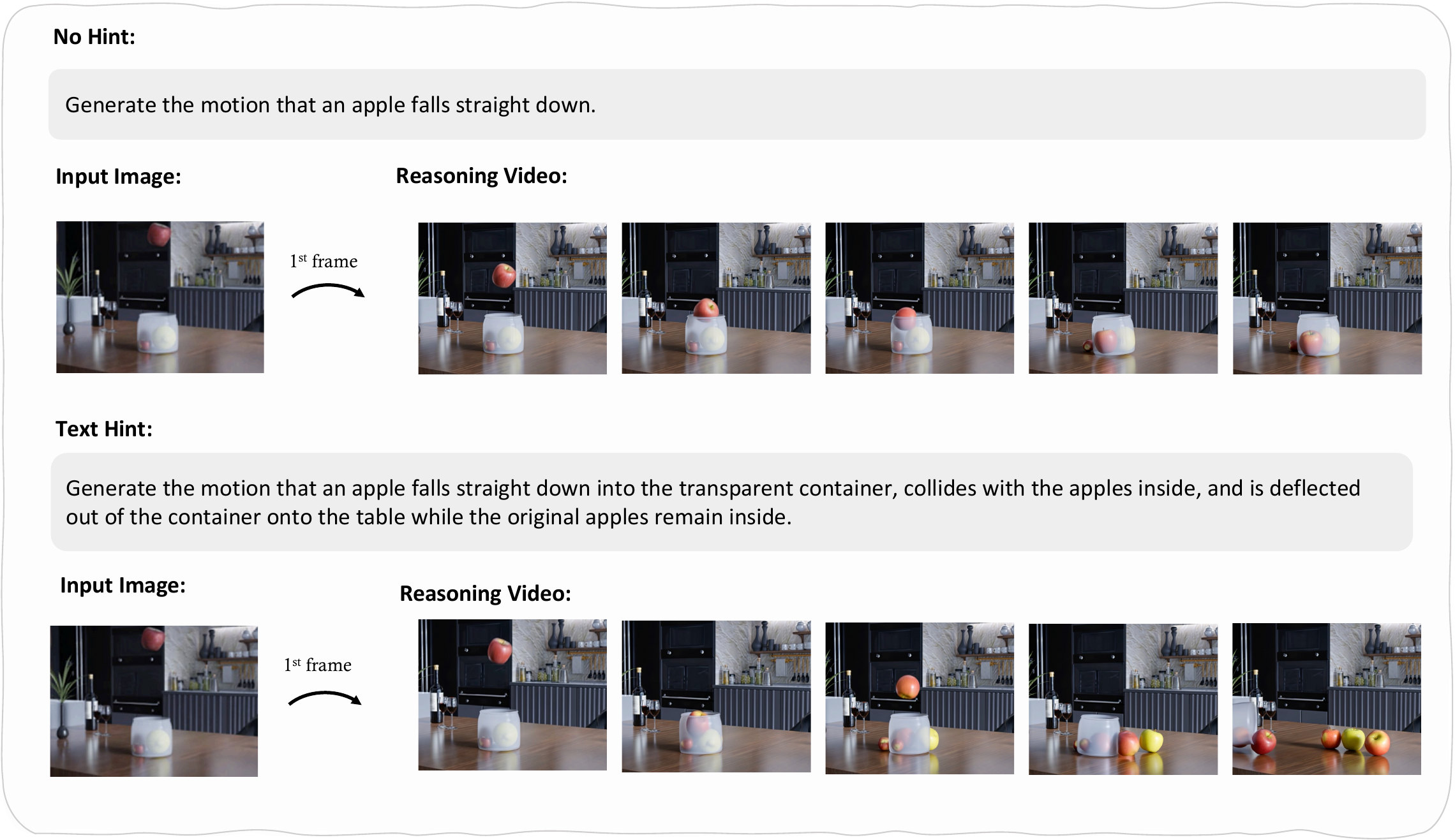}
\end{figure}

\begin{figure}[h]
\centering
\includegraphics[width=\linewidth]{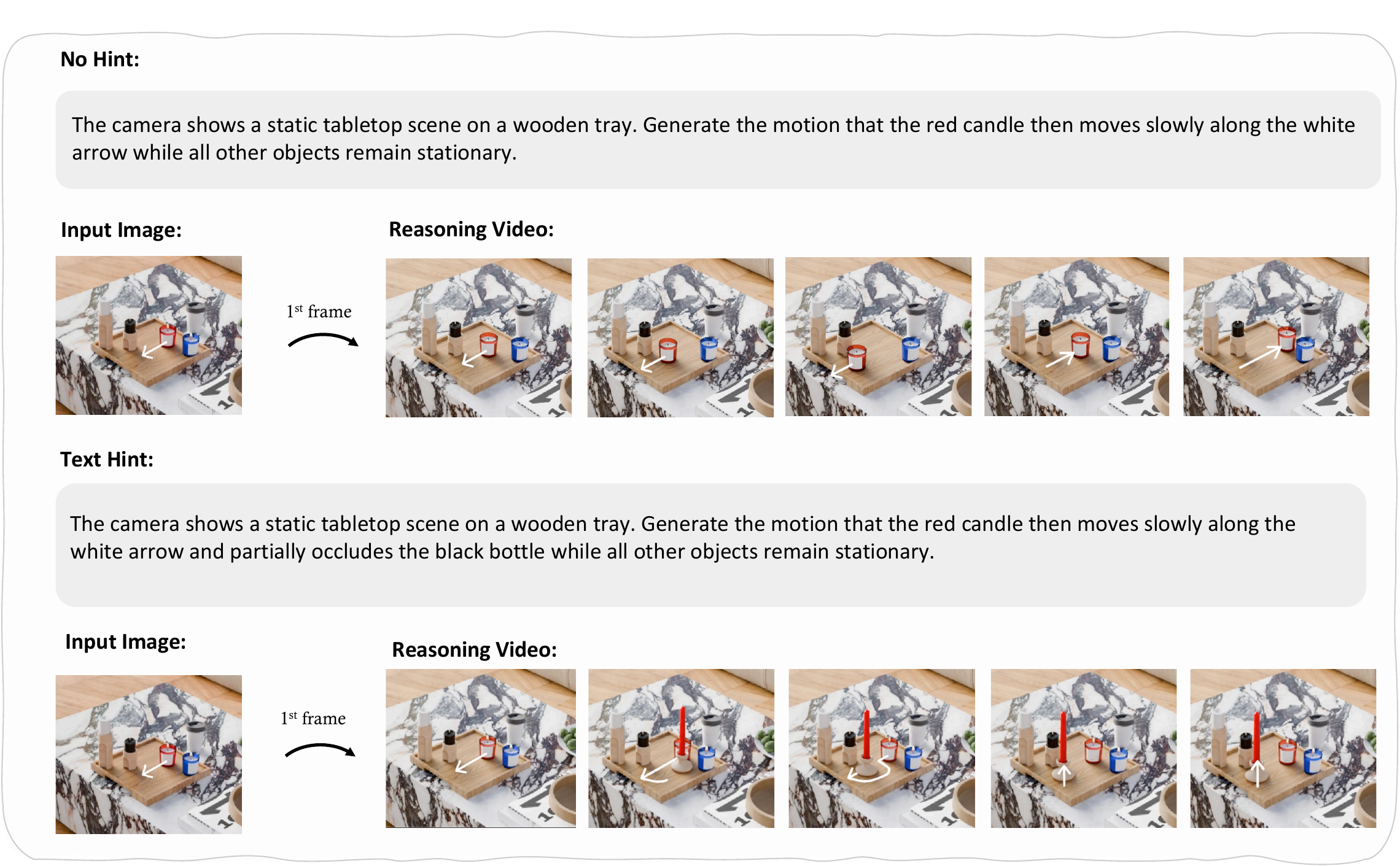}
\end{figure}

\begin{figure}[h]
\centering
\includegraphics[width=\linewidth]{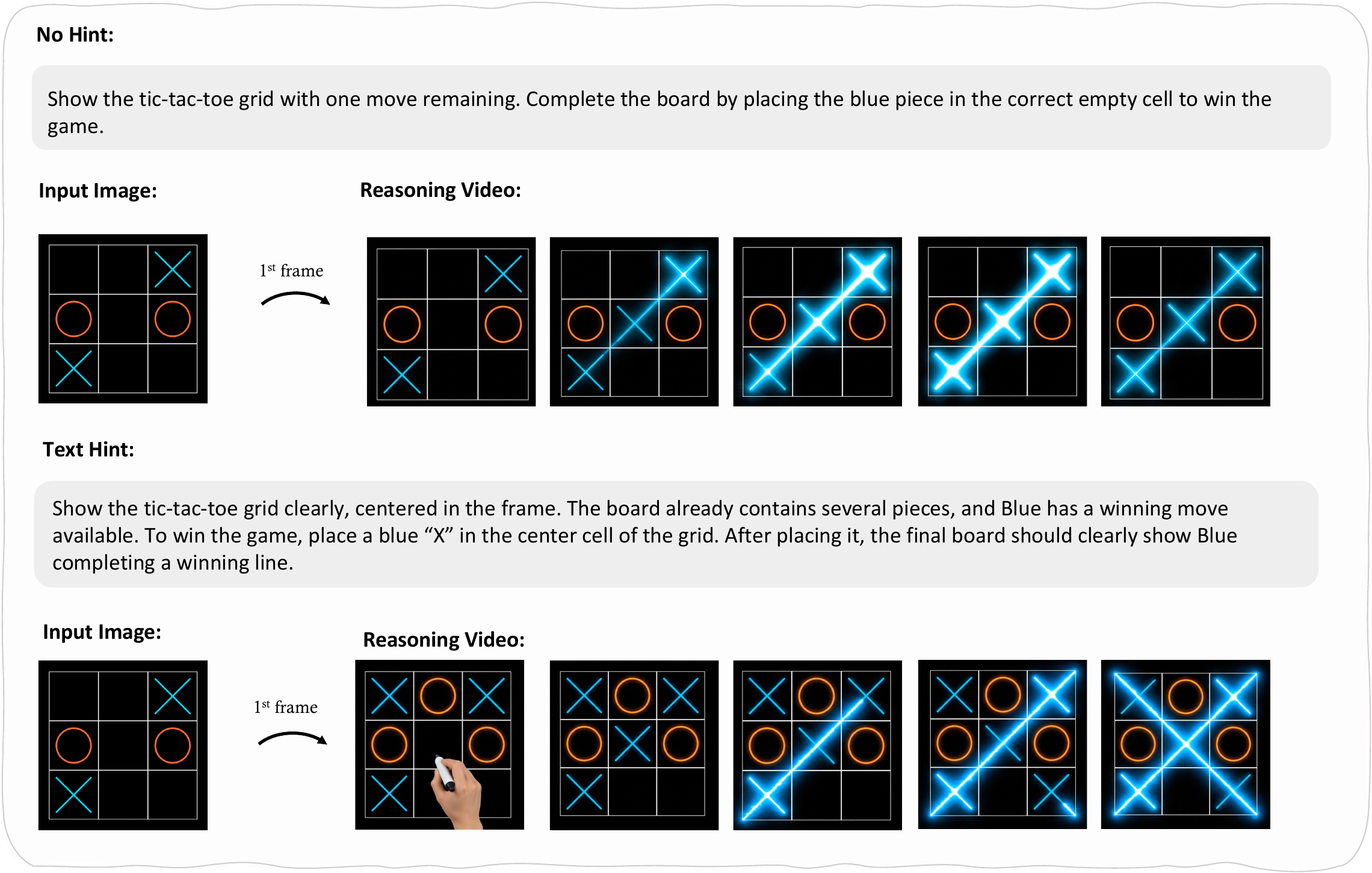}
\end{figure}

\begin{figure}[h]
\centering
\includegraphics[width=\linewidth]{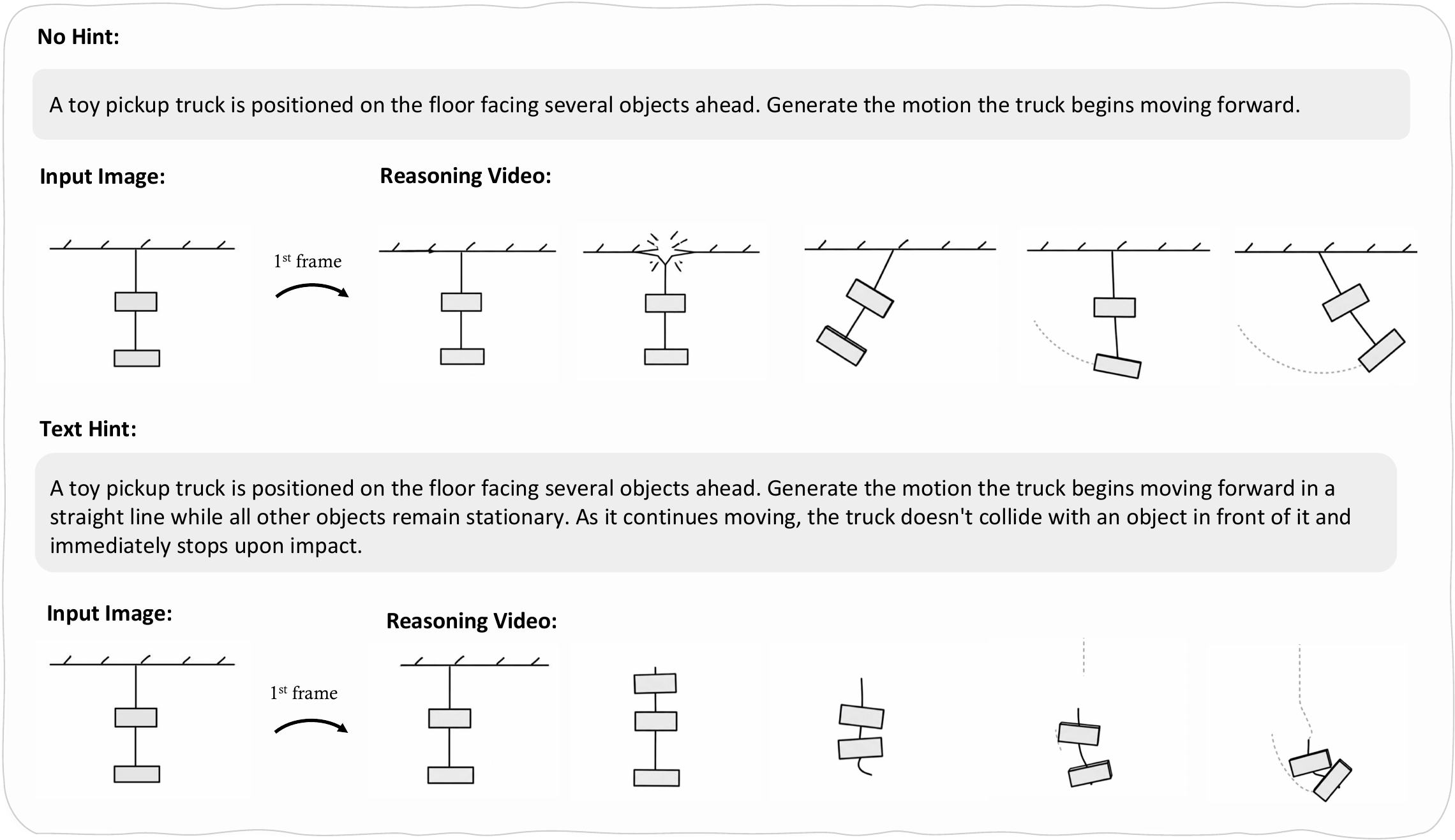}
\end{figure}

\begin{figure}[h]
\centering
\includegraphics[width=\linewidth]{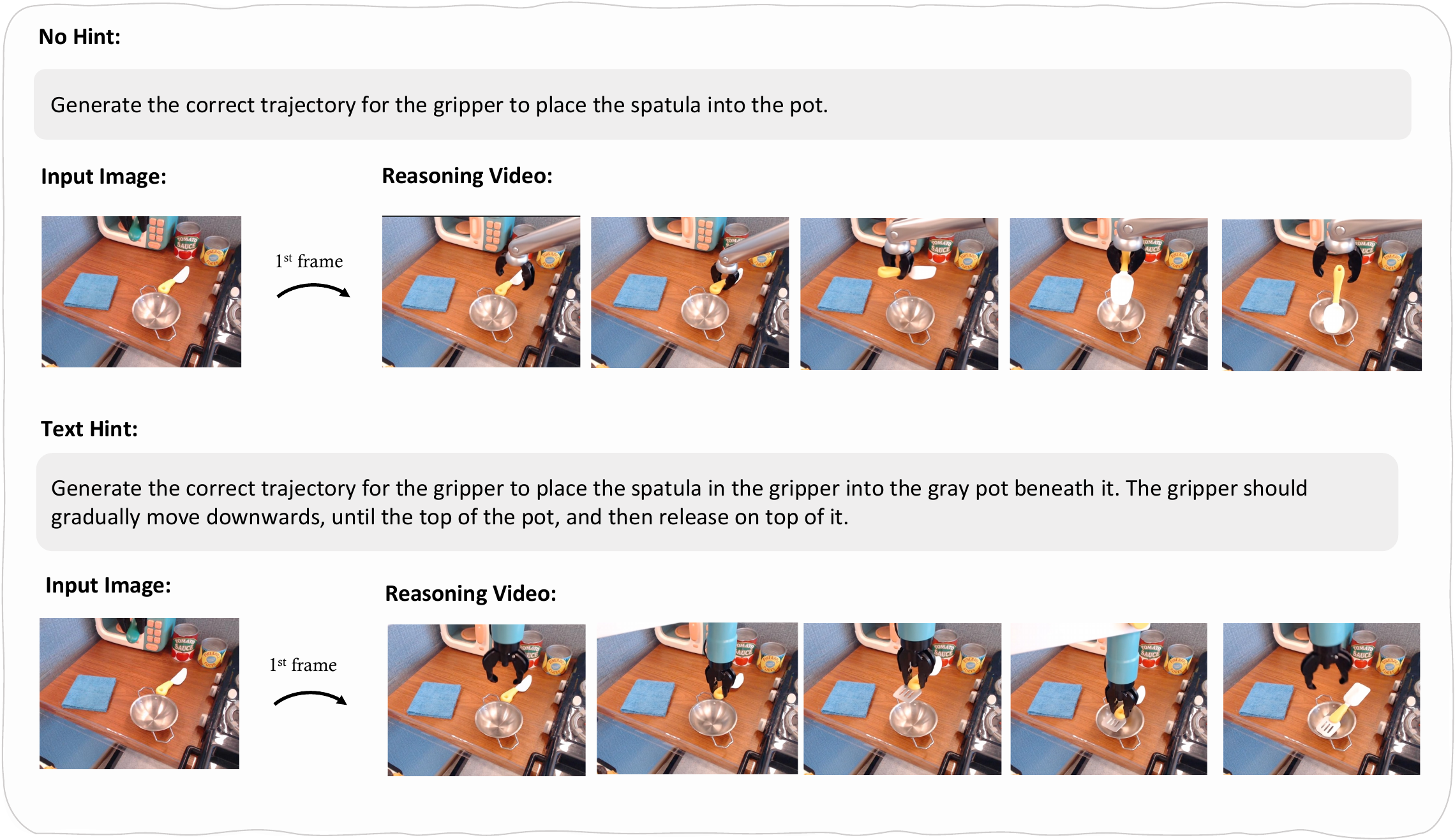}
\end{figure}

\begin{figure}[h]
\centering
\includegraphics[width=\linewidth]{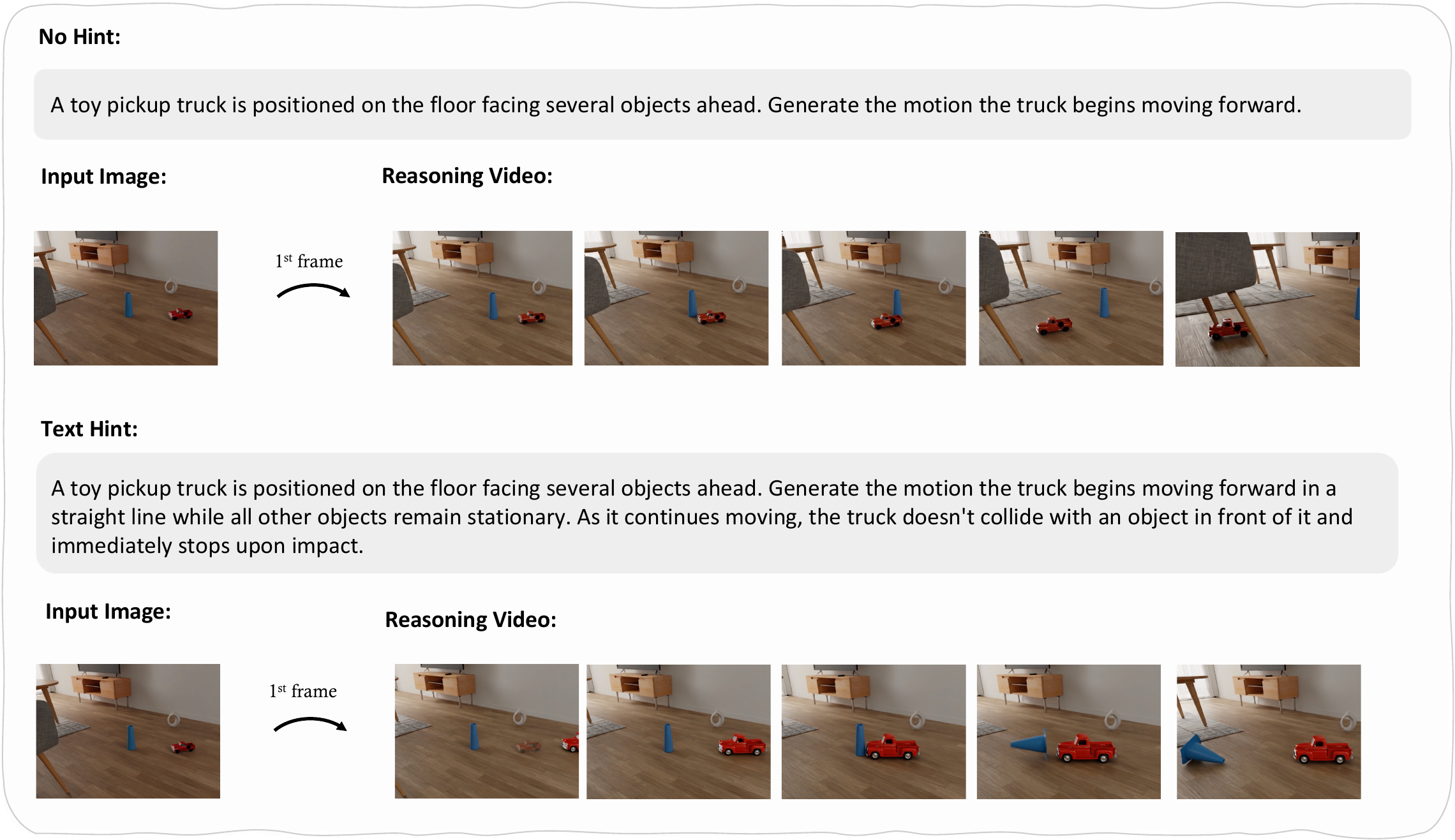}
\end{figure}

\begin{figure}[h]
\centering
\includegraphics[width=\linewidth]{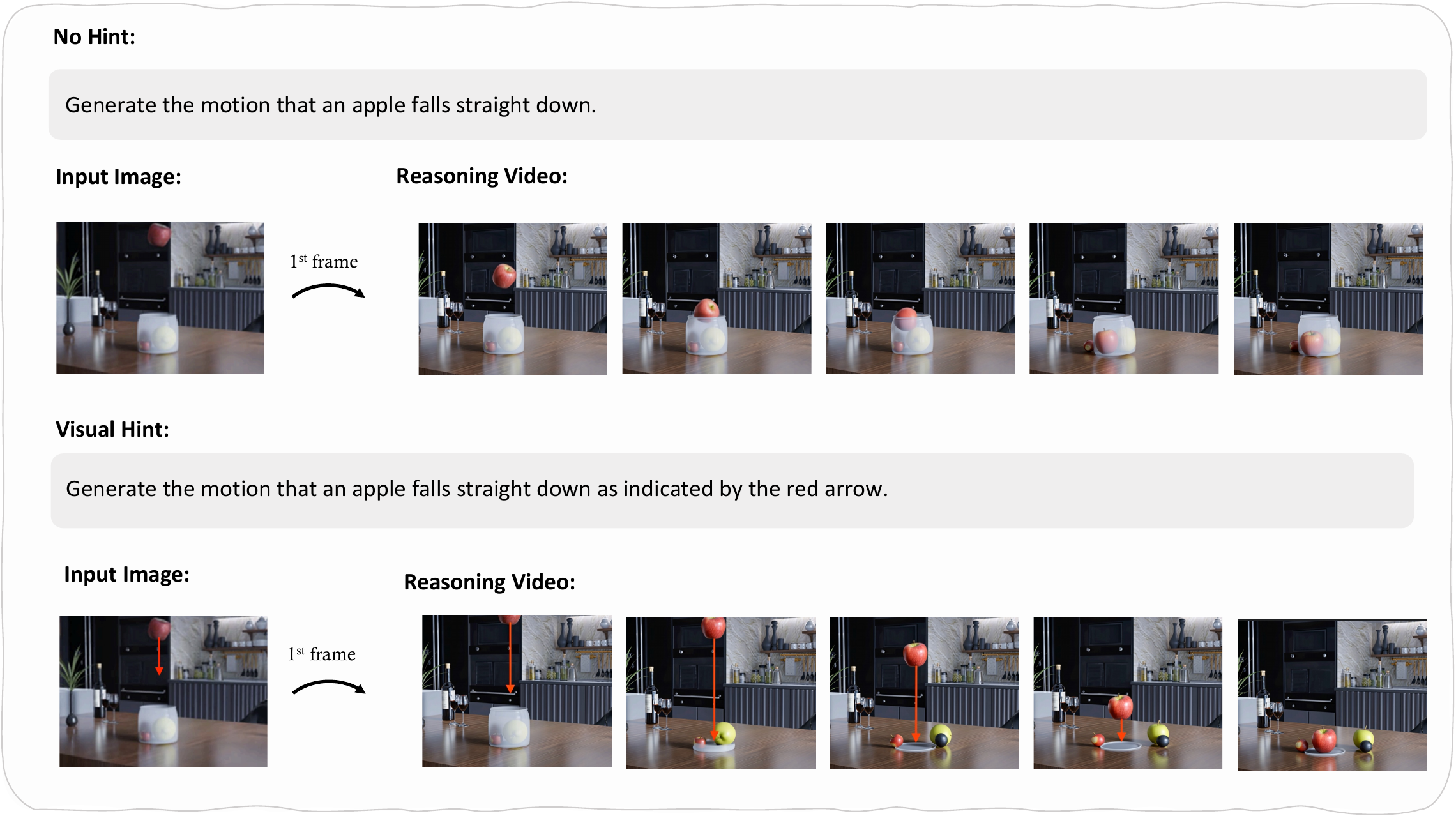}
\end{figure}

\begin{figure}[h]
\centering
\includegraphics[width=\linewidth]{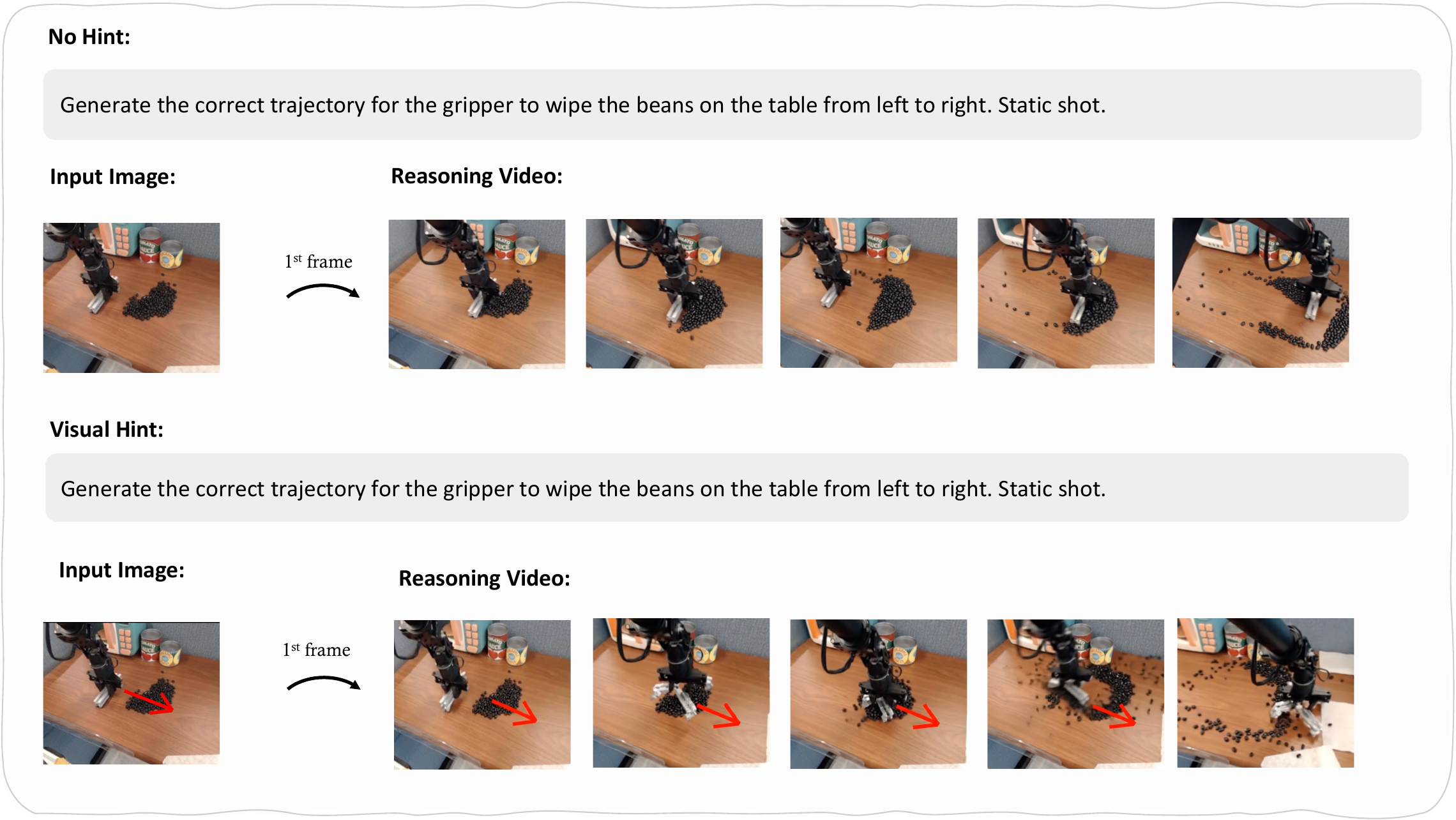}
\end{figure}

\begin{figure}[h]
\centering
\includegraphics[width=\linewidth]{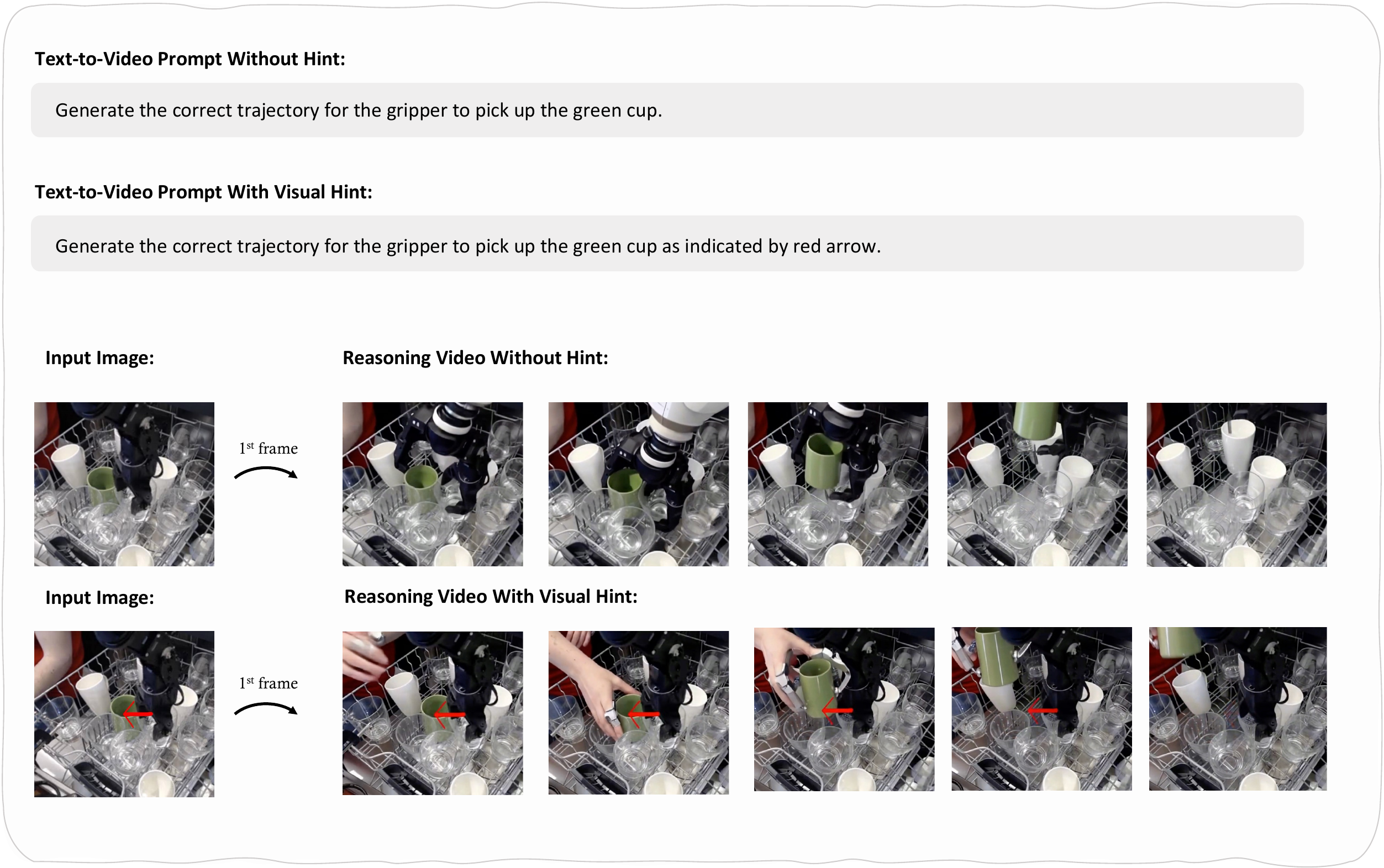}
\end{figure}

\end{document}